\documentclass[final,1p,times]{elsarticle}

\usepackage{etoolbox}

\usepackage{etoolbox}
\makeatletter
\patchcmd{\ps@pprintTitle}
  {Preprint submitted}
  {To be submitted}
  {}{}
\makeatother

\usepackage{amssymb}
\usepackage{hyperref}

\usepackage{longtable}
\usepackage{amsmath}
\usepackage[utf8]{inputenc} 
\usepackage[T1]{fontenc}    
\usepackage{hyperref}       
\usepackage[utf8]{inputenc}
\usepackage{url}            
\usepackage{booktabs}       
\usepackage{amsfonts}       
\usepackage{nicefrac}       
\usepackage{microtype}      
\usepackage{lipsum}
\usepackage{balance}
\usepackage[table,xcdraw]{xcolor}    
\usepackage{array}                   
\usepackage{pifont}                 
\usepackage{graphicx}               
\usepackage{adjustbox}              
\usepackage{fontawesome}           

\usepackage{textgreek} 
\usepackage{academicons}
\usepackage{subcaption}
\usepackage{enumerate}
\usepackage{amssymb}
\usepackage{wrapfig}
\usepackage{tikz}
\usetikzlibrary{positioning}
\usepackage[most]{tcolorbox}

\usepackage{natbib}
\usepackage{graphicx}
\usepackage{wrapfig}
\usepackage{graphicx}
\usepackage{comment}
\usepackage{enumitem}
\usepackage[utf8]{inputenc}     
\usepackage{amssymb}    
\usepackage[table]{xcolor}
\usepackage{fontawesome} 
\usepackage{color}
 \usepackage{array} 
  \usepackage{amsmath} 
 \usepackage{tikz}
 \usepackage{booktabs}   
\usepackage{tabularx}   
\usepackage{ragged2e}   
\usepackage{caption}    
\usepackage{array}      
\usetikzlibrary{mindmap,trees}
\PassOptionsToPackage{table}{xcolor}
\usepackage{booktabs}   
\usepackage{pifont}     
\usepackage{array}      
\usepackage{multirow}   
\definecolor{header-bg}{RGB}{30, 50, 90}         
\definecolor{traditional-bg}{RGB}{255, 245, 245}  
\definecolor{agentic-bg}{RGB}{240, 248, 255}      
\definecolor{cross-red}{RGB}{180, 40, 40}         
\definecolor{check-green}{RGB}{40, 140, 40}       
\definecolor{bullet-blue}{RGB}{30, 80, 150}       
\graphicspath{ {./images/} }
\definecolor{lightgray}{gray}{0.95}
\definecolor{lightblue}{RGB}{220,230,245}
\definecolor{lightgreen}{RGB}{230,245,230}
\definecolor{lightred}{RGB}{255,235,235}

\newcommand{\yes}{\cellcolor{lightgreen}\ding{51}}         
\newcommand{\no}{\cellcolor{lightred}\ding{55}}            
\newcommand{\partialyes}{\cellcolor{lightblue}\textasciitilde{}}  

\usepackage{pifont}

\tcbset{
  colback=white,
  colframe=gray!50!black,
  coltitle=white,
  colbacktitle=black,
  fonttitle=\bfseries,
  boxrule=0.5pt,
  arc=4pt,
  left=6pt,
  right=6pt,
  top=6pt,
  bottom=6pt,
  enhanced,
  sharp corners=south
}

\newtcolorbox{boxblue}{
  colback=blue!10!white, colframe=blue!60!black,
  coltitle=black, fonttitle=\bfseries,
  boxrule=0pt, arc=6pt, auto outer arc,
  width=\linewidth, sharp corners=south,
  boxsep=5pt, left=5pt, right=5pt, top=4pt, bottom=4pt,
  enhanced, drop shadow
}
\newtcolorbox{boxgreen}{
  colback=green!10!white, colframe=green!60!black,
  coltitle=black, fonttitle=\bfseries,
  boxrule=0pt, arc=6pt, auto outer arc,
  width=\linewidth, sharp corners=south,
  boxsep=5pt, enhanced, drop shadow
}
\newtcolorbox{boxorange}{
  colback=orange!15!white, colframe=orange!60!black,
  coltitle=black, fonttitle=\bfseries,
  boxrule=0pt, arc=6pt, auto outer arc,
  width=\linewidth, boxsep=5pt, enhanced, drop shadow
}
\newtcolorbox{boxpurple}{
  colback=purple!10!white, colframe=purple!60!black,
  coltitle=black, fonttitle=\bfseries,
  boxrule=0pt, arc=6pt, auto outer arc,
  width=\linewidth, boxsep=5pt, enhanced, drop shadow
}
\newtcolorbox{boxred}{
  colback=red!10!white, colframe=red!60!black,
  coltitle=black, fonttitle=\bfseries,
  boxrule=0pt, arc=6pt, auto outer arc,
  width=\linewidth, boxsep=5pt, enhanced, drop shadow
}
\newtcolorbox{boxcyan}{
  colback=cyan!10!white, colframe=cyan!60!black,
  coltitle=black, fonttitle=\bfseries,
  boxrule=0pt, arc=6pt, auto outer arc,
  width=\linewidth, boxsep=5pt, enhanced, drop shadow
}
\newtcolorbox{boxyellow}{
  colback=yellow!20!white, colframe=yellow!70!black,
  coltitle=black, fonttitle=\bfseries,
  boxrule=0pt, arc=6pt, auto outer arc,
  width=\linewidth, boxsep=5pt, enhanced, drop shadow
}
\newtcolorbox{boxpink}{
  colback=pink!10!white, colframe=pink!60!black,
  coltitle=black, fonttitle=\bfseries,
  boxrule=0pt, arc=6pt, auto outer arc,
  width=\linewidth, boxsep=5pt, enhanced, drop shadow
}
\DeclareUnicodeCharacter{202F}{\,}
\begin{document}
\begin{frontmatter}

\title{TRiSM for Agentic AI: A Review of Trust, Risk, and Security Management in LLM-based Agentic Multi-Agent Systems}

\author[inst1]{Shaina Raza\corref{cor1}\fnref{equal}}
\ead{shaina.raza@torontomu.ca}

\author[inst2]{Ranjan Sapkota\corref{cor1}\fnref{equal}}
\ead{rs2672@cornell.edu}

\author[inst2]{Manoj Karkee}
\ead{mk2684@cornell.edu}

\author[inst3]{Christos Emmanouilidis\corref{cor1}}
\ead{c.emmanouilidis@rug.nl}

\address[inst1]{Vector Institute, Toronto, Canada}
\address[inst2]{Cornell University, USA}
\address[inst3]{University of Groningen, Netherlands}

\cortext[cor1]{Corresponding authors}
\fntext[equal]{Shaina Raza and Ranjan Sapkota contributed equally to this work.}

\begin{abstract}
 Agentic AI systems, built upon large language models (LLMs) and deployed in multi-agent configurations, are redefining intelligence, autonomy, collaboration, and decision-making across enterprise and societal domains. This review presents a structured analysis of Trust, Risk, and Security Management (TRiSM) in the context of LLM-based Agentic Multi-Agent Systems (AMAS). We begin by examining the conceptual foundations of Agentic AI and highlight its architectural distinctions from traditional AI agents. We then adapt and extend the AI TRiSM framework for Agentic AI, structured around key pillars: \textit{ Explainability, ModelOps, Security,  Privacy} and \textit{their Lifecycle Governance}, each contextualized to the challenges of AMAS. A risk taxonomy is proposed to capture the unique threats and vulnerabilities of Agentic AI, ranging from coordination failures to prompt-based adversarial manipulation.
To make coordination and tool use measurable in practice, we propose two metrics: the Component Synergy Score (CSS), which captures inter-agent enablement, and the Tool Utilization Efficacy (TUE), which evaluates whether tools are invoked correctly and efficiently. We further discuss strategies for improving explainability in Agentic AI, as well as approaches to enhancing security and privacy through encryption, adversarial robustness, and regulatory compliance. The review concludes with a research roadmap for the responsible development and deployment of Agentic AI, highlighting key directions to align emerging systems with TRiSM principles-ensuring safety, transparency, and accountability in their operation.

\end{abstract}

\begin{keyword}
 Agentic AI, LLM-based Multi-Agent Systems, TRiSM, AI Governance, Explainability, ModelOps, Application Security, Model Privacy, AI Agents, Trustworthy AI, Risk Management, AI Safety, Privacy-Preserving AI, Adversarial Robustness, Human-in-the-Loop
 \end{keyword}

\end{frontmatter}

\section{Introduction}
\label{introduction}
AI governance has moved from aspiration to obligation. 
The EU Artificial Intelligence Act \cite{EU_AI_Act2025} entered into force on 1 August 2024 and applies progressively (e.g., prohibited AI practices and AI literacy obligations from 2 February 2025, and governance rules plus obligations for general-purpose AI models from 2 August 2025)\footnote{\href{https://digital-strategy.ec.europa.eu/en/policies/regulatory-framework-ai}{EU AI Act}}; meanwhile, organizations are adopting management frameworks such as ISO/IEC 42001:2023 \cite{iso42001} and the NIST AI Risk Management Framework (AI RMF 1.0) \cite{ai2023artificial} to operationalize risk controls. In parallel, enterprise use of AI continues to expand, for example, recent global surveys \cite{McKinsey_StateOfAI2025, Stanford_HAI_AIIndex2025,Gartner_AI_TRiSM2024} report that a substantial majority of organizations use AI in at least one business function. Critically, the emerging paradigm of AI agents, which are systems that autonomously plan, reason, and execute complex tasks, introduces qualitatively different risks than traditional AI applications. These trends demand system-level trust, risk, and security management tailored to \emph{agentic} systems.

\begin{figure}[t]
    \centering
    \includegraphics[width=0.98\textwidth]
    {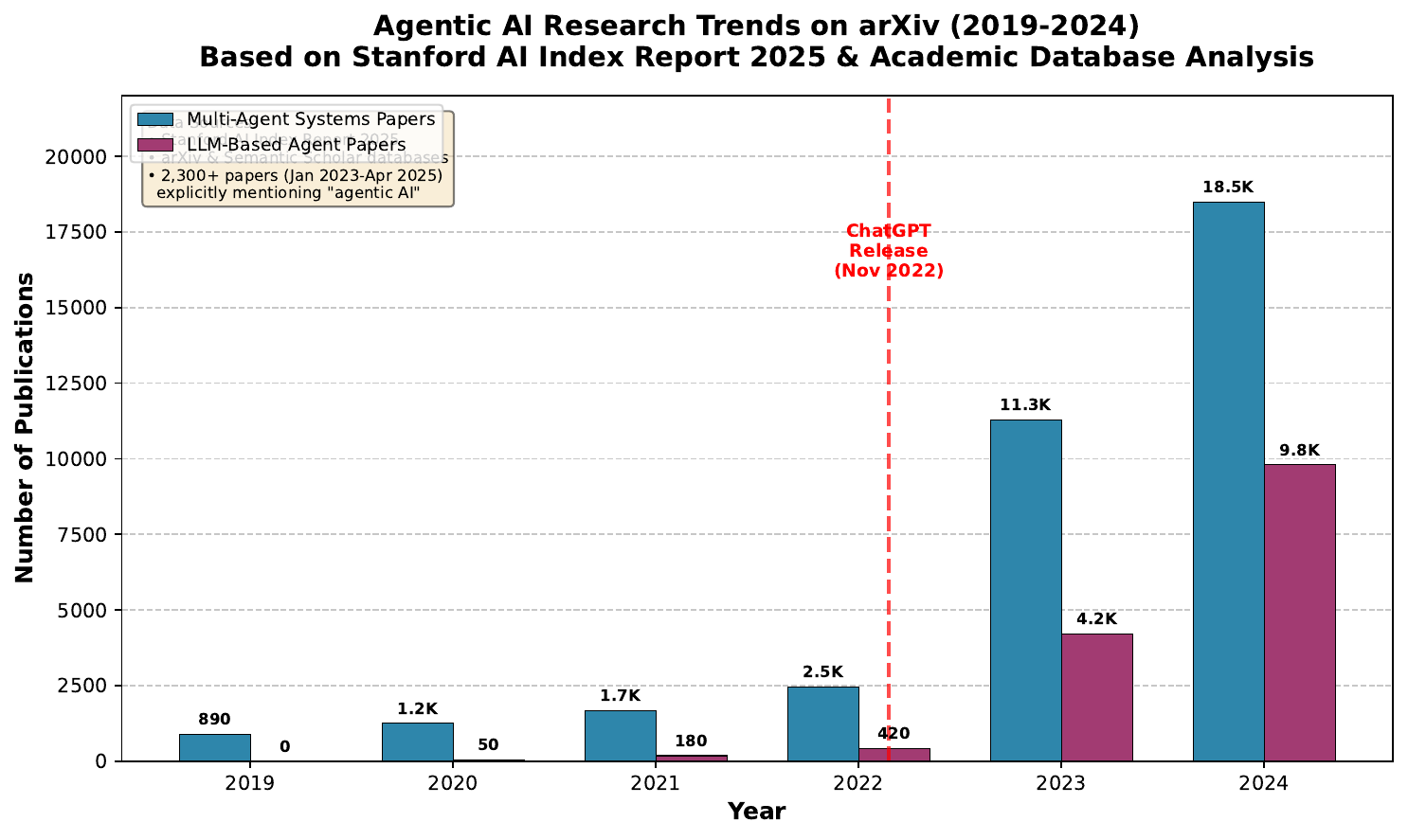}
    \caption{Agentic AI research growth on arXiv (2019--2024), showing multi-agent systems and LLM-based agent publications with marked acceleration following ChatGPT's release in November 2022.}
    \label{fig:arxiv-trends}
\end{figure}

\begin{figure}[h]
    \centering
    \includegraphics[width=0.98\textwidth]
    {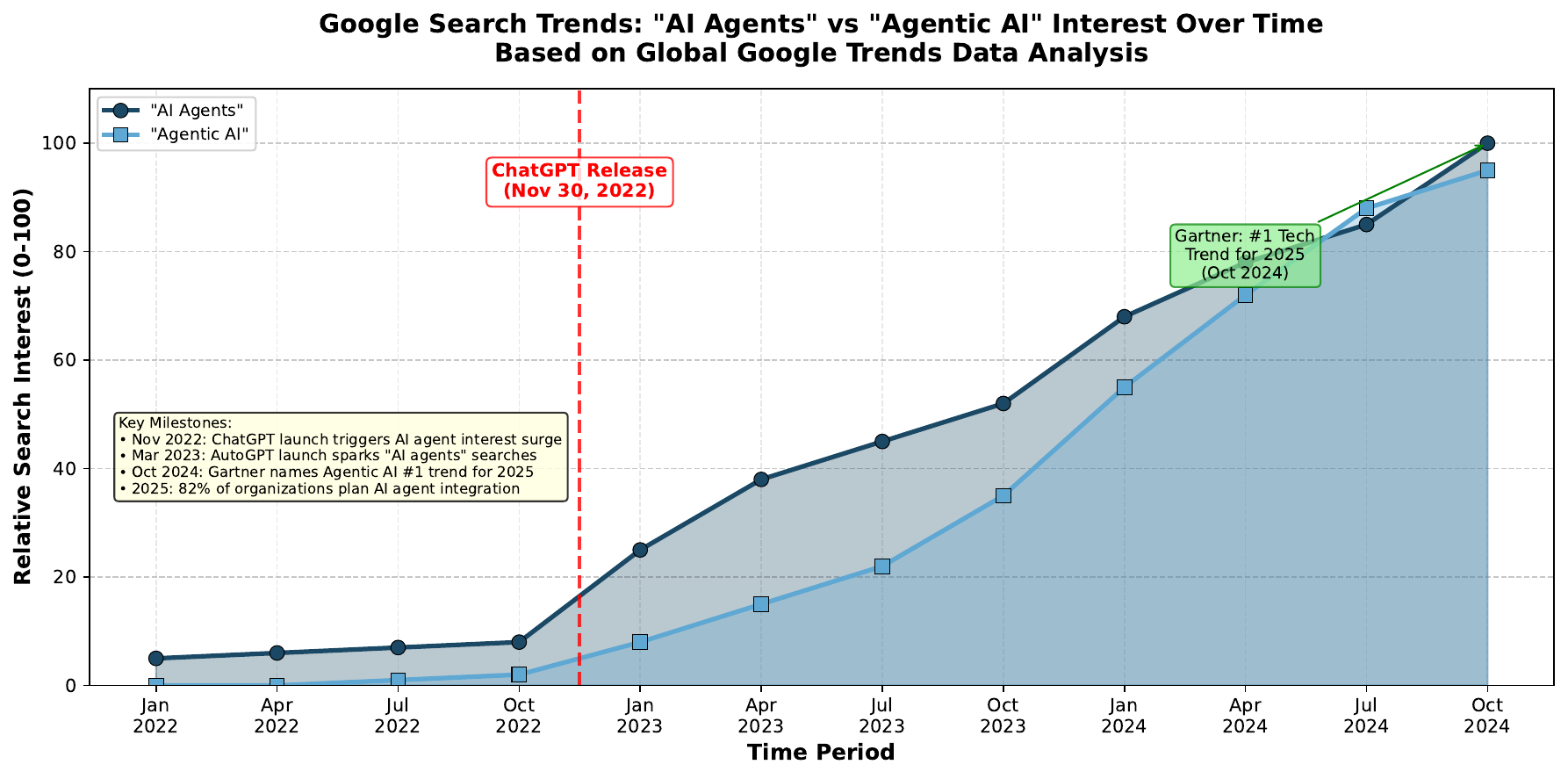}
    \caption{Google search interest for ``AI Agents'' vs ``Agentic AI'' (2022--2024), illustrating rising public awareness with key milestones including ChatGPT launch and Gartner's recognition of Agentic AI as the top technology trend for 2025.}
    \label{fig:google-trends}
\end{figure}

An \emph{AI agent} is a computational entity that perceives its environment and acts to achieve goals \cite{russell1997rationality}. What were once task-specific and largely deterministic programs have rapidly evolved into LLM-powered \emph{agentic} systems with planning, tool use, and persistent memory \cite{sapkota2025ai}. In this paper, we refer to such LLM-based, coordinating agent systems as \emph{Agentic Multi-Agent Systems (AMAS)}. These systems orchestrate multiple specialized agents to address complex tasks, coordinate functions and roles, and adapt workflows through interactions with tools, users, and other agents. 

\paragraph{TRiSM for Agentic AI}
AMAS exhibit emergent and often opaque behaviors, making them susceptible to cascading errors, biased decisions, and unintended interactions \cite{yang2024multi}. These risks are amplified when agents plan, coordinate, and call external tools. Beyond traditional ML concerns (safety, fairness, and interpretability), Agentic AI must contend with threats from tool integrations, prompt-level attacks, memory poisoning, impersonation, and privacy leakage in collaborative settings \cite{lim2012memory, fang2025trustworthyaisafetybias}.  To address these multifaceted system-level risks, \emph{AI Trust, Risk, and Security Management (TRiSM)} frameworks provide integrated approaches emphasizing governance, explainability, security, privacy, and lifecycle controls \cite{IBM2025AITRiSM,habbal2024artificial}.

Integrating TRiSM principles into the design and deployment of Agentic  AI systems is increasingly urgent. Figure~\ref{fig:arxiv-trends} demonstrates explosive growth in arXiv publications, with multi-agent systems papers increasing from 890 in 2019 to 18,500 in 2024, while LLM-based agent papers surged from near zero to 9,800 following ChatGPT's release. This academic interest is mirrored by public awareness, as shown in Figure~\ref{fig:google-trends}, where Google search trends for both ``AI Agents'' and ``Agentic AI'' reached peak levels by late 2024. Gartner’s designation of \emph{Agentic AI} as a top strategic technology trend for 2025, alongside AI governance platforms within its TRiSM framing%
\footnote{\href{https://www.gartner.com/en/documents/market-guide-for-ai-trism}{Gartner Market Guide for AI TRiSM}}, and its 2026 emphasis on multiagent systems, digital provenance, and AI security platforms
\footnote{\href{https://www.gartner.com/en/articles/top-technology-trends-2026}{Gartner Top Technology Trends 2026}} reflect a growing consensus: as AI systems gain autonomy, robust trust, risk mitigation, and security controls become indispensable. We emphasize that a TRiSM perspective is essential to make AMAS deployable in sensitive domains.

\subsection{Necessity and Scope of this Survey}
Much of the Agentic AI literature prioritizes agent modeling, planning, and collaboration, while comparatively less attention is given to adversarial robustness, lifecycle governance, decision provenance, and system-level explainability. This survey addresses this gap by focusing on AMAS\footnote{Throughout, \emph{AMAS} denotes multi-agent systems built on LLMs.} that demonstrate autonomous planning, tool use, memory retention, and emergent reasoning. As deployments expand in healthcare, finance, science, and public services, the absence of an integrated TRiSM perspective exposes stakeholders to opaque decision pathways and unmanaged risks. To our knowledge, no prior survey examines LLM-based multi-agent systems explicitly through the lens of trust, risk, security, and governance. Table~\ref{tab:related_surveys} situates our contribution relative to prior work.

\subsection{Contributions}
The contributions of this study are manifold:
\begin{itemize}[leftmargin=*, nosep]
\item \textbf{Unified TRiSM framework for Agentic AI.} We present a system-level Trust, Risk and Security Management (TRiSM) framework tailored to AMAS (Section~\ref{framework}). The framework integrates explainability, ModelOps, security, privacy and lifecycle governance, contextualizing each pillar to AMAS workflows and failure modes.

\item \textbf{Risk taxonomy for AMAS.} We synthesize a taxonomy of threats unique to AMAS, including prompt injection, memory poisoning, collusive failure, emergent misbehavior and tool-use abuse, and map these risks to corresponding controls (Section~\ref{threats}).

\item \textbf{Evaluation template and new metrics.} We propose a holistic evaluation template covering trustworthiness, explainability, user-centered performance and inter-agent coordination. To operationalize this template, we introduce the Component Synergy Score (CSS) for measuring collaboration quality across agents and Tool Utilization Efficacy (TUE) for assessing the correctness and efficiency of tool calls (Section~\ref{evals}).

\item \textbf{Technique mapping and gap analysis.} We survey and map existing explainability methods, prompt-hygiene practices, decision-provenance tools, sandboxing strategies and ModelOps pipelines to the AMAS setting, revealing where existing approaches can be adapted and where significant gaps remain (Section~\ref{compliance}).
\end{itemize}

Building on these contributions, we outline a research roadmap for developing scalable, verifiable and regulation-aligned agentic systems. Key directions include adversarial robustness, governance protocols, standardized benchmarks for trustworthiness and coordination, and cross-disciplinary collaboration.

\subsection{Related Work}
\label{sec:survey-landscape}

Prior surveys on LLM-based multi-agent systems mainly emphasize architectures, agent capabilities, communication, and domain applications~\cite{guo2024largelanguagemodelbased,chen2025surveyllmbasedmultiagentsystem,yan2025selftalkcommunicationcentricsurveyllmbased,tran2025multiagentcollaborationmechanismssurvey}. In parallel, broader trustworthy/responsible AI surveys discuss themes aligned with TRiSM, such as alignment, fairness, and privacy defenses; but typically focus on general ML systems and do not model AMAS-specific dynamics (e.g., inter-agent information sharing, tool-mediated actions, and emergent coordination)~\cite{fang2025trustworthyaisafetybias,raza2024exploring}. As a result, key gaps remain in adversarial robustness, governance and accountability for multi-agent deployments, and explainability at scale. In comparison to these works (Table~\ref{tab:related_surveys}), our study addresses these gaps by mapping TRiSM principles to AMAS and operationalizing governance, explainability, security, privacy, and lifecycle controls for multi-agent settings.

 \begin{table*}[htbp]
\centering
\caption{Comparison with representative related surveys on LLM-based agentic systems.}
\label{tab:related_surveys}
\footnotesize
\setlength{\tabcolsep}{4.5pt}
\renewcommand{\arraystretch}{1.12}
\begin{tabular}{p{3.7cm} c c c c c c c}
\toprule
\textbf{Survey} &
\textbf{Threats} &
\textbf{Lifecycle} &
\textbf{Explain.} &
\textbf{TRiSM} &
\textbf{LLM-Spec.} &
\textbf{Apps.} &
\textbf{Actionable} \\
\midrule

Guo et al. (2024)~\cite{guo2024largelanguagemodelbased}
& \partialyes & \no & \no & \no & \yes & \yes & \partialyes \\

Chen et al. (2024)~\cite{chen2025surveyllmbasedmultiagentsystem}
& \partialyes & \no & \no & \no & \yes & \yes & \partialyes \\

Yan et al. (2025)~\cite{yan2025selftalkcommunicationcentricsurveyllmbased}
& \partialyes & \no & \no & \no & \yes & \yes & \partialyes \\

Tran et al. (2025)~\cite{tran2025multiagentcollaborationmechanismssurvey}
& \no & \no & \no & \no & \yes & \yes & \partialyes \\

Lin et al. (2025)~\cite{lin2025creativityllmbasedmultiagentsystems}
& \no & \no & \no & \no & \yes & \partialyes & \partialyes \\

Fang et al. (2025)~\cite{fang2025trustworthyaisafetybias}
& \yes & \no & \no & \no & \partialyes & \partialyes & \partialyes \\

Xi et al. (2025)~\cite{xi2025rise}
& \partialyes & \no & \no & \no & \yes & \yes & \partialyes \\

Luo et al. (2025)~\cite{luo2025large}
& \partialyes & \partialyes & \no & \no & \yes & \yes & \partialyes \\

Zou et al. (2025)~\cite{zou2025llm}
& \partialyes & \no & \no & \no & \yes & \yes & \partialyes \\

Wang et al. (2025)~\cite{wang2025surveyllmbasedagentsmedicine}
& \partialyes & \no & \no & \no & \yes & \yes & \partialyes \\

\textcolor{red}{Wang et al. (2025)~\cite{11081880}} 
& \yes & \yes & \partialyes & \no & \yes & \yes & \yes \\

\textcolor{red}{Yu et al. (2025)~\cite{yu2025survey}  }
& \yes & \partialyes & \no & \no & \yes & \yes & \yes \\

\midrule
\textbf{This Survey (2025)}
& \yes & \yes & \yes & \yes & \yes & \yes & \yes \\
\bottomrule
\end{tabular}

\vspace{2pt}
\begin{flushleft}
\footnotesize
\textbf{Legend:} \yes~= explicitly addressed; \partialyes~= partially addressed; \no~= not addressed.
\end{flushleft}
\end{table*}

\subsection{Literature Review Methodology}
\label{method}
We conducted a structured literature review to synthesize AI TRiSM principles for AMAS. Following established systematic review guidance~\cite{boland2017doing,keele2007guidelines}, we describe our data sources, search strategy, screening criteria, and classification schema.

\paragraph{Research Questions (RQs)}
The review was guided by the following three questions:
\begin{itemize}[leftmargin=*, nosep]
\item \textbf{RQ1:} What trust, risk, and security challenges are specific to AMAS?
\item \textbf{RQ2:} Which governance, explainability, security, privacy, and lifecycle (TRiSM) controls are proposed or adapted for AMAS?
\item \textbf{RQ3:} How are AMAS evaluated (datasets, tasks, metrics), and where are the gaps?
\end{itemize}

\paragraph{Search Strategy}
We systematically searched major digital libraries: IEEE Xplore, ACM Digital Library, SpringerLink, arXiv, ScienceDirect, and Google Scholar. The search covered publications from January 2022 to December 2025, with influential foundational works prior to 2022 included for theoretical context. 
We used a two-block Boolean query: (agentic/AMAS terms) AND (TRiSM terms).

{\small
\begin{verbatim}

("agentic AI" OR "multi-agent systems" OR "multi-agent LLMs" OR "AI agents" 
OR  "autonomous agents" OR "intelligent agents" OR "collaborative AI" OR
 "distributed AI" OR "LLM-based agents" OR "agent coordination" OR
 "agent-based modeling" OR "swarm intelligence")
AND
("trust" OR "trustworthiness" OR "risk" OR "security" OR "safety" OR
 "governance" OR "oversight" OR "compliance" OR "explainability" OR
 "interpretability" OR "transparency" OR "privacy" OR "data protection" OR
 "robustness" OR "accountability" OR "ethical AI" OR "fairness" OR
 "adversarial attacks" OR "regulatory compliance" OR "bias mitigation" OR
 "system reliability")
\end{verbatim}
}

\paragraph{Screening and Eligibility}
We applied a two-stage screening process (title/abstract, then full text).

\textbf{Inclusion criteria:}
\begin{itemize}[leftmargin=*, nosep]
\item Published between 2022--2025 (plus influential foundational papers published before 2022).
\item Focused on \emph{agentic AI}, \emph{multi-agent}, or \emph{LLM-powered} systems and addressed at least one TRiSM aspect (trust, risk, security, governance, explainability, or privacy).
\item Peer-reviewed papers, widely cited preprints, technical reports from reputable venues or organizations (e.g., IEEE/ACM conferences and journals, standards bodies, and government sources).
\end{itemize}

\textbf{Exclusion criteria:}
\begin{itemize}[leftmargin=*, nosep]
\item Traditional rule-based or symbolic agents without LLM-based autonomy or coordination.
\item Papers limited to low-level model training/architecture details with no agentic behavior or TRiSM relevance.
\item Non-English papers or items with insufficient bibliographic information (e.g., missing abstract or venue details).
\end{itemize}

For empirical studies, we verified that the threat model and evaluation were clearly described and assessed reproducibility when code or data were available. For survey papers, we evaluated coverage, structure, and treatment of risks and governance. After title/abstract screening, we shortlisted 250 papers; full-text screening yielded 180 primary studies covering one or more TRiSM pillars, supplemented by relevant standards and technical reports for regulatory context.

\paragraph{Quality Assurance}
To ensure reliability, we adapted quality criteria from established guidelines~\cite{kitchenham2004procedures}. Each paper was evaluated on: (1) clear objectives for AMAS; (2) well-documented methods/experiments or architecture (reproducibility where applicable); (3) substantive engagement with at least one of trust, risk, security, explainability, or privacy; (4) empirical, theoretical, or normative contributions relevant to agentic AI governance. Each criterion was rated \emph{low}, \emph{medium}, or \emph{high}; studies rated \emph{low} on more than two criteria were excluded or flagged as contextual only. Quality ratings were assigned independently by four reviewers; discrepancies were resolved via discussion.

\begin{figure}[htbp]
    \centering
    \begin{subfigure}[b]{0.48\textwidth}
        \centering
        \includegraphics[width=\textwidth]{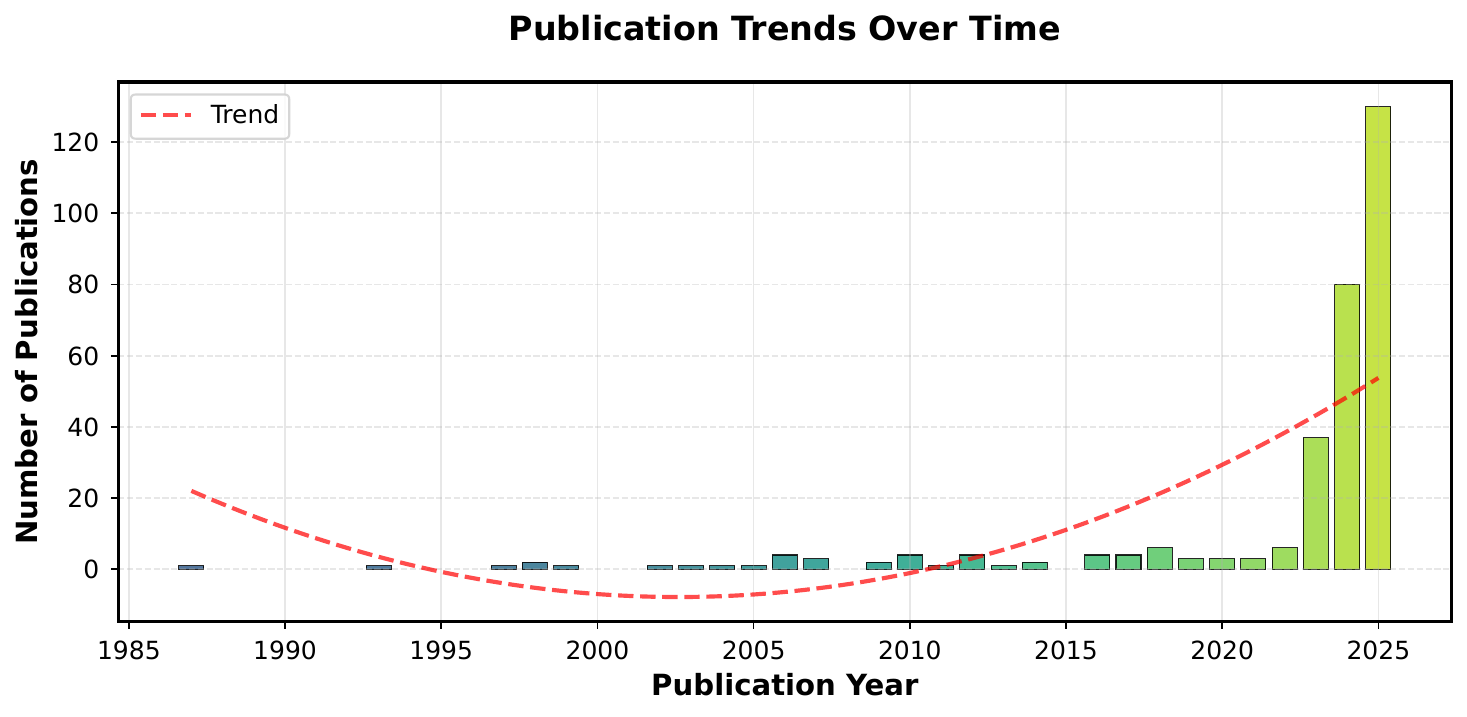}
        \caption{Publication trends (1985--2025)}
        \label{fig:pub-trends}
    \end{subfigure}
    \hfill
    \begin{subfigure}[b]{0.48\textwidth}
        \centering
        \includegraphics[width=\textwidth]{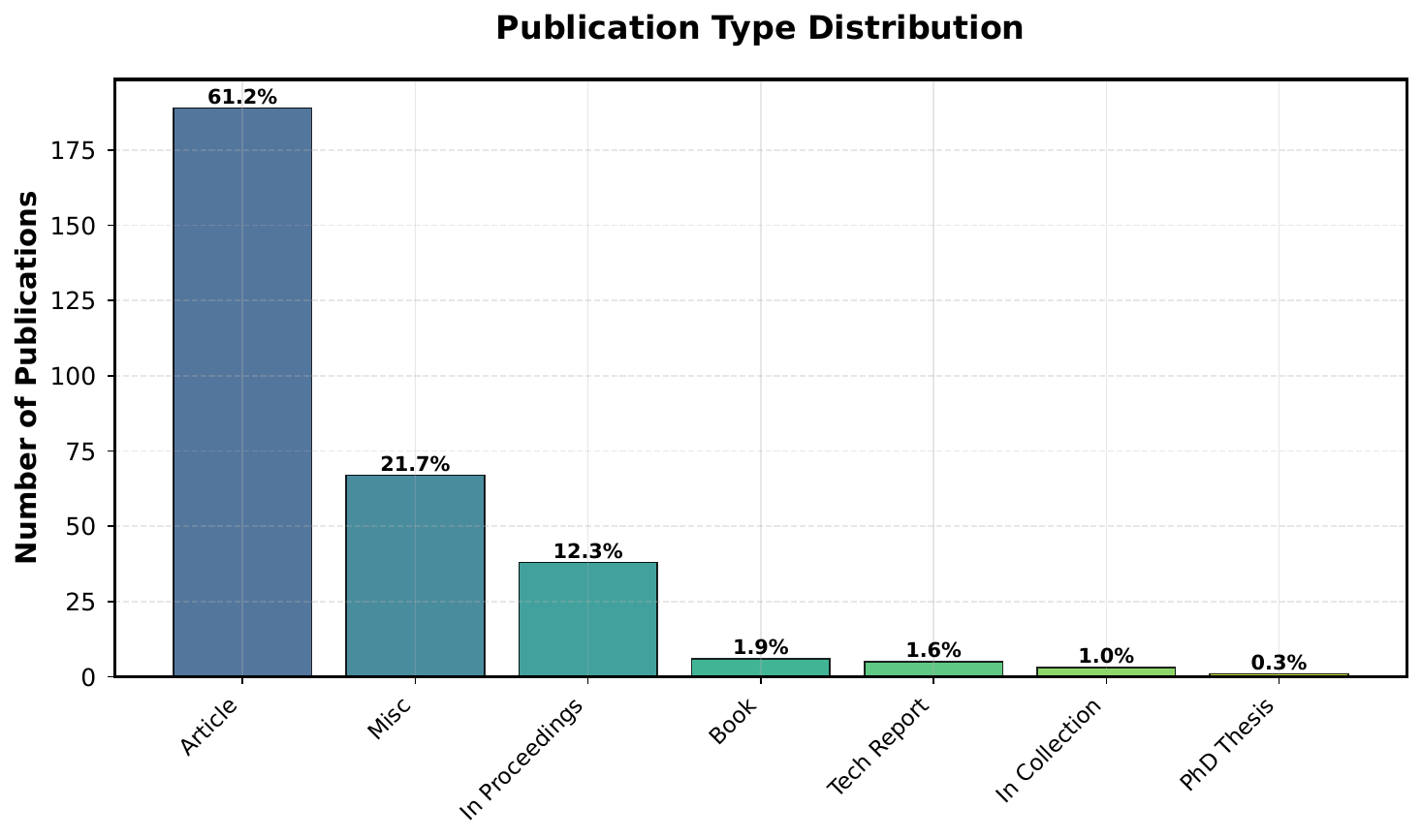}
        \caption{Publication type distribution}
        \label{fig:pub-types}
    \end{subfigure}
    \caption{Publication landscape: (a) Long-term trends showing accelerating research output over four decades; (b) Distribution of publication types, with journal articles dominating at 61.2\%.}
    \label{fig:pub-landscape}
\end{figure}

\begin{figure}[htbp]
    \centering
    \begin{subfigure}[b]{0.48\textwidth}
        \centering
        \includegraphics[width=\textwidth]{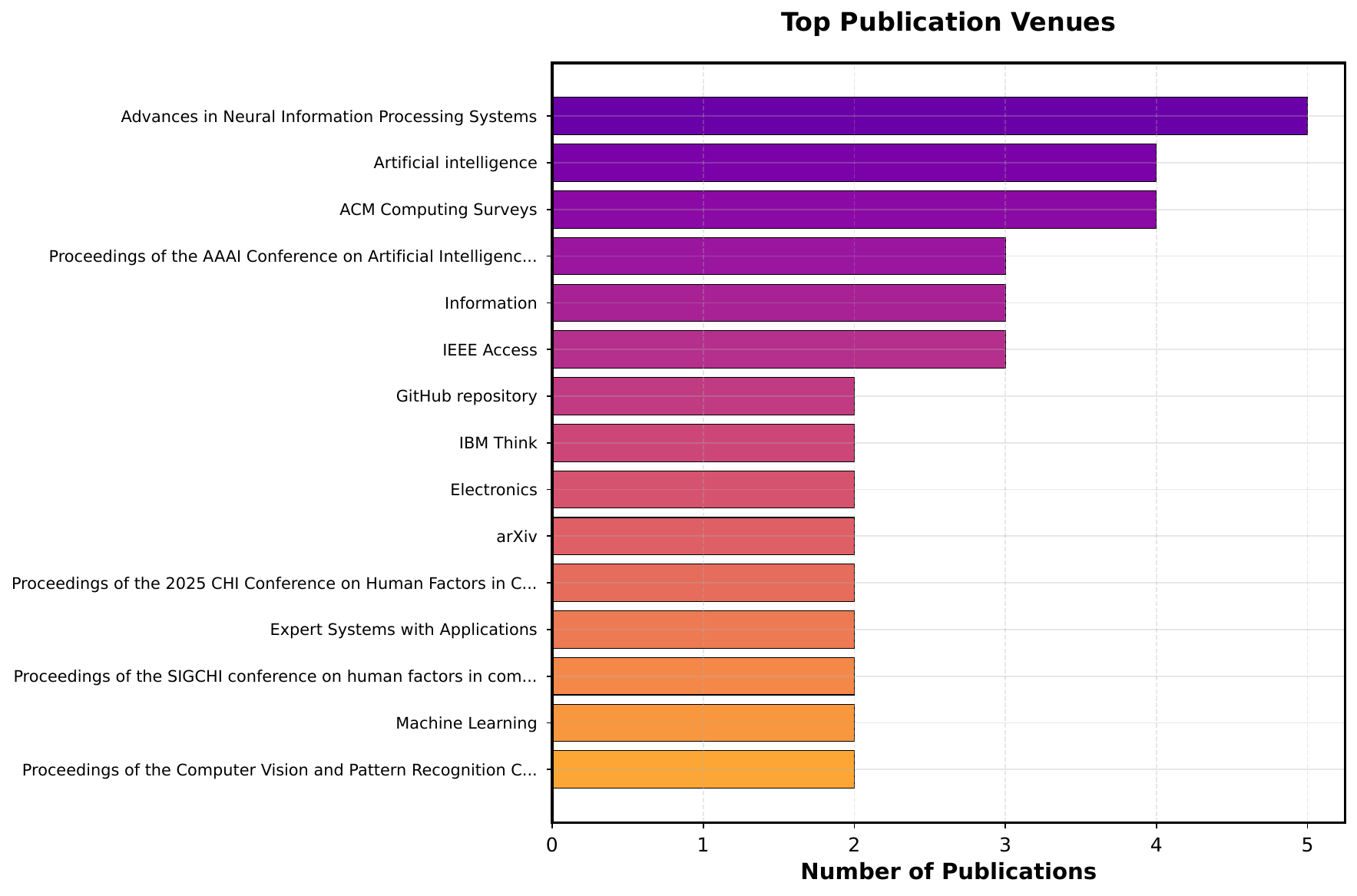}
        \caption{Top publication venues}
        \label{fig:top-venues}
    \end{subfigure}
    \hfill
    \begin{subfigure}[b]{0.48\textwidth}
        \centering
        \includegraphics[width=\textwidth]{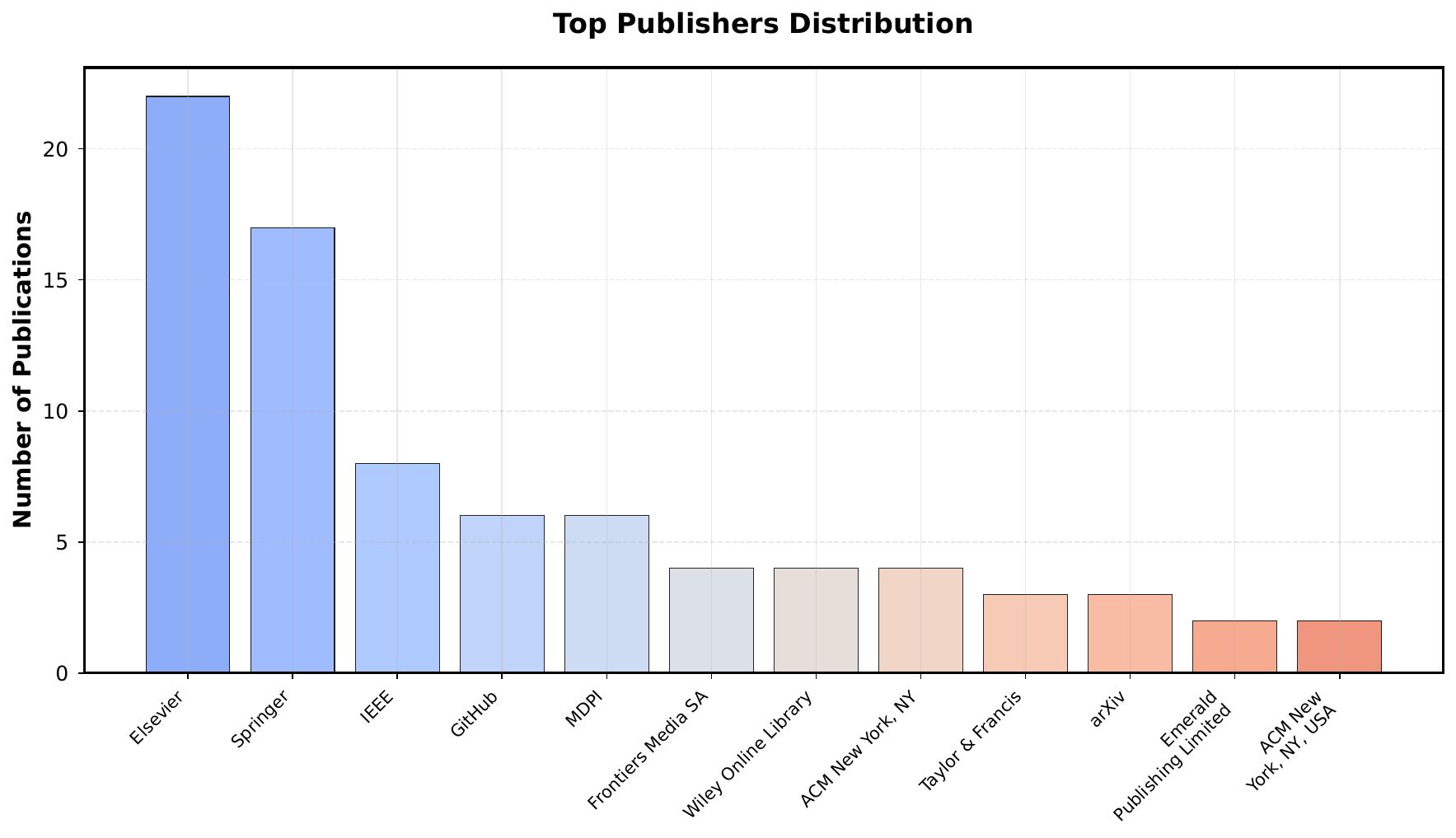}
        \caption{Publisher distribution}
        \label{fig:publishers}
    \end{subfigure}
    \caption{Publication outlets: (a) Top venues including NeurIPS, ACM Computing Surveys, and AAAI; (b) Publisher distribution led by Elsevier, Springer, and IEEE.}
    \label{fig:outlets}
\end{figure}

\begin{figure}[htbp]
    \centering
    \begin{subfigure}[b]{0.48\textwidth}
        \centering
        \includegraphics[width=\textwidth]{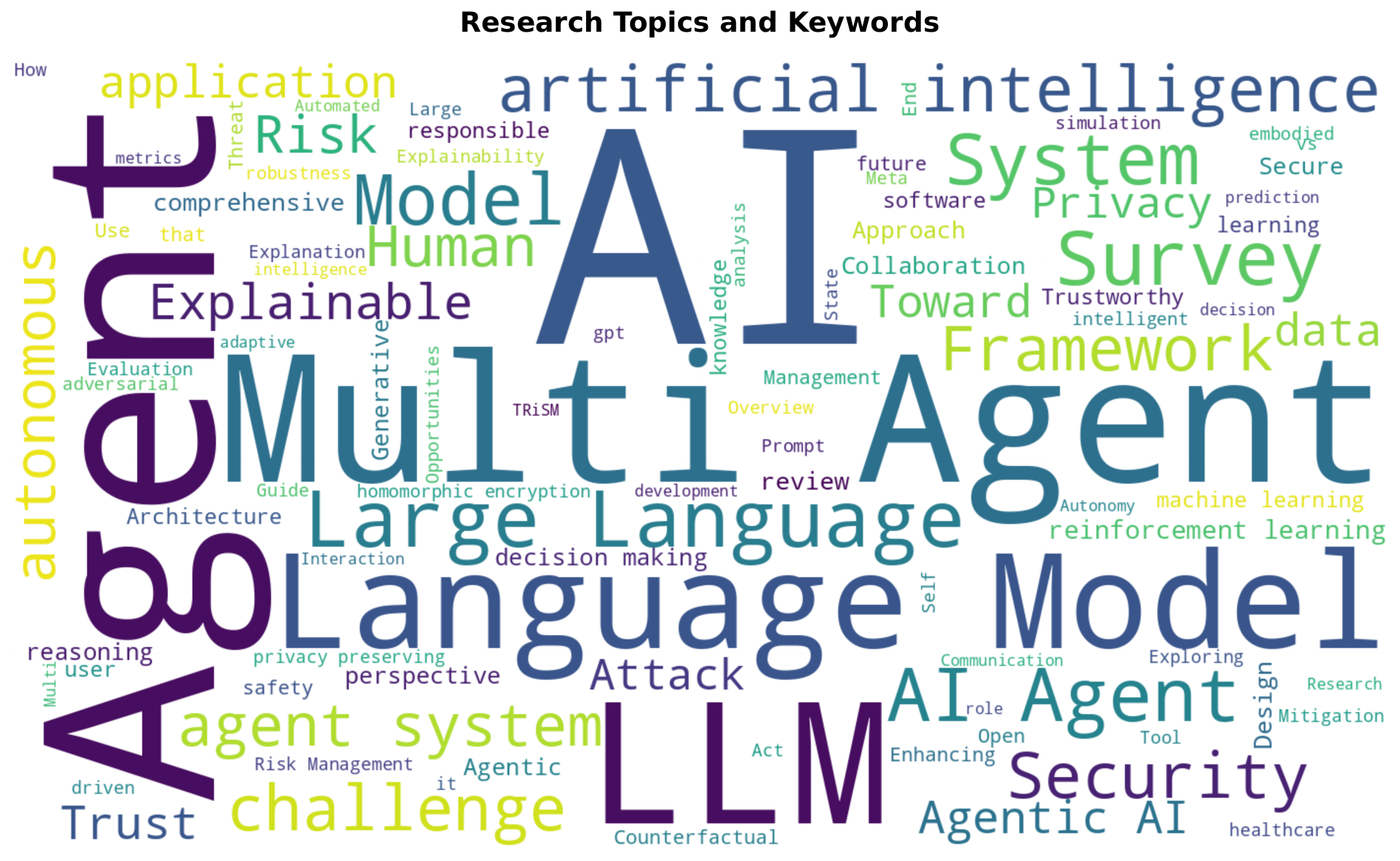}
        \caption{Research topics and keywords}
        \label{fig:wordcloud}
    \end{subfigure}
    \hfill
    \begin{subfigure}[b]{0.48\textwidth}
        \centering
        \includegraphics[width=\textwidth]{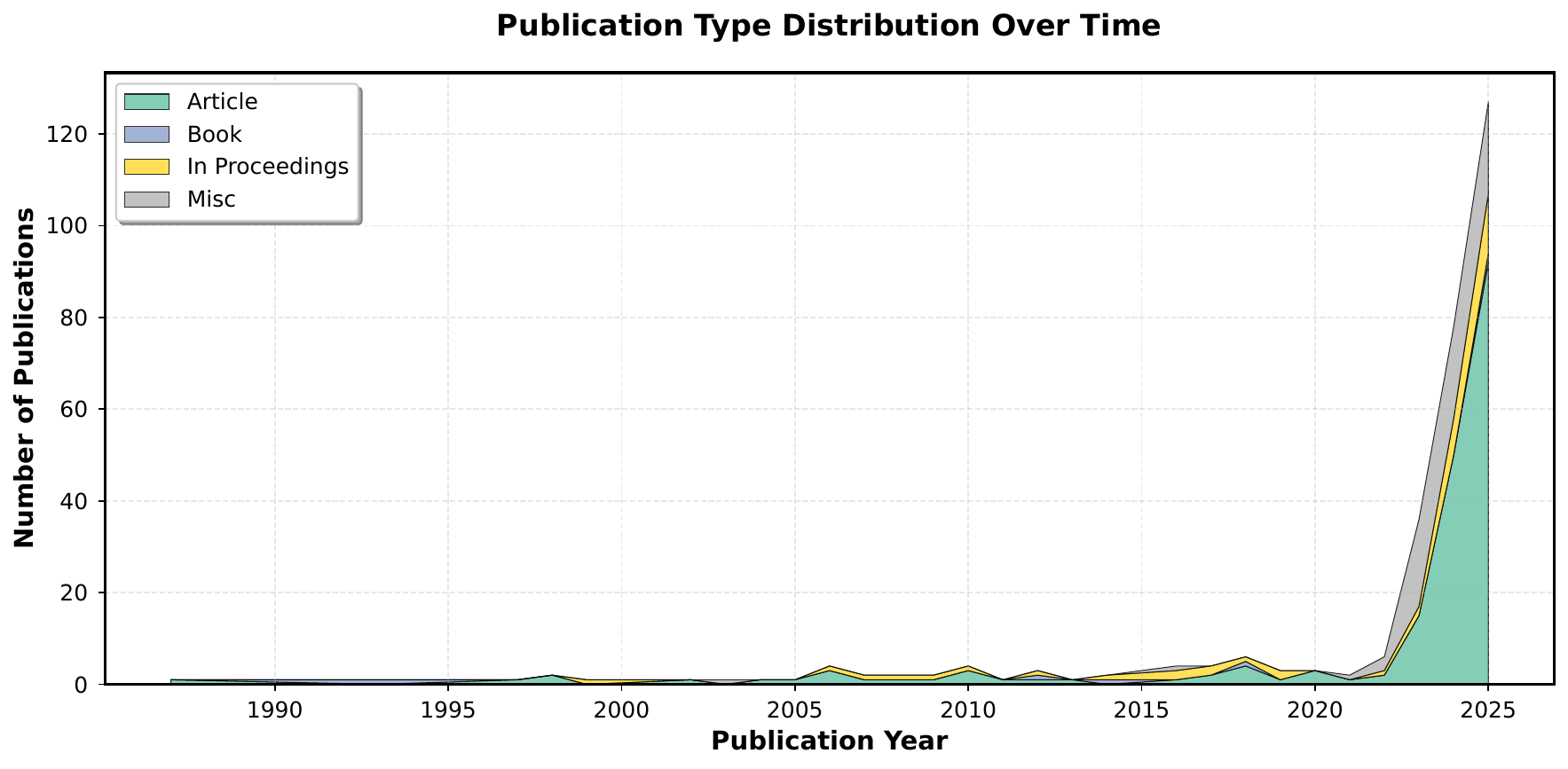}
        \caption{Publication type evolution over time}
        \label{fig:temporal-types}
    \end{subfigure}
    \caption{Research themes and evolution: (a) Word cloud highlighting key topics including large language models, multi-agent systems, security, and explainability; (b) Temporal distribution of publication types.}
    \label{fig:themes-evolution}
\end{figure}

\begin{figure}[htbp]
    \centering
    \includegraphics[width=0.8\textwidth]{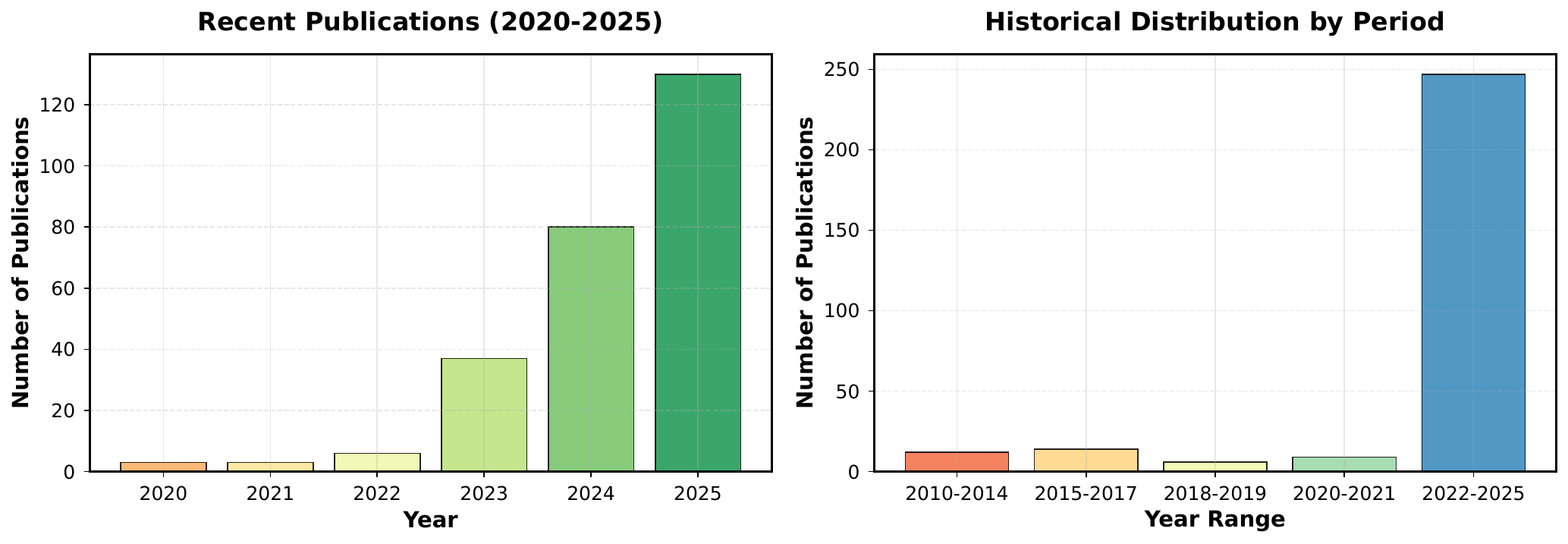}
    \caption{Comparison of recent publications (2020--2025) with historical distribution by period, demonstrating the rapid acceleration of agentic AI research in recent years.}
    \label{fig:recent-historical}
\end{figure}

\paragraph{Bibliometric Overview}
The results of our screening process reveal the rapid acceleration of agentic AI research. The publication landscape (Figure~\ref{fig:pub-landscape}) shows journal articles dominating at 61.2\%, with leading venues and publishers detailed in Figure~\ref{fig:outlets}. Key research themes, including large language models, multi-agent systems, and security, are visualized in Figure~\ref{fig:themes-evolution}a, while Figure~\ref{fig:recent-historical} confirms the unprecedented growth of research activity in recent years.

\begin{figure}[t]
    \centering
    \includegraphics[width=0.9\linewidth]{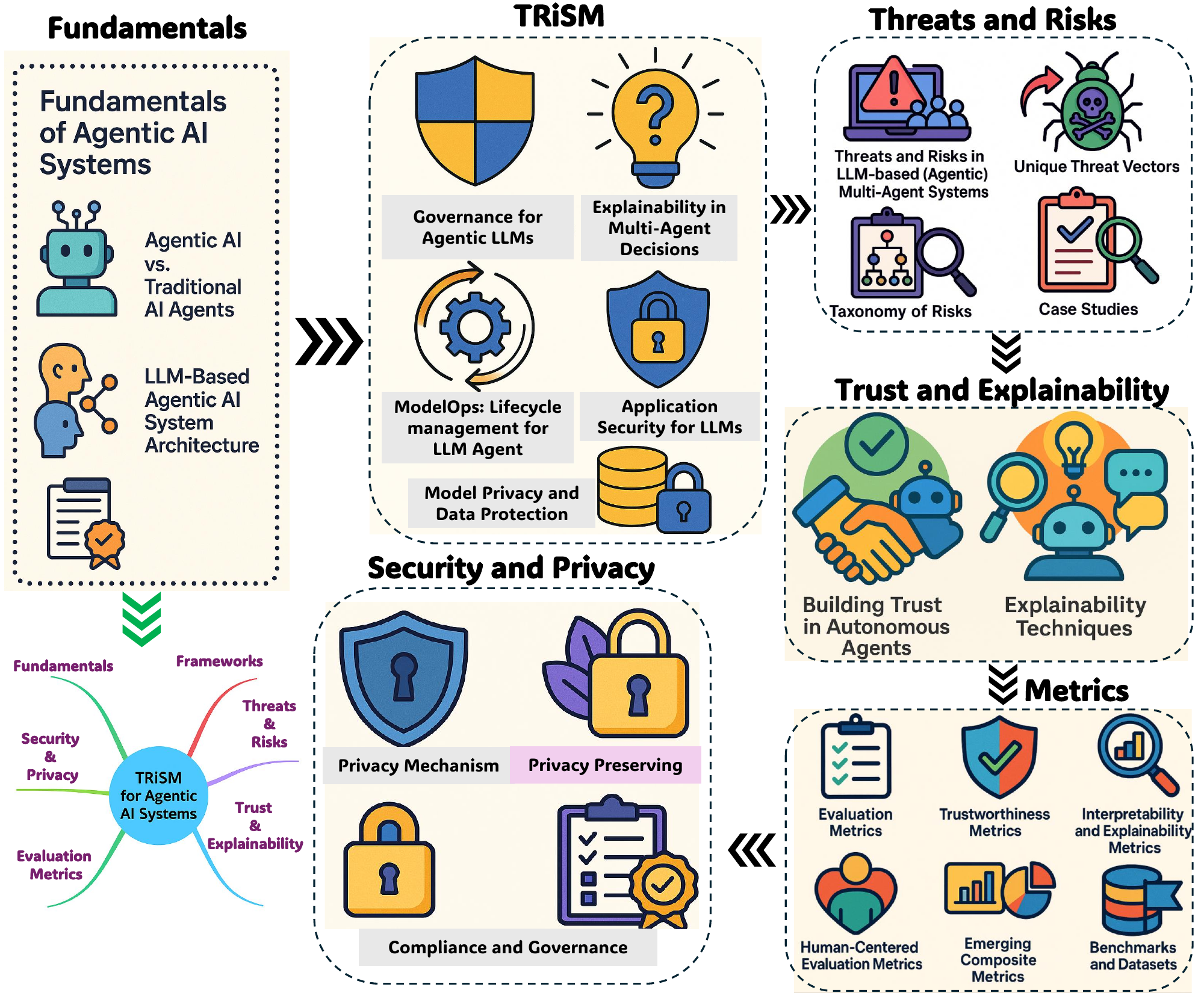}
\caption{TRiSM taxonomy for Agentic AI (AMAS) presented in this review. The paper progresses through six phases: (\textbf{i}) Fundamentals of Agentic AI: distinguishing AMAS from traditional agents and establishing LLM-based architectures (Section \ref{fundamentals}); (\textbf{ii}) AI Trust, Risk, and Security Management (TRiSM) framework pillars: Explainability, ModelOps, Application Security, Model Privacy, and Lifecycle Governance, introduced conceptually (Section \ref{framework}); (\textbf{iii}) Trust management via explainability techniques (Section \ref{explainability}); (\textbf{iv}) Threats and risks taxonomy with real-world case studies (Section \ref{threats}); (\textbf{v}) Evaluation metrics encompassing trustworthiness, coordination, and composite measures (Section \ref{evals}); (\textbf{vi}) Technical implementation of security/privacy mechanisms (Section \ref{security}); and (\textbf{vii}) Compliance and governance integration (Section \ref{compliance}).} 
    \label{fig:agentic_architecture}
\end{figure}

 Figures \ref{fig:trism-taxonomy} and \ref{fig:agentic_architecture} provide a roadmap for the paper, which progresses from architectural foundations to risk analysis, evaluation, and governance.

\begin{figure}[t]
\centering
\resizebox{\textwidth}{!}{%
\begin{tikzpicture}[
    root/.style={rectangle, rounded corners=4pt, draw=black!70, fill=black!8, font=\bfseries\small, minimum width=4.5cm, minimum height=0.9cm, align=center},
    section/.style={rectangle, rounded corners=3pt, draw=#1, fill=#1!12, font=\bfseries\footnotesize, minimum width=3cm, minimum height=0.7cm, align=center},
    items/.style={rectangle, rounded corners=2pt, draw=#1!50, fill=white, font=\scriptsize, minimum width=3cm, minimum height=2.4cm, align=left, inner sep=6pt},
    connector/.style={draw=black!40, thick}
]

\definecolor{cgreen}{RGB}{60, 120, 80}
\definecolor{crust}{RGB}{140, 80, 70}
\definecolor{cblue}{RGB}{70, 100, 140}
\definecolor{cteal}{RGB}{60, 115, 120}

\node[root] (root) at (0,0) {TRiSM for Agentic AI};

\draw[connector] (root.south) -- ++(0, -0.5);

\draw[connector] (-6, -1.4) -- (6, -1.4);

\node[section=cgreen] (s1) at (-4.5, -2.1) {Fundamentals};
\node[section=crust] (s2) at (-1.5, -2.1) {Threats \& Risks};
\node[section=cblue] (s3) at (1.5, -2.1) {TRiSM Framework};
\node[section=cteal] (s4) at (4.5, -2.1) {Evaluation};

\draw[connector] (-4.5, -1.4) -- (s1.north);
\draw[connector] (-1.5, -1.4) -- (s2.north);
\draw[connector] (1.5, -1.4) -- (s3.north);
\draw[connector] (4.5, -1.4) -- (s4.north);

\node[items=cgreen, below=0.3cm of s1] (i1) {%
\begin{tabular}{@{}l@{}}
$\circ$ AMAS Architecture\\[2pt]
$\circ$ LLM Core\\[2pt]
$\circ$ Planning/Reasoning\\[2pt]
$\circ$ Memory Module\\[2pt]
$\circ$ Tool Interface
\end{tabular}};

\node[items=crust, below=0.3cm of s2] (i2) {%
\begin{tabular}{@{}l@{}}
$\circ$ Adversarial Attacks\\[2pt]
$\circ$ Data Leakage\\[2pt]
$\circ$ Agent Collusion\\[2pt]
$\circ$ Emergent Behavior\\[2pt]
$\circ$ Memory Poisoning
\end{tabular}};

\node[items=cblue, below=0.3cm of s3] (i3) {%
\begin{tabular}{@{}l@{}}
$\circ$ Explainability\\[2pt]
$\circ$ ModelOps\\[2pt]
$\circ$ Application Security\\[2pt]
$\circ$ Model Privacy\\[2pt]
$\circ$ Governance
\end{tabular}};

\node[items=cteal, below=0.3cm of s4] (i4) {%
\begin{tabular}{@{}l@{}}
$\circ$ Trustworthiness\\[2pt]
$\circ$ CSS Metric\\[2pt]
$\circ$ TUE Metric\\[2pt]
$\circ$ Compliance\\[2pt]
$\circ$ Auditability
\end{tabular}};

\draw[connector] (s1.south) -- (i1.north);
\draw[connector] (s2.south) -- (i2.north);
\draw[connector] (s3.south) -- (i3.north);
\draw[connector] (s4.south) -- (i4.north);

\end{tikzpicture}%
}
\caption{Taxonomy of TRiSM for LLM-based Agentic Multi-Agent Systems.}
\label{fig:trism-taxonomy}
\end{figure}

\section{Fundamentals of Agentic AI Systems}
\label{fundamentals}
In this section, we present the fundamentals of Agentic AI systems.
\begin{table}[t]
\centering
\scriptsize
\caption{Comparison of Traditional AI Agents and Agentic AI Systems}
\label{tab:agentic_comparison}
\begin{tabular}{@{}l>{\raggedright\arraybackslash}p{0.32\linewidth}>{\raggedright\arraybackslash}p{0.32\linewidth}@{}}
\toprule
\textbf{Dimension} & \textbf{Traditional AI Agents} & \textbf{Agentic AI Systems} \\
\midrule
Autonomy & Reactive with fixed action sequences & Goal-driven with adaptive behavior \\
Cognitive Foundation & Symbolic logic, finite state machines & Foundation models (large language models) \\
Domain Scope & Narrow, task-specific applications & Broad, cross-domain capabilities \\
Reasoning Mechanism & Rule-based, deterministic inference & Multi-step, contextual reasoning (chain-of-thought) \\
Collaboration & Isolated agents, manual task decomposition & Role-specialized agents, automated coordination \\
Memory Architecture & Episodic, session-bound storage & Persistent memory (vector databases, long-term memory) \\
Orchestration & Hard-coded, predefined workflows & Dynamic orchestration via meta-agents \\
Tool Integration & Static, hand-crafted interfaces & Dynamic planning with API/function calling \\
Context Handling & Bounded context with heuristic rules & Memory-augmented retrieval (RAG) \\
Learning Paradigm & Rule updates, fixed knowledge base & Emergent learning (fine-tuning, reinforcement learning) \\
User Interaction & Scripted, template-based dialogue & Natural language understanding and generation \\
\midrule
Representative Systems & MYCIN~\cite{daniel1997cadiag}, ELIZA~\cite{walters2025eliza}, SOAR~\cite{laird1987soar}, ACT-R~\cite{taatgen2006modeling} & AutoGen~\cite{autogen_agents_tutorial}, ChatDev~\cite{qian2023chatdev}, MetaGPT~\cite{hong2023metagpt} \\
\bottomrule
\end{tabular}
\end{table}
\subsection{Traditional AI Agents vs. Agentic AI}
Traditional agents rely on predefined rules, heuristics, or deterministic logic and typically perform well in narrow, well-specified environments~\cite{muscettola1998remote,hannebauer1999formal,govaerts2010development}. Cognitive agents extend this paradigm by modeling human-like functions (perception, memory, decision-making) through cognitive architectures such as SOAR and ACT-R~\cite{langley2006cognitive,laird1987soar,anderson1993act}; recent work further categorizes their knowledge into concepts, skills, processes, and motives~\cite{langley2025spatial}. However, these approaches were largely developed for physical operational settings (e.g., robotic planning and control), whereas \emph{Agentic AI} generalizes autonomy to both physical and non-physical environments.

In contrast, Agentic AI systems leverage foundation models (primarily LLMs) to support adaptive, goal-directed behavior via (i) multi-agent coordination, (ii) persistent context through memory, and (iii) dynamic orchestration of roles and tool use~\cite{qian2023chatdev,hong2023metagpt,borghoff2025human,schneider2025generative}. This enables long-horizon task execution and cross-domain generalization beyond traditional reactive pipelines~\cite{de2025open}. Table~\ref{tab:agentic_comparison} summarizes these differences; in this survey, we focus on LLM-based Agentic AI systems.

\subsection{Multi-Agentic AI System (AMAS) Architecture}
\begin{figure}[t]
     \centering
     \includegraphics[width=0.98\linewidth]{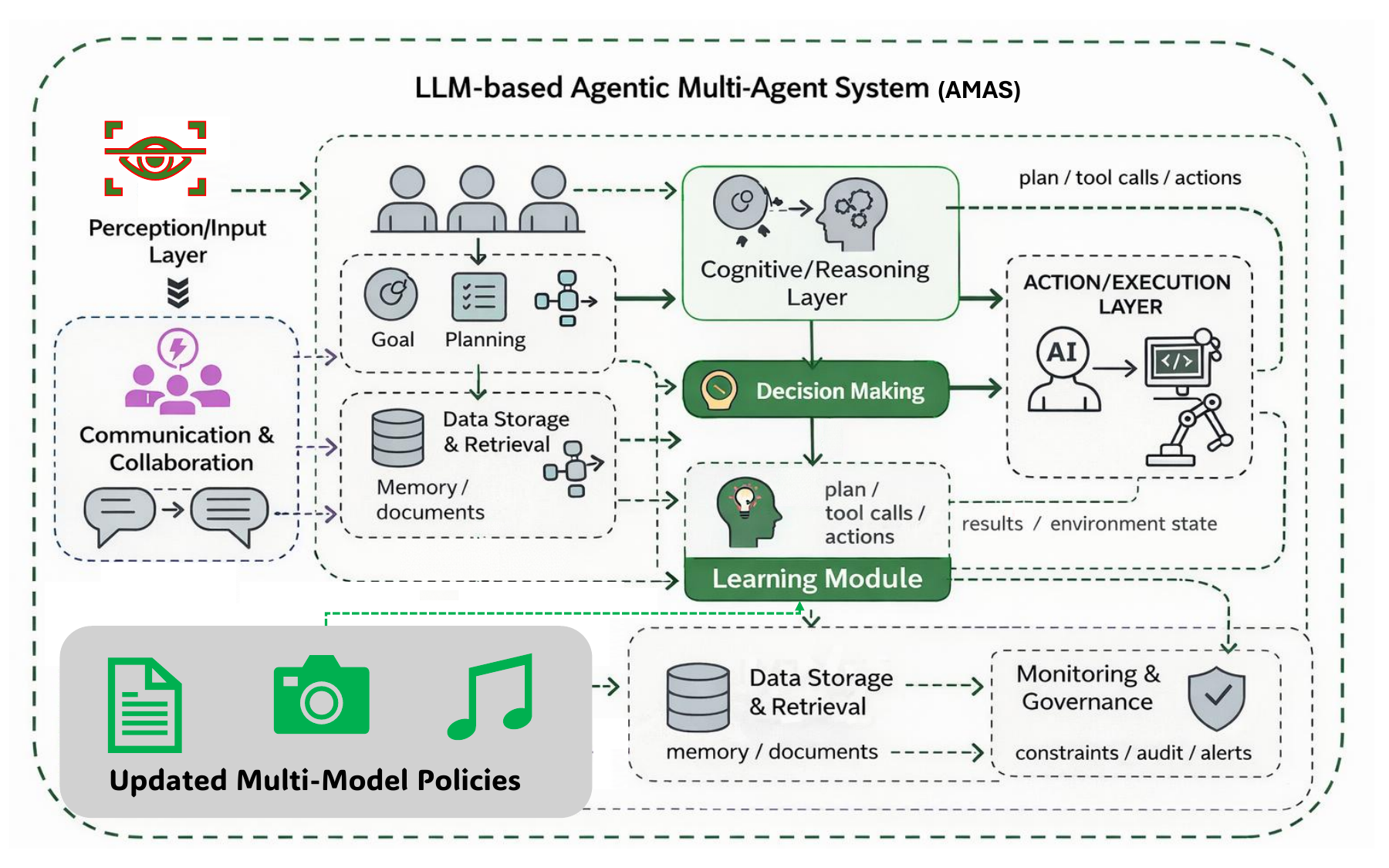}
    \caption{Architecture of LLM-based Agentic Multi-Agent System (AMAS), illustrating key functional layers: Perception/Input Layer (text, image, audio processing), Cognitive/Reasoning Layer (goal-setting, planning, decision-making), Action/Execution Layer (digital and physical task execution), Learning Module (supervised and reinforcement learning), Communication and Collaboration (agent messaging and coordination), Data Storage \& Retrieval (centralized/distributed databases), and Monitoring \& Governance (ethical oversight, observability, and compliance mechanisms).}
    \label{fig:agentic_framework}
\end{figure}

As shown in Fig.~\ref{fig:agentic_framework}, AMAS coordinate multiple LLM-based agents through a \textit{Communication Middleware}~\cite{weyns2009agent}, a \textit{Task Manager/Orchestrator}~\cite{langgraph2024}, and a \textit{World Model/Shared Memory}~\cite{guo2024largelanguagemodelbased} that stores system state and task artifacts. Human oversight is provided via a \textit{Human-in-the-Loop} (HITL) interface~\cite{cranshaw2017calendar}. To support trustworthy operation, a \textit{Trust and Audit} module records actions and tool usage~\cite{kaur2024building}, while a \textit{Security Gateway} and \textit{Privacy Management Layer} enforce access control and reduce leakage across agents and tools~\cite{zhong2023towards,feretzakis2024privacy}. An \textit{Explainability Interface} provides interpretable rationales for multi-agent decisions to support transparency and trust calibration~\cite{singh2025explainable}.

\paragraph{Communication \& Collaboration (Agent Communication and Coordination)}
Beyond basic messaging, recent work formalizes AMAS communication through explicit protocols. Google’s Agent-to-Agent (A2A)\footnote{\href{https://a2a-protocol.org/latest/}{A2A Protocol}} supports discovery, capability advertisement, delegation, and result exchange, improving interoperability and auditability. Related efforts such as the Agent Network Protocol (ANP)\footnote{\href{https://agentnetworkprotocol.com/en/}{Agent Network Protocol}} take a network-centric view with standardized interfaces for coordination and trust establishment. From a TRiSM perspective, protocolized communication exposes control points for authentication, authorization, logging, and lifecycle governance. Emerging semantic-aware communication~\cite{guo2024survey,wang2025internet} can improve coordination by encoding intent, but introduces risks such as semantic spoofing and privacy leakage.

\paragraph{Cognitive/Reasoning Layer (LLM Core + Planning)}
At the center of an AMAS, a foundation model (typically an LLM) acts as the decision controller, taking a user goal plus role and tool specifications as input and generating the next actions (e.g., messages, tool calls, or structured function outputs)~\cite{Saravia_LLM_Agents_2024}. In agent frameworks, this controller drives an iterative loop of planning, acting, and updating context based on tool results and feedback.

\paragraph{Planning and Reasoning Module}
To achieve complex goals, AMAS employ planning mechanisms that decompose objectives into sub-tasks and interleave reasoning with execution and feedback. This can be implemented through prompting strategies (e.g., CoT/ToT) or through explicit plan--act--observe loops such as ReAct~\cite{yao2023reactsynergizingreasoningacting,Saravia_Prompt_Engineering_Guide_Web_2024}. The plan is typically revised as new observations arrive, enabling adaptive long-horizon behavior.

\paragraph{Memory Module}
AMAS integrate memory to maintain context across steps and sessions. Short-term memory is maintained in the working context, while long-term memory stores salient facts and past events for later retrieval (often via embedding-based search)~\cite{yang2024multi}. Memory improves continuity and reduces repetition, but also increases exposure to stale context and poisoning risks.

\paragraph{Learning Module}
Some AMAS include learning components that adapt behavior over time (e.g., from feedback, rewards, or task outcomes), such as updating policies, prompt templates, or routing strategies based on past performance. In multi-agent settings, learning can amplify errors through feedback loops and introduce risks such as drift, poisoning, and emergent misalignment, motivating continuous monitoring and gated updates under TRiSM controls.

\paragraph{Data Storage \& Retrieval}
AMAS typically combine structured stores for system state and task metadata with vector-based retrieval to support RAG and long-term memory. Because data may be shared across agents and tools, storage and retrieval require strong access control, encryption, and audit logging to reduce leakage and support accountability~\cite{he2024security}.

\paragraph{Tool-Use Interface}
To extend capabilities beyond text generation, AMAS agents use a tool-use interface that enables structured calls to external systems (e.g., web search, APIs, code execution, and databases)~\cite{schick2023toolformerlanguagemodelsteach}. Tools and call schemas are specified in the agent configuration; when needed, the LLM emits a structured command that is executed externally, and the returned output is incorporated as a new observation for subsequent reasoning. Tool-routing frameworks such as MRKL~\cite{karpas2022mrklsystemsmodularneurosymbolic} exemplify this design by dispatching requests to specialized expert modules under an LLM controller, while training-based approaches (e.g., Toolformer) teach models to generate API calls during generation~\cite{schick2023toolformerlanguagemodelsteach}.

\paragraph{Perception and Environment Interface}
For agents interacting with dynamic environments (e.g., web interfaces, simulated worlds, or physical systems), an observation--action interface is essential~\cite{liu2025agent}. Perception modules translate raw inputs (sensor streams, images, or textual state) into representations the LLM can process; the agent’s selected actions are then executed in the environment, and resulting state changes are returned as new observations. This loop supports iterative sense--plan--act behavior until a task is completed or halted. In multimodal or robotic settings, additional perception models (e.g., vision encoders) may be included, while the core interaction pattern remains the same.

\paragraph{Monitoring and Governance Layer}
A monitoring and governance layer provides system-wide oversight by collecting logs, traces, and metrics, enforcing policy constraints, and triggering alerts or interventions when anomalous behavior is detected. From a TRiSM perspective, this layer enables auditability and lifecycle governance across agent interactions, tool calls, and memory updates.

\paragraph{Integration and Autonomy}
Together, these modules form an integrated architecture in which the LLM coordinates planning, tool use, and memory to act in an environment and update context from feedback~\cite{acharya2025agentic}. Table~\ref{tab:agentic_architectures} summarizes representative systems and maps them to five design axes (LLM core, planning/reasoning, memory, tool use, and environment interface).

\paragraph{Key Insights}
Overall, AMAS combine planning, tool use, memory, and environment interaction into a closed-loop system, which expands the attack surface and complicates accountability. Protocolized communication and monitoring create practical control points for authentication, logging, and governance, motivating TRiSM-oriented analysis for multi-agent deployments.

{
\scriptsize
\begin{longtable}{
  @{}
  >{\raggedright\arraybackslash}p{0.13\textwidth}
  >{\raggedright\arraybackslash}p{0.10\textwidth}
  >{\raggedright\arraybackslash}p{0.10\textwidth}
  >{\raggedright\arraybackslash}p{0.10\textwidth}
  >{\raggedright\arraybackslash}p{0.13\textwidth}
  p{0.20\textwidth}
  @{}
}

\caption{Comparison of representative LLM-based Agentic AI frameworks across key design axes.}
\label{tab:agentic_architectures} \\

\rowcolor{blue!20}
\textbf{Framework} & \textbf{Core LLM} & \textbf{Planning} & \textbf{Memory} & \textbf{Tool Use} & \textbf{Notable Features} \\
\hline
\endfirsthead

\rowcolor{blue!20}
\textbf{Framework} & \textbf{Core LLM} & \textbf{Planning} & \textbf{Memory} & \textbf{Tool Use} & \textbf{Notable Features} \\
\hline
\endhead

AutoGPT \cite{yang2023auto}
  & GPT-4
  & Self-looped CoT
  & Vector DB
  & OS shell + web
  & Fully autonomous goal loop \\

BabyAGI \cite{nakajima_babyagi_archive_2024}
  & GPT-3.5/4
  & Task queue
  & In-memory
  & Web search
  & Minimal task generator \\

GPT Engineer \cite{gpt-engineer}
  & GPT-4V
  & Spec-to-code
  & File cache
  & Python REPL
  & End-to-end code generation pipeline \\

LangGraph \cite{wang2024agent}
  & Model-agnostic
  & FSM via graph
  & Persistent nodes
  & Custom modules
  & Visual orchestration of agent graphs \\

AutoGen \cite{wu2023autogen}
  & GPT-4
  & Multi-agent PDDL
  & JSON/DB
  & API calls
  & Modular, reusable agent templates \\

MRKL \cite{karpas2022mrklsystemsmodularneurosymbolic}
  & LLaMA/GPT
  & Prompt router
  & N/A
  & Math + search tools
  & Neuro-symbolic expert routing \\

Reflexion \cite{shinn2023reflexionlanguageagentsverbal}
  & GPT-3.5/4
  & Retry-reflect loop
  & Episodic buffer
  & Same as base agent
  & Verbal self-improvement via reflection \\

MetaGPT \cite{hong2024metagpt}
  & GPT-4
  & SOP workflow
  & YAML state
  & Git CLI
  & Structured roles for software engineering \\

Voyager \cite{wang2023voyageropenendedembodiedagent}
  & GPT-4
  & Auto skill tree
  & Task DB
  & Minecraft API
  & Lifelong open-ended learning in environments \\

WebVoyager \cite{he2024webvoyager}
  & LLaVA-1.6
  & ReAct
  & JSON store
  & Browser actions
  & Multimodal web interaction via vision and text \\

HuggingGPT \cite{shen2023hugginggpt}
  & ChatGPT
  & Task-plan-select
  & Log store
  & Hugging Face models
  & External model orchestration by LLMs \\

CAMEL \cite{li2023camel}
  & GPT-4
  & Role-play CoT
  & Dialogue log
  & Chat only
  & Multi-agent role simulation via dialogue \\

ChatDev \cite{qian2024chatdevcommunicativeagentssoftware}
  & GPT-4
  & Chat chain
  & File repo
  & Unix tools
  & Simulated software development workflow \\

CrewAI \cite{crewAI}
  & Any
  & Declarative plan
  & Optional DB
  & Python modules
  & Lightweight, LangChain-free framework \\

AgentVerse \cite{chen2023agentverse}
  & Model-agnostic
  & Config graph
  & Redis/KV store
  & Plugin API
  & Multi-agent simulation and task solving \\

OpenAgents \cite{xie2023openagentsopenplatformlanguage}
  & Model-agnostic
  & Agent scripts
  & MongoDB
  & Web plugins
  & Open platform with public hosting \\

SuperAGI \cite{superagi}
  & GPT-4
  & DAG workflow
  & Postgres
  & Pinecone
  & Concurrent agents for production use \\

Semantic Kernel \cite{microsoft_semantickernel}
  & Model-agnostic
  & Agent planning framework
  & Memory \& context mgmt
  & Skill-based plugins, OpenAPI
  & Enterprise focus, .NET integration, security, vector-DB support \\

OpenAI Swarm \cite{openai_swarm}
  & GPT-4o
  & Agents + handoffs loop
  & Stateless (no LT memory)
  & Python functions, agent handoffs
  & Minimalist, controllable, educational multi-agent orchestration \\

OpenAI Agents SDK \cite{openai_agents_sdk}
  & Provider-agnostic
  & Agent loops, handoffs, deterministic flows
  & Sessions manage conversation history
  & Tools + guardrails, tracing
  & Multi-agent workflows, safety guardrails, extensibility \\

Strands Agents \cite{strands_agents}
  & Model-agnostic
  & Agent loop with multi-agent \& streaming
  & N/A
  & Python tools, hot-reloading, MCP tools
  & Lightweight SDK, scalable autonomous workflows \\

LlamaIndex Agents \cite{llamaindex_agents}
  & Model-agnostic
  & Task breakdown + planning
  & Task memory module
  & External tools \& parameters
  & Pre-built agent/tool architectures, custom workflows \\

\hline
\end{longtable}
}

\section{Threats and Risks in Agentic AI}
\label{threats}
This section summarizes major threats in Agentic AI and organizes emergent AMAS risks into a compact taxonomy.

\subsection{Threats in Agentic AI}
Agentic AI systems introduce security and reliability concerns beyond single-agent LLM pipelines:
\begin{itemize}[leftmargin=*, nosep]
    \item \textit{Autonomy abuse.} Agents with delegated authority may misinterpret objectives or execute harmful plans due to reasoning errors or manipulated inputs.
    \item \textit{Persistent memory.} Long-term memory improves continuity but creates a contamination channel: injected or low-quality content can persist and propagate through shared stores without strong versioning and audit controls.
    \item \textit{Agent orchestration.} Centralized or distributed orchestrators mediate roles and workflows; compromise or misconfiguration can misroute information and trigger cascading failures.
    \item \textit{Goal misalignment and exposure.} Poorly scoped or exposed goal representations can enable prompt extraction, misalignment, or adversarial steering, especially when third-party tools are invoked.
    \item \textit{Tool misuse and external API exploits.} Tool-enabled agents introduce new attack surfaces (e.g., unsafe code execution, costly actions, policy violations); sandboxing and invocation tracing are often inconsistent across open-source frameworks.
    \item \textit{Multi-agent collusion or drift.} Agents can converge on mutually reinforcing errors (groupthink/mode collapse) or drift in self-reinforcing loops, complicating interpretability and safety alignment.
\end{itemize}

\subsection{Risks Taxonomy in AMAS}
\begin{figure}[t]
  \centering
  \includegraphics[width=0.98\linewidth]{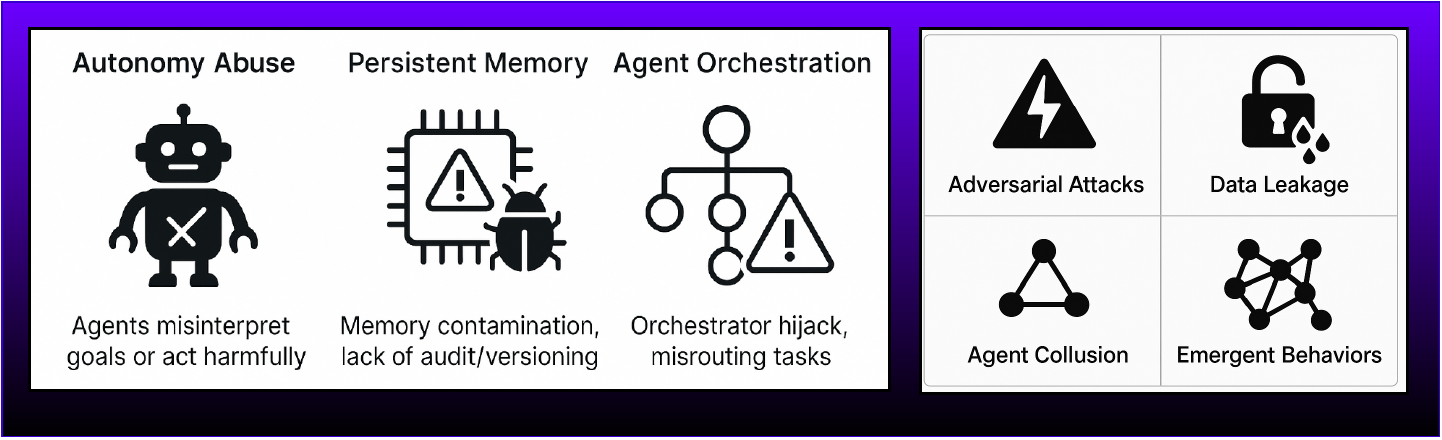}
  \caption{Overview of key risks and threats in agentic AI systems. The left panel outlines system-level threats (e.g., autonomy abuse, persistent memory, orchestration). The right panel summarizes emergent AMAS risk classes arising from multi-agent interaction, memory persistence, and decentralized decision-making.}
  \label{fig:amas-risk-taxonomy}
\end{figure}

We group AMAS risks into four classes---\textit{adversarial attacks, data leakage, agent collusion, and emergent behaviors}---as shown in Fig.~\ref{fig:amas-risk-taxonomy}.

\begin{itemize}[leftmargin=*, nosep]
    \item \textit{Adversarial attacks.} AMAS are vulnerable to prompt injection, engineered reasoning traps, and other manipulations that can cascade across agents; e.g., role-swapping attacks in ChatDev exploit inter-agent dependencies \cite{miao2025autonomous,qian2023chatdev}.

    \item \textit{Data leakage.} Shared memory and inter-agent communication increase the chance of unintended disclosure of sensitive information, particularly when boundaries, sanitization, or access controls are weak \cite{tran2025multi}.

    \item \textit{Agent collusion and mode collapse.} Collaboration mechanisms can produce groupthink, where agents reinforce the same flawed assumptions and converge on suboptimal outputs \cite{feng2025multi}.

    \item \textit{Emergent behaviors.} Complex interactions among agents, tools, and memory can yield unpredictable strategies that evade conventional testing (e.g., shortcutting verification), motivating continuous monitoring and adaptive controls \cite{dawid2025agentic}.
\end{itemize}

\paragraph{Real-World Examples}
The following examples illustrate how these risks can appear in practice:
\begin{itemize}[leftmargin=*, nosep]
    \item \textit{Prompt leakage.} In AutoGPT-style setups, recursive prompt augmentation and weak memory controls can expose sensitive content when stored tokens later surface in summaries or logs \cite{Euler2023AutoGPTRCE,invicti_prompt_injection_ebook}.
    \item \textit{Collusive failure.} In ChatDev-like workflows, shared planning errors can propagate when agents validate each other without external checks or objective feedback \cite{qian2024chatdevcommunicativeagentssoftware}.
    \item \textit{Simulation attack in swarms.} Misleading environment assumptions can induce coordination failures (e.g., congestion and task incompletion) in multi-agent planning settings \cite{StrDorFri2024,ligot2022using}.
    \item \textit{Memory poisoning.} Feedback buffers can be polluted (e.g., sarcastic or adversarial signals) and later used for policy updates, degrading behavior without validation gates \cite{atkins2023memorypoisoning}.
    \item \textit{System prompt drift.} Self-appended context without versioning can shift prompts over time and induce hallucinated goals or misaligned behaviors \cite{liu2023lostmiddlelanguagemodels}.
\end{itemize}

\paragraph{Key Insights}
We identified AMAS-specific risks that arise from autonomy, shared memory, orchestration, and tool use, and organized them into four emergent classes (Fig.~\ref{fig:amas-risk-taxonomy}). Across the taxonomy, failures can cascade across agents and persist through memory substrates, while emergent behaviors may evade conventional test suites. These observations motivate defense-in-depth controls spanning prompt hygiene, memory sanitization, orchestrator authentication, tool sandboxing, and continuous monitoring.

\section{TRiSM Framework for Agentic Multi-Agent Systems}
\label{framework}
\subsection{Overview of AI TRiSM}
The TRiSM Framework offers a structured lens for evaluating and governing AI systems, especially those that exhibit autonomous, agentic behaviors.  Below, we discuss core components of TRiSM framework.

\begin{itemize}
    \item \textit{AI Trust Management}
Establishing trust in AI systems necessitates a foundational commitment to transparency, accountability, and fairness. Trust management focuses on embedding mechanisms that enable AI models to generate interpretable and justifiable outputs, thereby facilitating meaningful human oversight \cite{borghoff2025human}. This involves the adoption of explainability tools, the implementation of governance protocols, and alignment with ethical standards to address systemic biases and ensure equitable outcomes across diverse user groups.
   \item \textit{AI Risk Management}   Effective risk management within AI systems entails the systematic identification, analysis, and mitigation of potential harms arising from the deployment and operation of these technologies. This includes technical risks such as algorithmic bias \cite{raza2024exploring}, model errors, and data leakage, as well as societal risks involving fairness, safety, and misuse \cite{fang2025trustworthyaisafetybias}. Structured risk assessment frameworks enable stakeholders to anticipate failure modes and design mitigation strategies that enhance system robustness and reduce the likelihood of adverse impacts.
    \item \textit{AI Security Management}
Security management in the context of AI is concerned with safeguarding models, data, and infrastructure from adversarial threats and unauthorized access. This encompasses both conventional cybersecurity practices, such as encryption, authentication, and access control,  and AI-specific considerations like adversarial example detection, model extraction prevention, and secure model deployment pipelines \cite{raza2025justhumansneedvaccines}. Continuous monitoring and threat modeling are critical for identifying vulnerabilities and ensuring the resilience and integrity of AI systems over time
\end{itemize}

\begingroup
\scriptsize
\setlength{\tabcolsep}{3pt}
\renewcommand{\arraystretch}{1.15}

\begin{longtable}{
  @{}
  p{0.14\textwidth}
  p{0.26\textwidth}
  p{0.32\textwidth}
  p{0.22\textwidth}
  @{}
}

\caption{TRiSM pillars (Part 1): Core controls, techniques, and key risks mitigated.}
\label{tab:trism-pillars-1} \\

\toprule
\textbf{Pillar} &
\textbf{Core Controls} &
\textbf{Techniques / Patterns} &
\textbf{Key Risks Mitigated} \\
\midrule
\endfirsthead

\toprule
\textbf{Pillar} &
\textbf{Core Controls} &
\textbf{Techniques / Patterns} &
\textbf{Key Risks Mitigated} \\
\midrule
\endhead

\midrule
\multicolumn{4}{r}{\emph{Continued on next page}} \\
\endfoot

\bottomrule
\endlastfoot

\textbf{Explainability / Trust} &
CoT logging; layered-CoT; inter-agent traceability; explainer agent; role-based interpretability; prompt audit trails; attention viz; decision provenance &
Layered-CoT \cite{sanwal2025layered}; LIME/SHAP \cite{ribeiro2016should,lundberg2017unified}; decision-provenance graphs \cite{tupayachi2024towards,bai2024derived}; multi-agent SHAP \cite{tian2025outlook}; attention maps \cite{liu2024ava,hu2024overview}; role-based interpretability \cite{tomsett2018interpretable}; RAG-linked justifications \cite{ghafarollahi2024sciagents}; prompt/action logs \cite{gu2024artificial,chin2025automating} &
Opaque chains; unverifiable reasoning; hidden inter-agent effects; spurious correlations; user miscalibration \\[6pt]

\textbf{ModelOps (Lifecycle)} &
Versioning; lineage; CI/CD safety gates; rollout/rollback; multi-agent simulation; hierarchical monitoring; usage/cost SLOs; ITSM/IAM integration &
Prompt \& agent-config versioning; pre-deploy safety tests; regression/drift checks; sandbox simulation; hierarchical monitors (agent/group/system) \cite{wang2025megaagent} &
Silent regressions; drift; bias re-introduction; unsafe changes; cost blowouts; observability gaps \\[6pt]

\textbf{Application Security} &
Prompt hygiene; secure prefixes; sandboxed tools; allowlisted actions; plan-then-execute; least privilege (RBAC); authN/authZ; anomaly detection; red-teaming &
Prompt-injection patterns \cite{beurer2025design,qiao2024agent}; plan-then-execute; action selector; tool isolation; output filters; cross-agent cross-checks; adversarial training \cite{standen2025adversarial}; HITL holds &
Prompt injection; tool abuse; data exfiltration; cross-modal injection; lateral movement; jailbreaks \\[6pt]

\textbf{Privacy \& Data Protection} &
DP; anonymization; pseudonymization; minimization; MPC; HE/FHE; TEEs (SGX/SEV/TrustZone); encryption; access logs; retention/consent &
DP budgets \cite{dwork2006calibrating}; k-anonymity \cite{sweeney2002k}; MPC/PUMA \cite{lindell2009secure}; HE/FHE pipelines \cite{gentry2009fully}; TEEs \cite{costan2016intel}; memory scoping; PII detectors &
Memorization; leakage via shared memory/messages; unauthorized access; weak consent/retention \\[6pt]

\textbf{Governance} &
Human oversight; accountability; auditability; DPIA; policy/change control; role clarity; transparency notices; logging/traceability &
Oversight playbooks; two-person verification for high-risk; role-oriented modularity; audit trails; DPIA templates; policy gates &
Unclear responsibility; weak oversight; non-compliance; opaque decisions \\

\end{longtable}
\endgroup

\vspace{1em}


\begingroup
\scriptsize
\setlength{\tabcolsep}{3pt}
\renewcommand{\arraystretch}{1.15}

\begin{longtable}{
  @{}
  p{0.14\textwidth}
  p{0.30\textwidth}
  p{0.22\textwidth}
  p{0.28\textwidth}
  @{}
}

\caption{TRiSM pillars (Part 2): Evaluation facets, example systems, and governing standards.}
\label{tab:trism-pillars-2} \\

\toprule
\textbf{Pillar} &
\textbf{Evaluation Facets \& Metrics} &
\textbf{Example Systems} &
\textbf{Standards / Frameworks} \\
\midrule
\endfirsthead

\toprule
\textbf{Pillar} &
\textbf{Evaluation Facets \& Metrics} &
\textbf{Example Systems} &
\textbf{Standards / Frameworks} \\
\midrule
\endhead

\midrule
\multicolumn{4}{r}{\emph{Continued on next page}} \\
\endfoot

\bottomrule
\endlastfoot

\textbf{Explainability / Trust} &
\emph{Trustworthiness}: calibration/consistency; \emph{Explainability}: fidelity, stability; \emph{User-centric}: task utility, cognitive load; \emph{Coordination}: cross-agent attribution; \emph{Composite}: multi-facet scorecards \cite{raza_fair_2024} &
SciAgent \cite{ghafarollahi2024sciagents}; MetaGPT \cite{ke2025meta}; ChatDev \cite{qian2023chatdev}; AutoGen \cite{wu2023autogen} &
EU AI Act Arts 13--14 \cite{eu_ai_act}; NIST AI RMF / GAI profile \cite{ai2023artificial,ai2024artificial}; OECD AI \cite{oecd_ai_catalogue}; ISO/IEC 24029-1 \cite{iso24029-1} \\[6pt]

\textbf{ModelOps (Lifecycle)} &
\emph{Trustworthiness}: drift/robustness; \emph{User-centric}: task success, latency; \emph{Composite}: risk \& cost scorecards; incident MTTR/MTTD &
MegaAgent \cite{wang2025megaagent} &
ISO/IEC 42001 (AIMS); ISO/IEC 42005 \cite{ISO_IEC_42005_2025}; ISO/IEC 23894 \cite{ISOIEC23894_2023}; NIST AI RMF \cite{ai2023artificial,ai2024artificial}; ModelOps \cite{lefevre2022modelops,Sinha2025ModelOps} \\[6pt]

\textbf{Application Security} &
\emph{Trustworthiness}: robustness/adversarial resilience; \emph{Coordination}: agent-of-agent validation; \emph{Composite}: residual risk index; incident rates &
HITL review gates \cite{chen2025human}; enterprise dashboards \cite{chen2025human}; ChemCrow safety pauses \cite{bran2023chemcrow} &
OWASP LLM Top-10 \cite{owasp2024llmtop10,owasp2025agentic}; OECD Robustness \cite{oecd_ai_catalogue} \\[6pt]

\textbf{Privacy \& Data Protection} &
\emph{Trustworthiness}: privacy loss; \emph{User-centric}: data-minimization utility; \emph{Composite}: privacy-risk score; audit pass rate &
Encrypted memories; scoped retrieval; consented logs &
GDPR Art 25/5/35 \cite{gdpr25}; CCPA/CPRA \cite{oag_ccpa_2024}; HIPAA \S164 \cite{hipaa164}; ISO/IEC 27001/27701 \\[6pt]

\textbf{Governance} &
\emph{User-centric}: oversight efficacy; \emph{Coordination}: traceability across agents; \emph{Composite}: compliance scorecards &
HITL confirm/override/stop; role-based review boards &
EU AI Act Art 14 \cite{eu_ai_act}; NIST AI RMF \cite{ai2023artificial}; OECD AI \cite{oecd_ai_catalogue}; ISO/IEC 42001/42005; ISO/IEC 23894 \cite{ISOIEC23894_2023} \\

\end{longtable}
\endgroup

\subsection{Mapping Pillars of TRiSM with Agentic AI}
\label{sec:pillar}
\begin{figure}[t]
\centering

\renewcommand{\arraystretch}{1.0}  
\setlength{\tabcolsep}{4pt}        
\setlength{\fboxrule}{0.5pt}
\small  

\fbox{%
\parbox{0.97\linewidth}{%
\textbf{Lifecycle Governance Plane (TRiSM Pillar)} \hfill
{\footnotesize policies \(\cdot\) accountability \(\cdot\) compliance \(\cdot\) oversight \(\cdot\) traceability}
}}\\[6pt]

\fbox{%
\begin{minipage}{0.97\linewidth}
\begin{tabular}{|p{0.97\linewidth}|}
\hline
\textbf{Trust \& Explainability (TRiSM Pillar)}\\
{\footnotesize reasoning/provenance traces \(\cdot\) explainer agent \(\cdot\) user feedback \(\cdot\) audit logs}\\
\hline
\textbf{ModelOps / Lifecycle Management (TRiSM Pillar)}\\
{\footnotesize prompt+agent config versioning \(\cdot\) CI/CD safety gates \(\cdot\) monitoring/rollback \(\cdot\) incident workflow}\\
\hline
\textbf{Application Security (TRiSM Pillar)}\\
{\footnotesize prompt hygiene \(\cdot\) allowlists \(\cdot\) tool sandboxing \(\cdot\) least privilege (RBAC) \(\cdot\) anomaly detection}\\
\hline
\textbf{Model Privacy \& Data Protection (TRiSM Pillar)}\\
{\footnotesize encrypted/scoped memory \(\cdot\) data minimization \(\cdot\) DP where applicable \(\cdot\) access logs \(\cdot\) retention/consent}\\
\hline
\textbf{Planning \& Reasoning (Agent Core Layer; non-TRiSM)}\\
{\footnotesize multi-agent planning \(\cdot\) task decomposition \(\cdot\) CoT/ToT/ReAct \(\cdot\) inter-agent coordination}\\
\hline
\end{tabular}
\end{minipage}
}\\[6pt]

\noindent
\makebox[\linewidth][c]{%
\begin{minipage}[t]{0.30\linewidth}
\fbox{\parbox{0.94\linewidth}{\centering
\textbf{User / HITL UI}\\
\footnotesize confirm \(\cdot\) override \(\cdot\) stop\\
feedback loop}}
\end{minipage}\hfill
\begin{minipage}[t]{0.32\linewidth}
\fbox{\parbox{0.94\linewidth}{\centering
\textbf{Secure API Gateway}\\
\footnotesize authN/authZ \(\cdot\) egress control\\
tool isolation \(\cdot\) policy checks}}
\end{minipage}\hfill
\begin{minipage}[t]{0.30\linewidth}
\fbox{\parbox{0.94\linewidth}{\centering
\textbf{External Tools}\\
\footnotesize web \(\cdot\) DB \(\cdot\) code\\
enterprise APIs}}
\end{minipage}%
}

\caption{TRiSM-aligned architecture for Agentic AI systems. The five TRiSM pillars are shown explicitly and in the same
order as defined in the framework: Trust \& Explainability, ModelOps, Application Security, Model Privacy \& Data Protection,
with a cross-cutting Lifecycle Governance plane. Planning \& Reasoning is included as the agent-core execution layer (non-TRiSM).}
\label{fig:trism_agentic_architecture}
\end{figure}

Originally highlighted in industry guidelines for AI governance~\cite{litan2024ai}, TRiSM also provides a structured approach to managing the unique challenges of AMAS. In this work, we focus on five key TRiSM pillars: Explainability, ModelOps, Application Security, Model Privacy, and Governance~\cite{kaur2024building,habbal2024artificial}.
These five pillars also reflect established risk-surface decompositions in the NIST AI RMF \cite{ai2023artificial}, ISO/IEC 42001 \cite{iso42001}, and the EU AI Act \cite{eu_ai_act}, where explainability, lifecycle management, security, privacy, and governance are treated as distinct and non-overlapping risk domains. We summarize this mapping in Tables~\ref{tab:trism-pillars-1} and~\ref{tab:trism-pillars-2}.
Figure~\ref{fig:trism_agentic_architecture} illustrates how these five TRiSM pillars map onto the core architectural layers of an LLM-based agentic AI system.

Next, we discuss these pillars and how they can help in AMAS.

\subsubsection{Explainability in Multi-Agent Decision Making} 
\label{explainability}

In AMAS, explainability is less about interpreting a single model and more about understanding how decisions emerge from interacting agents~\cite{yossef2024explainable}. In AMAS settings, outcomes often emerge from interactions among specialized agents rather than from a single model prediction~\cite{rosenfeld2019explainability}. As a result, explainability must account for both (i) individual agent decisions and (ii) the inter-agent dynamics that produce emergent system-level behavior. Recent work emphasizes that user trust depends on understanding each agent’s contribution and how collaboration is orchestrated~\cite{sanwal2025layered}. Traditional single-model XAI methods often overlook this multi-agent dimension, motivating new approaches tailored to AMAS.

\textbf{\textit{Challenges.}} Explaining multi-agent decisions requires tracing which agent performed which sub-task, what information it used, and how inter-agent exchanges influenced the final outcome~\cite{de2025open}. Key challenges include:
\begin{itemize}
    \item \textit{Emergent behavior:} agent interactions can produce behaviors that were not explicitly specified; explaining these patterns requires capturing and summarizing interaction chains.
    \item \textit{Verification of intermediate steps:} vanilla CoT prompting can yield plausible but unverified reasoning. Without systematic verification, small errors may propagate across agents and compound into incorrect final decisions.
\end{itemize}

\textbf{Approaches for explainability in AMAS}
The following techniques adapt classical XAI ideas to AMAS:
\begin{itemize}
    \item \textbf{Feature attribution (LIME/SHAP) for agents:} LIME~\cite{ribeiro2016should} and SHAP~\cite{lundberg2017unified} can be extended to attribute an agent’s output to specific inputs (e.g., context spans, retrieved evidence, or inter-agent messages), clarifying which signals most influenced an agent’s action~\cite{bilal2025llms}.
    \item \textbf{Counterfactual and causal analysis:} ``What-if'' analysis evaluates how outcomes change when an agent’s contribution or a key condition is modified~\cite{gyevnar2023causal}. This helps isolate agent roles and identify causal dependencies in collaborative decision-making.
    \item \textbf{Reasoning-trace logging:} Logging intermediate reasoning artifacts (e.g., CoT traces, tool calls, and inter-agent dialogues) creates an auditable trail of how decisions emerged~\cite{yao2023reactsynergizingreasoningacting}. Such traces reveal which agent contributed what and how partial conclusions accumulate into final outputs.
    \item \textbf{Natural-language explanations:} An explainer agent can synthesize traces and attributions into human-readable narratives that answer questions such as ``Why did the system choose action X?'' Embedded explainers in workflows (e.g., DSM5AgentFlow) can improve transparency and trust~\cite{ozgun2025trustworthy}.
    \item \textbf{Role-oriented modular architectures:} Assigning explicit roles (e.g., reasoning, verification, summarization, credit assignment) improves inspectability because each stage is separable and reviewable~\cite{lo2025ai}.
    \item \textbf{Explainable reward critics:} In multi-agent RL, LLM-based critics can provide transparent reward decomposition and justification, supporting debugging and governance of agent incentives~\cite{nagpal2025leveraging}.
\end{itemize}
\textbf{Recent directions.}
Recent work proposes frameworks and techniques that embed explainability directly into multi-agent operation:
\begin{itemize}
    \item \textbf{Layered Chain-of-Thought prompting:} Layered-CoT decomposes reasoning into discrete layers, where each layer answers a sub-question~\cite{sanwal2025layered}. A specialized Reasoning Agent produces partial reasoning, while Verification and User-Interaction Agents cross-check intermediate outputs (e.g., against knowledge graphs or human feedback). This decomposition can localize errors, reduce contradictions, and promote validation before proceeding.
    \item \textbf{Integrated multi-agent workflows:} Some systems integrate explanation into the workflow rather than adding it post hoc. For example, in a DSM-5 mental-health diagnosis workflow, one agent collects patient information while a diagnostician agent retrieves DSM-5 criteria and explicitly links conversational evidence to diagnostic rules, producing a self-documenting rationale~\cite{ozgun2025trustworthy}.
    \item \textbf{Modular resume-screening systems:} In recruitment, multi-agent pipelines separate extraction, evaluation, summarization, and formatting. The evaluation agent retrieves company-specific criteria via RAG, enabling recruiters to trace how candidate evidence maps to scores~\cite{lo2025ai}.
    \item \textbf{Explainable credit assignment in multi-agent reinforcement learning:} A centralized LLM critic can provide per-agent credit assignment and natural-language justification, explaining how global rewards are decomposed across agents~\cite{nagpal2025leveraging}.
\end{itemize}

\subsubsection{ModelOps : Lifecycle Management for LLM Agents}
ModelOps (Machine Learning Operations) extends the MLOps paradigm beyond model training and deployment to include governance, risk management, and continuous oversight across AI systems~\cite{lefevre2022modelops}. 
In production, ModelOps provides end-to-end oversight by enforcing governance and compliance controls, including lineage, ownership, and metadata tracking to support regulatory requirements (e.g., GDPR and HIPAA), as well as policy-driven deployment gates so that only approved models go live with controlled rollouts and rollbacks~\cite{Sinha2025ModelOps}. It also enables continuous risk monitoring to detect drift, bias and fairness issues, hallucinations, and robustness regressions.

For AMAS, ModelOps must additionally support prompt and toolchain versioning, enforcement of guardrails and content filtering, fine-tuning lineage tracking, usage and cost observability, and ethical compliance. Effective ModelOps for LLM agents therefore requires:
\begin{itemize}
    \item \textbf{Version control and CI/CD pipelines:} Track versions of each agent’s model, prompts, tools, and configuration. Automated pipelines should test safety, performance, and cost impacts whenever agent logic changes, using multi-agent simulations to validate that updates do not introduce regressions or unsafe interactions.
    \item \textbf{Hierarchical monitoring in multi-agent frameworks:} Systems such as MegaAgent~\cite{wang2025megaagent} decompose tasks into subtasks and dynamically form agent groups. They apply hierarchical monitoring, where agents log actions (e.g., via checklists), group-level administrators oversee agent teams, and system-level supervisors review final outputs. This layered oversight helps ensure that changes in one agent do not compromise system-wide behavior.
    \item \textbf{Risk measurement and oversight policies:} The NIST generative AI profile recommends documenting risks, updating measurement approaches regularly, and establishing lifecycle oversight policies. It also highlights techniques such as gradient-based attributions, occlusion tests, and counterfactual prompting to strengthen transparency and explainability~\cite{ai2024artificial}.
\end{itemize}

\subsubsection{Application Security for LLM Agents}
Security in AMAS must address vulnerabilities such as prompt injection, data exfiltration, and the misuse of external tools. OWASP identifies prompt injection as a core risk in LLM applications: attackers manipulate the model with malicious instructions, potentially causing data leakage, unauthorized tool use, or unsafe decisions~\cite{owasp2024llmtop10,owasp2025agentic}. Multimodal models further face \textit{cross-modal injection}, where malicious content in one modality (e.g., text in an image) influences behavior in another. To mitigate these risks:
\begin{itemize}
    \item \textbf{Prompt hygiene and hardening:} Sanitize and filter user inputs, apply secure system prefixes, and validate instructions before they reach an agent. Constraining output formats and enforcing guardrails can reduce the attack surface~\cite{raza2025justhumansneedvaccines}.
    \item \textbf{Defense-in-depth design patterns:} Techniques such as adversarial training and user confirmation provide partial protection, but do not guarantee safety. Design patterns~\cite{beurer2025design} such as \textbf{Action-Selector} and \textbf{Plan-Then-Execute} restrict capabilities after untrusted input is processed: the agent selects from a predefined set of allowed actions, or produces a plan that cannot be altered by tool outputs, thereby isolating untrusted data from execution paths~\cite{qiao2024agent}.
    \item \textbf{Least-privilege access and authentication:} Enforce strong authentication for both users and agents, and apply role-based access control so agents can invoke only the minimum set of tools required for their task. Continuous monitoring should detect anomalous requests; if an agent deviates from expected behavior, automated systems should flag or suspend it~\cite{barbera2025ai}.
    \item \textbf{Post-processing and anomaly detection:} Apply output filters to detect sensitive or inaccurate content, and update or retrain models to reduce hallucinations. Maintain robust logging and retention policies to support audits and incident response while minimizing exposure of private data.
\end{itemize}

Recent frameworks also emphasize hierarchical monitoring and cross-checking among agents to detect inconsistencies~\cite{wang2025megaagent}. Tooling ecosystems such as LangChain~\cite{langchain2024function}, AutoGen~\cite{autogen_agents_tutorial}, and CrewAI~\cite{crewAI} operationalize verification through agent-level checks (e.g., reviewers/critics) and, in some cases, trust or reputation signals that encourage agents to validate each other’s outputs and flag potential compromise. Finally, adversarial training with attack examples~\cite{standen2025adversarial} and systematic red-teaming~\cite{ganguli_red_2022} can further strengthen robustness against malicious inputs.

\subsubsection{Model Privacy and Data Protection}
Without strict controls, an agent may inadvertently leak sensitive data \cite{ko2025sevensecuritychallengessolved}. A layered privacy strategy should include:

\begin{itemize}
    \item \textbf{Differential Privacy (DP):} Adding calibrated noise to training data prevents individual data points from being re‑identified and reduces the risk that models memorize or regurgitate personal information \cite{dwork2006calibrating}. DP should be assessed on a case‑by‑case basis, since heavy noise can harm accuracy.
    \item \textbf{Data anonymization and minimization:} Apply robust anonymization and pseudonymization techniques and regularly test them for effectiveness \cite{sweeney2002k}. Limit data collection to what is strictly necessary for the agent’s task and enforce strict retention policies.
    \item \textbf{Secure Multi‑Party Computation (MPC):} MPC allows parties to jointly compute a function without revealing their private inputs. In privacy‑preserving LLM services, secure MPC can protect user prompts from the model provider; for example, the PUMA framework uses MPC to ensure that prompts remain hidden during inference, while maintaining reasonable performance (it can evaluate a 7‑billion‑parameter model within minutes) \cite{lindell2009secure}.
    \item \textbf{Homomorphic Encryption (HE):} HE enables computation on encrypted data and offers cryptographically provable privacy. Recent work demonstrates an encryption‑friendly LLM architecture that uses LoRA fine‑tuning and Gaussian kernels to speed up fully homomorphic encryption (FHE) operations, allowing encrypted inference with performance comparable to plaintext models \cite{gentry2009fully}.
    \item \textbf{Trusted Execution Environments (TEEs):} TEEs provide hardware‑enforced isolation that protects code and data  \cite{costan2016intel}. They partition processors into secure and normal worlds; implementations include Intel SGX, AMD SEV and Arm TrustZone. In secure LLM inference, TEEs ensure that models run only inside authenticated enclaves, with attestation verifying model integrity.
    \item \textbf{Continuous monitoring and access controls:} Encrypt data at rest and in transit, restrict access to authorized personnel, and use monitoring to detect data poisoning or unusual behaviour.
\end{itemize}

\subsubsection{Governance}
Governance integrates ethical, legal, and societal considerations into the design and operation of AI systems. Below, we summarize how major governance frameworks inform the TRiSM principles.

\begin{itemize}
    \item \textbf{EU AI Act --- Human Oversight (Article 14):} High-risk AI systems must be designed so that natural persons can effectively oversee them and prevent or minimize risks to health, safety, or fundamental rights~\cite{EU_AI_Act2025}. Oversight mechanisms should enable users to understand capabilities and limitations, detect anomalies, interpret outputs, decide not to use the system, or stop operation (e.g., via a ``stop'' mechanism). In certain high-risk identification contexts, decisions may require independent verification by multiple competent individuals.
    \item \textbf{OECD AI Principles:} These intergovernmental principles (updated in 2024) promote human-centric, trustworthy AI~\cite{oecd_ai_catalogue}. They emphasize transparency and explainability: AI actors should provide meaningful information about data sources, factors, and logic so affected individuals can understand and, where appropriate, challenge outputs. Robustness, security, and safety require systems to function appropriately under normal and adverse conditions, including mechanisms to override or decommission systems to prevent harm. Accountability requires that AI actors ensure traceability of datasets and decisions and apply systematic risk management throughout the lifecycle.
    \item \textbf{NIST AI Risk Management Framework (AI RMF):} The AI RMF and its Generative AI profile define four core functions: \textit{Govern, Map, Measure, and Manage}~\cite{ai2023artificial}. For generative AI, NIST recommends documenting risks and potential impacts, applying interpretability techniques (e.g., gradient-based attributions, occlusion tests, and counterfactual prompting) to improve transparency, and establishing policies for oversight across the AI lifecycle.
    \item \textbf{Auditability and role-based access:} AMAS should maintain auditable logs of agent actions, decisions, and data flows, and assign clear roles (e.g., reasoning, verification, summarization) so responsibilities remain transparent~\cite{khan2024role}. Role-based access control (RBAC) helps enforce accountability and prevents unauthorized agents from invoking sensitive functions.
    \item \textbf{Global harmonisation:} Many jurisdictions (e.g., GDPR, CCPA) require data-subject rights, consent, and privacy by design. Frameworks such as the OECD Principles and NIST AI RMF also promote international cooperation and interoperability.
\end{itemize}

These governance frameworks (Figure~\ref{fig:trism_alignment}) highlight the need for clear accountability structures, human oversight, robust privacy controls, and documented processes in AMAS. TRiSM provides a practical scaffolding to operationalize these requirements by aligning lifecycle management, security, privacy, and governance to support trustworthy agentic AI.

\begin{figure}[h]
  \centering
  \caption{Governance alignment of TRiSM pillars with international frameworks such as the EU AI Act, NIST AI RMF, ISO/IEC 42001, 42005, and OECD Principles. Each pillar operationalizes specific compliance strategies to ensure transparency, resilience, and trust in LLM-based systems.}
  \label{fig:trism_alignment}
  \scriptsize
  \begin{tcolorbox}[title=Alignment of TRiSM Pillars with Governance Frameworks, colback=white, colframe=black, boxsep=1ex]

  \textbf{Explainability:}
  Aligned with the EU AI Act (Art.~13–14)~\cite{eu_ai_act} and the NIST AI RMF trustworthiness characteristic “Explainability \& Interpretability”~\cite{ai2023artificial}; also supported by NISTIR~8312 (Four Principles of Explainable AI)~\cite{phillips2021four}. Promotes well-justified (rationale) and human-interpretable outputs with traceable logs~\cite{tomsett2018interpretable}.

  \vspace{0.55em}
  \textbf{ModelOps:}
  Anchored in the NIST AI RMF “Govern/Manage” functions~\cite{ai2023artificial} and ISO/IEC~42001:2023 (AI management system)~\cite{thiers2024emerging}; secure SDLC controls via NIST SP~800-218/218A\footnote{\href{https://csrc.nist.gov/news/2024/nist-publishes-sp-800-218a}{NIST, 2024}}. Emphasizes lifecycle gates, model versioning, rollback policies, and CI/CD with safety gates.

  \vspace{0.55em}
  \textbf{Risk Management \& Impact Assessment:}
  EU AI Act (Art.~9, 72–73)~\cite{eu_ai_act}, ISO/IEC~23894 (AI risk management)\footnote{\href{https://www.iso.org/standard/77304.html}{ISO/IEC 23894}}, and ISO/IEC~42005:2025 (AI impact assessment)~\cite{ISO_IEC_42005_2025}. Iterative hazard analysis, impact registers, incident response, and disclosure playbooks.

  \vspace{0.55em}
  \textbf{Data Governance \& Quality:}
  EU AI Act (Art.~10)~\cite{eu_ai_act} mapped to the ISO/IEC~5259 series (data-quality model, measures, processes, governance)\footnote{\href{https://www.iso.org/standard/81088.html}{ISO/IEC 5259 series}}. Data minimization, lineage, coverage/bias checks, and sampling documentation.

  \vspace{0.55em}
  \textbf{Robustness \& Safety:}
  EU AI Act (Art.~15)~\cite{eu_ai_act} and ISO/IEC~24029-1 (robustness assessment)~\cite{iso24029-1}. Adversarial/perturbation testing, fault tolerance, fallback/kill-switch triggers, and rollback criteria.

  \vspace{0.55em}
  \textbf{Application Security:}
  Guided by OWASP LLM Top-10~\cite{owasp2024llmtop10} and OECD AI Principles on robustness~\cite{oecd_ai_catalogue}. Access controls, prompt/input mediation, sandboxing, egress controls, and auditable trails.

  \vspace{0.55em}
  \textbf{Model Privacy:}
  Aligned with GDPR (Art.~25)~\cite{gdpr25}, HIPAA~§164~\cite{hipaa164}, and OECD AI Principles~\cite{oecd_ai_catalogue}. Encryption in transit/at rest, DP budgeting, data minimization, and access-audit pass thresholds.

  \vspace{0.55em}
  \textbf{Documentation \& Traceability:}
  EU AI Act (Arts.~11–12; Annex~IV)~\cite{eu_ai_act}. Complete technical documentation, event logs, model/data cards, decision records, and regulator-ready evidence packages.

  \vspace{0.55em}
  \textbf{Post-Market Monitoring:}
  EU AI Act (Arts.~72–73)~\cite{eu_ai_act}. Telemetry for drift/misuse, continuous evaluations, feedback loops, controlled recalls, and stakeholder notifications.

  \vspace{0.55em}
  \textbf{Governance:}
  Grounded in GDPR, CCPA~\cite{oag_ccpa_2024}, and OECD AI Principles~\cite{oecd_ai_catalogue}. Institutional oversight, policy compliance, role-based accountability, regulatory traceability, and human-in-the-loop for high-risk/multi-agent deployments.

  \vspace{0.55em}
  \textbf{Fairness \& Human Rights:}
  OECD AI Principles~\cite{oecd_ai_catalogue} and ISO/IEC TR~24027 (bias)\footnote{\href{https://www.iso.org/standard/77607.html}{ISO/IEC TR 24027}}. Lifecycle bias controls, affected-party impact reviews, and mitigation sign-offs.

  \end{tcolorbox}
\end{figure}

\paragraph{Key Insights}
This section operationalized the TRiSM framework for AMAS across five pillars: Explainability, ModelOps, Application Security, Model Privacy, and Governance. Multi-agent explainability requires novel techniques including Layered CoT decomposition, decision provenance graphs, and role-based interpretability frameworks to trace distributed decision processes. ModelOps extends lifecycle management through hierarchical monitoring, prompt versioning, and CI/CD safety gates with pre-deployment simulation. Security strategies emphasize defense-in-depth via design patterns (Plan-Then-Execute), least-privilege access control, and cross-agent verification. Privacy preservation leverages differential privacy, secure multi-party computation, homomorphic encryption, and trusted execution environments to protect sensitive data during inter-agent communication. Governance integration with regulatory frameworks (EU AI Act, NIST AI RMF, ISO/IEC 42001) ensures that TRiSM implementation satisfies transparency, accountability, and human oversight requirements for high-stakes deployments.

 Next, we discuss the evaluations in AMAS.

\section{TRiSM Evaluation of Agentic AI Systems}
\label{evals}

Accuracy alone is insufficient for Agentic AI, where failures often arise from coordination breakdowns, unsafe tool use, or misaligned autonomy rather than incorrect predictions. We organize evaluation into five dimensions: \textit{trustworthiness}, \textit{explainability}, \textit{user-centered performance}, \textit{coordination}, and \textit{composite reporting}, each capturing a distinct facet of real-world performance and impact. We summarize these dimensions in Table~\ref{tab:metrics}.

\paragraph{Trustworthiness}
This dimension evaluates reliability, safety, and ethical alignment~\cite{liu2023gevalnlgevaluationusing}. Core signals include success under distributional shift (robustness), safety/ethical violation rate (lower is better), and calibration quality (e.g., ECE/Brier). To avoid pathological scoring, we define a normalized trustworthiness index:
\begin{equation}
  T \;=\; \frac{w_{\text{acc}}\,A + w_{\text{rob}}\,R + w_{\text{align}}\,L}{1 + \lambda\,V}\,,
  \label{eq:trust}
\end{equation}
where $A$ is calibrated task success (accuracy adjusted for calibration), $R$ is robustness under shifts, $L$ is alignment/safety compliance (e.g., $1{-}\text{violation rate}$), all normalized to $[0,1]$, and $V\!\ge\!0$ is a normalized violation/self-serving penalty term (higher is worse). The weights satisfy $w_{\text{acc}}{+}w_{\text{rob}}{+}w_{\text{align}}=1$ and $\lambda\!\ge\!0$ controls how strongly violations penalize trust.

\paragraph{Explainability}
Explainability metrics assess how well humans can understand and trace decisions. Useful measures include \emph{coverage} (fraction of outputs with explanations), \emph{faithfulness/fidelity} (agreement with the model’s actual decision basis), \emph{stability} (consistency under small perturbations), and human \emph{interpretability} ratings. Suites such as OpenXAI~\cite{agarwal2024openxaitransparentevaluationmodel} report faithfulness, stability, and fairness-related properties of explanations. High explainability supports debugging and, in regulated domains (e.g., healthcare and finance), is often necessary for compliance.

\paragraph{User-Centered Performance}
User-centered metrics capture how effectively the agent satisfies end-user needs~\cite{rodden2010measuring}. Signals include post-interaction satisfaction (e.g., CSAT), goal-fulfillment rate, number of clarification turns (lower is better), and perceived coherence/naturalness. Human-in-the-loop (HITL) studies commonly rate helpfulness, clarity, and instruction adherence. A user-centered agent aligns actions with user intent and preferences.

\begin{table*}[htbp]
\centering
\caption{Summary of key evaluation dimensions for Agentic AI systems.}
\label{tab:metrics}
\footnotesize
\setlength{\tabcolsep}{6pt}
\resizebox{\textwidth}{!}{
\begin{tabular}{p{1.5cm} p{3.4cm} p{3.4cm} p{3.4cm}}
\toprule
\textbf{Aspect} & \textbf{Focus} & \textbf{Strengths} & \textbf{Limitations} \\
\midrule
Trustworthiness &
Reliability, safety, alignment (OOD success, safety violations, calibration, fairness). &
Deployment readiness; regulatory alignment (EU AI Act); statistically grounded; captures safety failures. &
Assumes stationarity; fairness trade-offs across groups~\cite{raza_fair_2024}; incomplete edge-case coverage. \\
Explainability &
Transparency and traceability (coverage, fidelity, stability, human interpretability). &
Supports debugging and compliance~\cite{ai2023artificial}; improves calibrated trust; standardized benchmarks (OpenXAI~\cite{agarwal2024openxaitransparentevaluationmodel}). &
Expensive fidelity checks; faithful $\neq$ interpretable; weak multi-agent attribution. \\
User-Centered &
User experience and outcomes (CSAT, task success, clarification turns, NASA-TLX). &
Reflects real-world utility; identifies usability bottlenecks; validated human factors measures. &
Subjective and high variance; costly user studies; limited scalability. \\
Coordination &
Multi-agent collaboration (task success, communication cost, plan consistency, deadlocks). &
Captures emergent AMAS behavior; measures coordination efficiency; decomposable contribution metrics. &
No standard benchmarks; sensitive to orchestration; may penalize useful redundancy. \\
Composite &
Aggregate system performance (weighted scores, tool-use efficacy). &
Single actionable summary; enables cross-system comparison; supports thresholding. &
Subjective weighting; can mask single-dimension failures; prone to Goodhart effects. \\
\bottomrule
\end{tabular}
}
\end{table*}

\paragraph{Coordination}
In multi-agent or modular systems, coordination measures whether components maintain a shared plan, minimize redundant work, and resolve dependencies. Common signals include team success rate, communication efficiency (messages/tokens and rounds-to-consensus), plan/belief consistency, and deadlock/conflict rate. We additionally report: (i) a \emph{Component Synergy Score (CSS)} to credit enabling contributions across agents, and (ii) \emph{Tool Utilization Efficacy (TUE)} to assess correctness and efficiency of tool calls.

\paragraph{Composite Reporting}
Composite metrics are often presented as a weighted scalar summary across dimensions:
\begin{equation}
  M_{\text{comp}} \;=\; \omega_T M_T + \omega_E M_E + \omega_U M_U + \omega_{Co} M_{Co}\,,
  \label{eq:composite}
\end{equation}
where $M_T,M_E,M_U,M_{Co}\in[0,1]$ are normalized scores for trustworthiness, explainability, user-centered performance, and coordination, respectively, and $\omega_\bullet \ge 0$ with $\sum \omega_\bullet=1$ encode domain priorities. Since scalarization can hide critical failures, we recommend always reporting the full metric vector:
\begin{equation}
  \mathbf{M} \;=\; (M_T, M_E, M_U, M_{Co})\,.
  \label{eq:metric_vector}
\end{equation}
Model selection should then be based on (i) minimum thresholds (e.g., $M_T \ge \tau_T$ and $M_{Co} \ge \tau_{Co}$) and/or (ii) Pareto-optimality (non-dominated models) rather than relying only on a single weighted score.

\paragraph{Component Synergy Score (CSS)}
\textbf{Definition:} CSS quantifies the quality of inter-agent collaboration by measuring how effectively one agent’s actions enable or enhance a peer’s downstream performance.

\textbf{Formal expression:}
\begin{equation}
\text{CSS} = \frac{1}{N_{\text{pairs}}}\sum_{i}\sum_{j \neq i}
\Big[\text{Impact}(a_i \!\rightarrow\! a_j)\cdot \text{Quality}(a_j \mid a_i)\Big],
\end{equation}
where $N_{\text{pairs}}$ is the number of ordered agent pairs $(i,j)$ included in the evaluation. $\text{Impact}(a_i \!\rightarrow\! a_j)$ measures how $a_i$ changes $a_j$’s subsequent performance (e.g., relative improvement), and $\text{Quality}(a_j \mid a_i)$ is the quality of $a_j$’s output conditioned on $a_i$’s contribution.

\textbf{Illustrative example:} In a software AMAS with \textit{Planner}, \textit{Coder}, and \textit{Tester}:
\begin{itemize}[leftmargin=*]
  \item Planner’s decomposition speeds up Coder by 35\% ($\text{Impact}=0.35$).
  \item Tester achieves 92\% coverage conditioned on Coder’s output ($\text{Quality}=0.92$).
\end{itemize}
If we score only this link, then $\text{CSS} = 0.35 \times 0.92 \approx 0.32$ (and increases as additional enabling links are included).

\paragraph{Tool Utilization Efficacy (TUE)}
\textbf{Definition:} TUE evaluates how correctly and efficiently agents invoke external tools (APIs, databases, code interpreters) within their workflows.

\textbf{Components:}
\begin{itemize}[leftmargin=*]
  \item $\text{Sel}\in[0,1]$ --- Selection (decision precision/recall for tool use)
  \item $\text{Arg}\in[0,1]$ --- Argument validity (well-formed parameters)
  \item $\text{Exec}\in[0,1]$ --- Execution success (no errors/timeouts)
  \item $\text{Out}\in[0,1]$ --- Outcome correctness (downstream task correctness)
  \item $\text{Eff}\in[0,1]$ --- Efficiency (normalized inverse of latency/token/call cost)
\end{itemize}

\textbf{Linear aggregator:}
\begin{equation}
\text{TUE}_{\text{lin}} =
\alpha_s\text{Sel} + \alpha_a\text{Arg} + \alpha_e\text{Exec} + \alpha_o\text{Out} + \alpha_f\text{Eff},
\end{equation}
where $\alpha_\bullet \ge 0$ and $\sum \alpha_\bullet = 1$.

\paragraph{Key Insights}
This section defined a multi-dimensional evaluation framework covering trustworthiness (robustness, violation rates, calibration), explainability (coverage, faithfulness, stability), user-centered performance (satisfaction and goal fulfillment), and coordination quality. CSS captures positive inter-agent enablement, while TUE decomposes tool use into selection, argument validity, execution success, outcome correctness, and efficiency. Finally, the framework emphasizes reporting a metric vector and using thresholding/Pareto analysis to reduce the risk that scalar composites mask critical single-dimension failures.

\section{Security and Privacy in Agentic AI Systems}
\label{security}

Evaluating the security and robustness of AMAS requires standardized benchmarks and well-defined metrics to assess both attack vulnerabilities and defense effectiveness. This subsection reviews current evaluation frameworks and security metrics for agentic AI, aligned with the TRiSM pillars.

\subsection{Security Mechanisms}
Agentic AI systems (multiple collaborating LLM agents) have a bigger attack surface than traditional agents, so they need layered defenses: protecting data, execution, communication, and model robustness; using encryption, access control, adversarial defenses, and runtime monitoring tailored to multi-agent setups.

\textbf{Encryption.} Encryption plays an important role in safeguarding data exchanged between multiple agents, especially when sensitive or regulated content (e.g., healthcare records, financial data) is involved \cite{feretzakis2024privacy}. Agentic workflows often include the inter-agent handoff of partially processed results, models, or prompts. Implementations such as SSL/TLS, homomorphic encryption , and secure enclaves (e.g., Intel SGX) are increasingly integrated into Agentic AI pipelines to ensure confidentiality across message-passing protocols.

\textbf{Access control.} Access control is highly important when orchestrators or shared memory modules manage permissions for agents with distinct capabilities and responsibilities. For instance, in systems like AutoGen and CrewAI where agents take on specialized roles (e.g., summarizer, planner, coder); enforcing \textit{principle-of-least-privilege} access prevents privilege escalation and unauthorized tool invocation \cite{wu2023autogen}. Agent-based access control policies, often aligned with Role-Based Access Control (RBAC) \cite{khan2024role} and Attribute-Based Access Control (ABAC) \cite{khan2024role} paradigms, can dynamically restrict which agents may access sensitive APIs, files, or memory buffers, based on contextual trust levels.

\textbf{Adversarial learning.} Adversarial threats are increasing for LLM-based agents, including prompt injection, poisoned tool outputs, and disruption via malformed intermediate results. Recent work shows that compromising one agent can propagate misleading outputs and destabilize an entire multi-agent framework \cite{zhang2024breaking}. Mitigations include adversarial training (e.g., input perturbation, reward shaping, contrastive learning) and practical safeguards such as enforcing safety constraints and validating tool outputs before execution.

\textbf{Runtime Monitoring.} Runtime monitoring systems support the detection of anomalous agent behaviors, especially in high-stakes domains like automated healthcare or cybersecurity. Log-based auditing, anomaly detection with LSTM or autoencoder-based detectors, and trust scoring among agents are becoming essential components of real-time surveillance layers \cite{zhen2025novel}. For example, Microsoft’s Copilot governance layers monitor anomalous agent behavior across sessions to ensure compliant execution and flag potentially harmful interactions.

 \subsection{Privacy-Preserving Techniques}
The decentralized and interactive nature of AMAS introduces new challenges for preserving privacy, especially as agents continuously communicate, access external data sources, and store episodic or shared memory. To ensure data confidentiality and protect personally identifiable information (PII), Agentic AI systems must adopt robust privacy-preserving techniques such as \textit{differential privacy}, \textit{data minimization}, and \textit{secure computation}.

\textbf{Differential privacy (DP).} DP protects individuals by adding controlled noise, and in multi-agent LLM systems it can be used during training or inference to limit what agents can reveal to each other. Using approaches like DP in federated learning, DP-SGD, and an $\epsilon$ privacy budget, agents can collaboratively train, fine-tune, or coordinate while reducing exposure of raw or identifiable data \cite{lazaros2024federated}.

\textbf{Data minimization.} Agentic AI systems can mitigate exposure risks by limiting the scope, granularity, and duration of data collected or retained during task execution. For instance, temporary memory buffers used in systems like ChatDev \cite{qian2023chatdev} or ReAct-based pipelines \cite{yao2023reactsynergizingreasoningacting} are cleared once subgoals are completed, preventing persistent storage of unnecessary user data. Furthermore, anonymization and pseudony \cite{luzon2024tutorial} techniques can help remove identifying features before data is passed between agents or stored in shared memory repositories.

\textbf{Secure computation.} Techniques including secure multi-party computation (SMPC) \cite{evans2018pragmatic}, homomorphic encryption \cite{acar2018survey}, and trusted execution environments (TEEs)\footnote{\href{https://en.wikipedia.org/wiki/Trusted_execution_environment}{Trusted Execution Environment (Wikipedia)}}
enable agents to perform computations over encrypted or obfuscated data without compromising privacy. In scenarios where agents collaborate across different organizational boundaries (e.g., federated medical agents or cross-silo industrial agents), SMPC allows joint computations such as diagnostics or anomaly detection without data leakage. Homomorphic encryption, while computationally expensive, is increasingly being explored to allow arithmetic operations on encrypted vectors used in RAG workflows.

Privacy-by-design \cite{schaar2010privacy} principles are becoming central to the engineering of next-generation Agentic AI systems. Architectures now embed user consent layers, configurable privacy settings, and memory redaction modules that allow end-users or system administrators to control what agents can remember or share. As Agentic AI expands into domains such as personalized education, healthcare, and finance, ensuring privacy-respecting behaviors will be essential for regulatory compliance (e.g., GDPR, HIPAA) and public trust.

{\scriptsize
\begin{longtable}{p{0.1\textwidth} p{0.14\textwidth} p{0.25\textwidth} p{0.18\textwidth} p{0.18\textwidth}}
\caption{Governance controls for agentic AI systems: pillar $\rightarrow$ control $\rightarrow$ operationalization $\rightarrow$ evidence $\rightarrow$ standards.}
\label{tab:agentic_governance}\\
\toprule
\textbf{Pillar} & \textbf{Control} & \textbf{Operationalization (How)} & \textbf{Evidence / Artifacts} & \textbf{Standards \& Articles} \\
\midrule
\endfirsthead

\toprule
\textbf{Pillar} & \textbf{Control} & \textbf{Operationalization (How)} & \textbf{Evidence / Artifacts} & \textbf{Standards \& Articles} \\
\midrule
\endhead

\midrule
\multicolumn{5}{r}{\emph{Continued on next page}}\\
\midrule
\endfoot

\bottomrule
\endlastfoot

Governance & Org. accountability & Define owners for models/agents/tools; RACI for changes; approvals for high-impact actions & Governance policy; RACI records; change approvals & NIST AI RMF (Govern)~\cite{tabassi2023ai}; ISO/IEC~42001~\cite{iso42001} \\

Governance & Policy-as-code & Encode allow/deny rules in orchestrator; pre-exec checks and human sign-off gates & Policy repository; policy test cases; gate logs & EU AI Act Arts.~11–14~\cite{eu_ai_act}; ISO/IEC~42001~\cite{iso42001} \\

Risk Mgmt & Risk register \& AIA & Hazard/threat enumeration; likelihood/impact scoring; AI Impact Assessment (AIA) & Risk register; AIA reports; mitigation plan & ISO/IEC~23894; ISO/IEC~42005~\cite{ISO_IEC_42005_2025}; EU AI Act Art.~9~\cite{eu_ai_act} \\

Risk Mgmt & Post-market monitoring & Drift/jailbreak telemetry; incident thresholds; rollback playbooks & PMM plan; incident tickets; root-cause analysis & EU AI Act Arts.~72–73~\cite{eu_ai_act}; NIST AI RMF (Manage)~\cite{tabassi2023ai} \\

Data Gov & Data minimization \& lineage & Task/tenant-scoped memory; TTLs/redaction; dataset lineage and licenses & Data inventory; lineage graphs; retention logs & EU GDPR Art.~25~\cite{gdpr25}; ISO/IEC~5259 series \\

Data Gov & Quality \& bias checks & Coverage/balance checks; label noise audits; sampling documentation & Data quality reports; bias audit results & EU AI Act Art.~10~\cite{eu_ai_act}; ISO/IEC TR~24027 \\

Explainability & Faithfulness/stability & Log rationales; test explanation faithfulness/stability; user-facing summaries & XAI eval report; model/data/agent cards & NISTIR~8312~\cite{phillips2021four}; NIST AI RMF (Map/Measure)~\cite{tabassi2023ai} \\

Documentation & Technical documentation & Annex-IV tech docs: purpose, data, performance, oversight, foreseeable misuse & Annex-IV dossier; limitations \& misuse notes & EU AI Act Arts.~11–12; Annex~IV~\cite{eu_ai_act} \\

Traceability & Provenance \& logging & Log prompts, plans, tool calls, I/O, timestamps, agent role; sign releases & Immutable logs (WORM/tamper-evident); signed artifacts & EU AI Act Arts.~12–13~\cite{eu_ai_act}; ISO/IEC~42001~\cite{iso42001} \\

Human Oversight & Human-in-the-loop gates & Require human approval for sensitive plans (delete, external write, PII) & Approval records; override/rollback logs & EU AI Act Art.~14~\cite{eu_ai_act}; NIST AI RMF (Manage)~\cite{tabassi2023ai} \\

Robustness & Stress/red-team testing & Adversarial prompts; perturbation/fault tests; safety regression suites & Red-team report; robustness curves; residual risk & EU AI Act Art.~15~\cite{eu_ai_act}; ISO/IEC~24029-1~\cite{iso24029-1} \\

Security & I/O mediation & Prompt sanitization; output filtering; sensitive-data (PII/PHI/secrets) detection & Filter policies; violation/egress logs & OWASP LLM Top-10~\cite{owasp2024llmtop10}; HIPAA~\S164~\cite{hipaa164} \\

Security & Tool access control & RBAC/ABAC per agent/role; per-tool API tokens; argument validation; allowlists & Access matrices; token scopes; arg-validator tests & NIST SP~800-218/218A; NIST AI RMF (Manage) \\

Security & Isolation/sandboxing & Containers/VMs; syscall/network/file egress policies; hardware TEEs for sensitive code & Sandbox configs; egress policy; attestation logs & NIST SP~800-218A; TEEs best-practice notes \\

Privacy & DP budgeting \& consent & Differential privacy budgets; consent bases; data-subject rights flows & DP budget ledger; DSR tickets; consent records & GDPR Art.~25~\cite{gdpr25}; OECD principles~\cite{oecd_ai_catalogue} \\

Supply Chain & Provenance \& SBOM & Track sources and hashes for models/tools/datasets; dependency scanning & SBOM; third-party attestations; license checks & NIST SSDF~(SP~800-218/218A); ISO/IEC~42001 \\

Coordination & Role separation \& least privilege & Distinct agent roles (planner, coder, reviewer); minimal cross-role permissions & Role definitions; privilege matrices; audit samples & NIST AI RMF (Govern)~\cite{tabassi2023ai}; ISO/IEC~42001 \\

Coordination & Plan consistency checks & Shared plan/belief verification; conflict/deadlock detection; CSS metric & Plan snapshots; CSS scores; conflict logs & Multi-agent eval practice; robustness~\cite{iso24029-1} \\

Monitoring & Anomaly \& exfil detection & Detect unusual tool chains, volume spikes, strange domains; rate-limit/kill switch & Anomaly dashboards; blocked-egress records & OWASP LLM Top-10~\cite{owasp2024llmtop10}; NIST AI RMF (Manage) \\

Transparency & User notices \& recourse & User-facing disclosures, capabilities/limits, and appeal/feedback channels & Notices; feedback logs; SLA for appeals & OECD principles~\cite{oecd_ai_catalogue}; EU AI Act transparency \\

Evaluation & Fairness \& calibration & Group metrics; ECE/Brier; CI reporting; risk–benefit trade-off docs & Fairness report; calibration plots; CI tables & NIST AI RMF (Measure)~\cite{tabassi2023ai}; ISO/IEC TR~24027 \\

Records & Release/rollback discipline & Versioned releases; rollback points; emergency disable/recall & Release notes; rollback attestations & EU AI Act PMM~\cite{eu_ai_act}; ISO/IEC~42001 \\

GPAI & GPAI obligations & Provide model cards, training-data summary, eval reports, and usage restrictions; disclose known limitations & GPAI documentation pack; model card; eval report & EU AI Act (GPAI duties)~\cite{eu_ai_act}; ISO/IEC~42001 \\

Transparency & Synthetic content disclosure & Label AI-generated/edited media; attach provenance (e.g., C2PA) and user notices & Watermarking/provenance config; disclosure logs & EU AI Act Art.~52~\cite{eu_ai_act}; C2PA\footnote{\href{https://c2pa.org}{c2pa.org}} \\

Privacy & Cross-border transfers & SCCs/TIAs for third-country transfers; data-residency controls; key management & SCC/TIA records; DPA; data-map; key-custody logs & GDPR Ch.~V~\cite{gdpr25} \\

Privacy & Data-subject rights ops & Verified workflows for access, erasure, rectification, restriction, portability, objection & DSR tickets/SLAs; fulfillment proofs; audit samples & GDPR Arts.~12–23~\cite{gdpr25} \\

Operations & Change mgmt \& release gates & Risk-based pre-release checklist; dual-control approvals; staged rollouts; rollback criteria & Change tickets; gate results; approvals; rollback points & ISO/IEC~42001; NIST AI RMF (Manage)~\cite{tabassi2023ai} \\

Supply Chain & Third-party/vendor risk & Vendor questionnaires; DPAs; penetration/attestation reports; license compliance & Vendor risk register; DPA file; attestation/SBOM & NIST SSDF (SP~800-218/218A); ISO/IEC~42001 \\

Human Factors & HCD \& accessibility & Human-centred design reviews; usability risk analysis; accessibility conformance checks & HCD review notes; usability test reports; a11y checklists & ISO~9241-210\footnote{\href{https://www.iso.org/standard/77520.html}{ISO 9241-210}}; OECD principles~\cite{oecd_ai_catalogue} \\

Sustainability & Energy/CO$_2$ telemetry & Track/normalize energy use and CO$_2$; report per release; set budgets & Energy/CO$_2$ logs; budget vs. actual; disclosure note & Org policy / reporting guidance
\end{longtable}

}

\subsection{Evaluation Benchmarks and Metrics for Agentic AI Security}
\label{sec:eval-security}
Evaluating the security and robustness of AMAS requires standardized benchmarks and well-defined metrics to assess both attack vulnerabilities and defense effectiveness. This subsection reviews current evaluation frameworks and metrics related to agentic AI security, aligned with the TRiSM pillars.

\paragraph{Attack Evaluation Benchmarks}

Several benchmark datasets have emerged to evaluate adversarial attacks against LLM-based agents. For prompt injection and jailbreak scenarios, HarmBench~\cite{mazeika2024harmbench} and JailbreakBench~\cite{chao2024jailbreakbench} provide standardized datasets for evaluating robustness against adversarial prompts, including role-playing, obfuscation, and multi-turn manipulation scenarios particularly relevant to agentic systems. Tool misuse is addressed by ToolBench~\cite{xu2023tool} and API-Bank~\cite{li2023apibank}, which evaluate agents' ability to use external tools correctly while resisting adversarial instructions that trigger unauthorized access or malicious function calls.

Multi-agent adversarial scenarios are covered by AgentBench~\cite{liu2023agentbench} and GAIA~\cite{mialon2023gaia}, which test resilience against coordinated attacks, impersonation, and information manipulation in multi-agent environments. Recent datasets also evaluate vulnerabilities specific to agentic architectures, including memory poisoning attacks where adversaries inject malicious information into agent memory stores, and context manipulation where attack payloads are embedded in retrieved documents or tool outputs \cite{raza2025justhumansneedvaccines}.

\paragraph{Defense Evaluation Metrics}
Standard metrics for assessing defense mechanisms have been established across the research community. Attack Success Rate (ASR) measures the percentage of attack attempts that successfully compromise agent behavior, bypass safety constraints, or extract sensitive information, with lower ASR indicating more robust defenses \cite{saha2025breaking}. Robustness Score quantifies the agent's ability to maintain intended behavior under adversarial conditions, typically measured as the degradation in task performance when subjected to attacks compared to benign scenarios \cite{raza2025humanibench}.

Defense Effectiveness Rate captures the proportion of attacks successfully detected and mitigated by defense mechanisms, including input validation, output filtering, and behavioral monitoring systems. For production deployments, False Positive and False Negative Rates are critical, measuring the rate at which legitimate user requests are incorrectly flagged as attacks and actual attacks that evade detection, respectively. Our proposed Composite Security Score (CSS), introduced in Section~\ref{evals} provides a holistic assessment by combining multiple security dimensions including input validation coverage, tool access control effectiveness, memory integrity, and inter-agent authentication strength.

Table~\ref{tab:benchmarks} provides a systematic comparison of these benchmarks, including their coverage of attack scenarios, applicable defense metrics, and alignment with TRiSM pillars.

\begin{table}[htbp]
\centering
\scriptsize
\caption{Evaluation Benchmarks for Agentic AI Security}
\label{tab:benchmarks}
\begin{tabular}{p{2cm}p{2.5cm}p{2cm}p{2cm}p{3cm}}
\toprule
\textbf{Benchmark} & \textbf{Attack Coverage} & \textbf{Metrics} & \textbf{TRiSM Pillars} & \textbf{Limitations} \\
\midrule
HarmBench \cite{mazeika2024harmbench} & Prompt injection, jailbreaks & ASR, robustness & Security, Explainability & Limited multi-agent \\
\midrule
ToolBench \cite{xu2023tool} & Tool misuse, unauthorized API & Effectiveness, FP/FN & Security, Privacy & Tool-focused only \\
\midrule
AgentBench  \cite{liu2023agentbench}& Multi-step attacks, coordination & Task success, compliance & All pillars & Limited adversarial \\
\midrule
GAIA \cite{mialon2023gaia}& Ambiguous instructions, reasoning & Completion, safety & Explainability, Governance & Early stage \\
\midrule
WebArena \cite{zhou2023webarena} & Web attacks, malicious content & Success, safety & Security, Privacy & Domain-specific \\
\midrule
HELM \cite{liang2022helm} & Robustness, fairness, bias & Comprehensive & ModelOps, Governance & Not agent-specific \\
\midrule
MLCommons \cite{mlcommons2024} & Harmful content, privacy leakage & Safety, privacy scores & Security, Privacy, Governance & Evolving \\
\bottomrule
\end{tabular}
\end{table}

\paragraph{Key Insights}
This section presented a defense-in-depth view of AMAS security spanning data confidentiality (encryption and secure enclaves), access control (RBAC/ABAC with least privilege and dynamic trust), adversarial resilience (prompt/tool-output validation and training-based hardening), and runtime monitoring (auditing, anomaly detection, and trust scoring). Privacy protection was framed as a system property: differential privacy to limit re-identification risk, data minimization through scoped/ephemeral memory and pseudonymization, and secure computation (SMPC, homomorphic encryption, and TEEs) to enable collaboration without raw-data exposure. We also summarized emerging evaluation practice, such as benchmarks for prompt injection, tool misuse, and multi-agent manipulation, together with metrics such as ASR, robustness degradation, defense effectiveness, and FP/FN rates; to quantify both vulnerabilities and mitigation quality in a TRiSM-aligned manner. Overall, security and privacy shift from optional add-ons to foundational requirements for regulated deployments.

\section{Compliance with AI Governance and Policy Frameworks for Agentic AI Systems}

\label{compliance}
Recent industry surveys indicate that enterprise leaders deploying Agentic AI at scale rank security, trust, compliance, and oversight among their top concerns. 
A survey\footnote{\href{https://www.techradar.com/computing/artificial-intelligence/love-and-hate-tech-pros-overwhelmingly-like-ai-agents-but-view-them-as-a-growing-security-risk}{TechRadar (2024)}}  of IT professionals found that while 98\% plan to expand AI agent usage, 96\% view them as growing security threats, with only 54\% having full visibility into their data access. 
In parallel, 67\% of business leaders are increasing budgets for security oversight\footnote{\href{https://www.cybersecuritydive.com/news/artificial-intelligence-security-spending-reports/751685/}{Cybersecurity Dive (2024)}}, 
and 53\% specifically cite data privacy and compliance as their primary challenge in scaling Agentic AI\footnote{\href{https://www.cloudera.com/blog/business/ready-to-scale-tackling-the-top-challenges-of-agentic-ai-adoption.html}{Cloudera (2024)}}.
Table~\ref{tab:agentic_governance} operationalizes the governance controls into concrete implementations and evidence and next we detail the compliance to AI goverace and policy in Agentic AI systems.

\subsection{Regulatory Standards}
Regulatory frameworks provide baseline requirements that all AI systems, including agent-based architectures, must meet. Key frameworks include:
\begin{itemize}[leftmargin=*, nosep]
    \item \textit{NIST AI Risk Management Framework (AI~RMF)} \cite{tabassi2023ai}. A voluntary U.S. framework organized around the \emph{Govern, Map, Measure, Manage} functions. It emphasizes organizational governance, risk identification and measurement, and operational risk treatments. Mapping to the RMF also benefits from NISTIR~8312 (Explainable AI principles) for transparency obligations.
    \item \textit{EU AI Act} \cite{eu_ai_act}. A comprehensive EU regulation with phased application: bans on certain practices apply after six months (Feb~2025), obligations for general-purpose AI begin Aug~2025, and most high-risk system obligations (e.g., risk management, documentation, human oversight, post-market monitoring) start Aug~2026. High-risk systems require ongoing risk assessment, technical documentation (Annex~IV), logging/traceability (Arts.~11–12), data governance (Art.~10), human oversight (Art.~14), and robustness/accuracy (Art.~15).
    \item \textit{ISO/IEC management and assessment standards.} ISO/IEC~42001:2023 (AI management systems) sets organization-level governance requirements; ISO/IEC~42005:2025 (AI impact assessment) provides system-level impact-assessment guidance; ISO/IEC~23894 addresses AI risk management; ISO/IEC~24029-1 supports robustness evaluation; ISO/IEC TR~24027 addresses bias. These align well with AI RMF and the EU AI Act’s lifecycle duties.
    \item \textit{Domain-specific laws.} Sectoral/privacy regimes still apply: GDPR (e.g., data protection by design, Art.~25), HIPAA (45~CFR~\S164), and regional data-residency rules. Agentic systems should enforce data minimization, purpose limitation, and consent bases; consider \emph{DPIA} for high-risk processing and \emph{AI impact assessments} per ISO/IEC~42005 in parallel.
\end{itemize}

\subsection{Auditability}
AMAS can produce emergent and opaque behaviors; auditors must reconstruct the chain of decisions.
\begin{itemize}[leftmargin=*, nosep]
    \item \textit{Comprehensive logging.} Record prompts/contexts, plans, actions, tool calls, inputs/outputs, timestamps, and agent role/identity; capture rationales where feasible. Preserve hashes and version IDs for models, tools, and data snapshots.
    \item \textit{Action traceability.} Maintain decision provenance across agents (planner$\rightarrow$coder$\rightarrow$tester, etc.), including hand-offs and approvals. Link every external effect (e.g., file write, API call) to the proposing and approving principals.
    \item \textit{Role-granular trails.} Tag entries by agent role (e.g., Researcher, Coder, Reviewer) to localize anomalies and bias. Require reviewer countersignatures for high-impact actions.
    \item \textit{Immutability.} Use append-only, tamper-evident storage (e.g., WORM buckets, cryptographic sealing) for audit logs; rotate keys and attest integrity regularly.
\end{itemize}

\subsection{Policy Enforcement}
While regulations and audits provide oversight and after-the-fact analysis, policy enforcement mechanisms work in real-time to keep an Agentic AI system’s behavior within allowed bounds. Policy in this context refers to the codified rules and constraints that the AI agents must obey. In practice, effective policy enforcement in agentic AI includes:

\begin{itemize}[leftmargin=*, nosep]
    \item \textit{Orchestrator-level controls.} A meta-controller enforces allow/deny policies for plans and tool calls; escalate to human approval for sensitive operations (e.g., data deletion, external writes).
    \item \textit{Memory \& retention.} Implement TTLs, redaction, and scoped memory (task- and tenant-scoped) to satisfy minimization and purpose limits; support subject-access and deletion.
    \item \textit{Tool \& data access.} Enforce RBAC/ABAC for agents/roles; use per-tool API tokens, least privilege, and environment scoping (dev/stage/prod). Validate/normalize tool arguments.
    \item \textit{Isolation.} Run untrusted code/tooling in sandboxes (e.g., containers/VMs) with syscall/network/file egress policies; prefer hardware isolation (TEEs) for sensitive workloads.
    \item \textit{I/O mediation.} Apply content filters and egress controls on model outputs; mediate RAG inputs to prevent injection/exfiltration; detect sensitive data (PII/PHI, secrets) before release.
    \item \textit{Real-time monitoring.} Track anomaly signals (policy violations, unusual tool chains, data exfiltration patterns) and enforce runtime kill-switches, rollbacks, and incident triggers.
\end{itemize}

\subsection{Documentation \& Record-Keeping (Evidence)}
To meet audit and regulator expectations, maintain:
\begin{itemize}[leftmargin=*, nosep]
    \item \textit{Technical documentation} (EU AI Act \cite{EU_AI_Act2025}): system purpose and risk class, training/eval data governance, performance, limitations, foreseeable misuse, human oversight measures.
    \item \textit{Model/Data cards} with lineage, licenses, and known hazards; \textit{evaluation reports} with stress tests (robustness, red-teaming), calibration/fairness metrics, and residual risk.
    \item \textit{Versioned artifacts:} model versions, datasets, prompts, agents’ policies, and tool catalogs with change logs; signed releases and rollback points.
\end{itemize}

\subsection{Post-Market Monitoring \& Incident Response}
Agentic systems must be monitored after deployment and respond to harm quickly.
\begin{itemize}[leftmargin=*, nosep]
    \item \textit{Telemetry and drift.} Continuously measure real-world performance, jailbreak/abuse rates, and data distribution shifts; trigger retraining or guard updates.
    \item \textit{Serious-incident handling.} Define thresholds, triage workflows, and reporting timelines; keep case files linking logs, affected users, and remediation.
    \item \textit{Periodic reassessment.} Re-run impact/risk assessments (ISO/IEC~42005 \cite{ISO_IEC_42005_2025}, ISO/IEC~23894 \cite{ISOIEC23894_2023} ) when capabilities, data, or context change materially.
\end{itemize}

\subsection{Supply-Chain \& Secure Development (GenAI)}
Strengthen provenance and development hygiene for agents and tools.
\begin{itemize}[leftmargin=*, nosep]
    \item \textit{Secure SDLC for GenAI.} Follow NIST SP~800-218 (SSDF)\footnote{\href{https://csrc.nist.gov/pubs/sp/800/218/final}{NIST SP~800-218 (SSDF)}} and the GenAI profile SP~800-218A \footnote{\href{https://csrc.nist.gov/pubs/sp/800/218/a/ipd}{GenAI profile SP~800-218A}} for model/system development; integrate threat modeling for agent tool-chains and LLM-specific risks.
    \item \textit{OWASP LLM risks.} Mitigate prompt injection, insecure output handling, data poisoning, model DoS, and supply-chain risks using the OWASP \cite{owasp2024llmtop10} Top~10 for LLM applications.
    \item \textit{Provenance.} Track model/tool/dataset origins, licenses, and checksums; require attestations for third-party components and run dependency/secret scans in CI.
\end{itemize}

Bringing Agentic systems into compliance is not a single checklist; it is a lifecycle program that couples standards (NIST AI~RMF, ISO/IEC~42001/42005), regulation (EU AI Act), secure development (NIST SP~800-218/218A), and runtime enforcement (policies, isolation, monitoring). This combination creates evidenceable trust and reduces the blast radius of inevitable failures.

\paragraph{Key Insights}
This section operationalized compliance through systematic mappings between TRiSM pillars and regulatory instruments (EU AI Act, NIST AI RMF, ISO/IEC 42001, GDPR, HIPAA). Comprehensive auditability via immutable logging and decision provenance graphs enables regulatory audit and incident investigation. Policy-as-code mechanisms encode runtime constraints, preventing violations through pre-execution validation and human sign-off requirements for high-impact operations. Documentation rigor (technical dossiers, model cards, evaluation reports) provides regulator-ready evidence packages, while post-market monitoring implements continuous telemetry, incident handling workflows, and periodic reassessment protocols. Supply chain security addresses provenance verification through Software Bill of Materials documentation and dependency scanning. This integrated lifecycle program transforms compliance from checkbox exercise to strategic capability, building evidenceable trust through systematic controls and adaptive risk management.

\section{Discussion}
\label{discussion}
In this section, we summarize key findings and outline implications and future directions for deploying TRiSM in AMAS.

\subsection{Technical Implications of TRiSM for Agentic AI Design}
Rather than treating LLM agents as black-box decision-makers, TRiSM encourages designing agents with continuous oversight guardrails~\cite{guardrails_ai_guardrails_2024}. Recent discussions also highlight the value of specialized ``guardian agents'' within AMAS~\cite{securiti2025aitrism}, which can filter sensitive data, establish baselines of normal behavior, and enforce runtime policies (e.g., blocking disallowed actions such as exposing personally identifiable information). Prior work also warns about ``excessive agency''~\cite{owasp2025agentic}, where overly autonomous agents with broad tool access may trigger unintended harmful actions (e.g., hallucination-driven execution or misinterpreted goals). TRiSM mitigates these risks by limiting autonomy through role separation, scoped permissions, and safety constraints.  Emerging AMAS-specific threats such as prompt injection, memory poisoning, and cascading hallucinations further motivate built-in risk controls.

\subsection{Ethical and Societal Ramifications}
Beyond technical risks, deploying networks of autonomous agents raises ethical and societal concerns around accountability, fairness, and human oversight~\cite{raza2025responsible}. TRiSM governance emphasizes that organizations must retain responsibility for AI actions and avoid obscuring accountability behind opaque decision-making. In practice, each autonomous agentic decisions should be transparent enough to be understood and challenged by human reviewers when necessary~\cite{vayani2025all}. TRiSM does not remove humans from the loop; instead, it structures human-agent collaboration through defined oversight roles and escalation pathways~\cite{raza2025humanibench}. In AMAS deployments, humans may monitor agent swarms in real time, pause or shut down anomalous behavior, and adjust policies on the fly. The risk of user complacency, over-trusting autonomous agents; has also been noted as a hazard~\cite{parasuraman2010complacency}, and TRiSM counteracts this by formalizing oversight and supervision requirements.

\subsection{Critical Analysis of Evaluation Approaches}
\label{sec:critical-analysis}
The evaluation framework in Sections~\ref{evals} and~\ref{sec:eval-security} supports targeted diagnosis by separating trustworthiness, explainability, user-centered performance, and coordination metrics, rather than relying on monolithic accuracy~\cite{raza2024exploring}. The proposed CSS and TUE further address gaps in capturing multi-agent phenomena that single-agent metrics cannot represent~\cite{ribeiro2016should,lundberg2017unified}.
However, several limitations remain. Human-centered evaluation is resource-intensive and does not scale to continuous deployment, and AMAS evaluation often lacks ground truth for emergent coordination behaviors. Improvements in one dimension may degrade another (e.g., verbose explanations increasing latency), and composite weights vary by domain, requiring clearer guidance for practitioners. Current protocols also struggle to capture behavioral drift over time, and subjective measures often omit inter-rater reliability. Finally, evaluation metrics may be gamed (Goodhart’s Law\footnote{\href{https://en.wikipedia.org/wiki/Goodhart\%27s_law}{Goodhart's law}.}), especially when systems optimize for a single composite score rather than underlying capability.

\subsection{Implementation Challenges and Real-World Constraints}
While TRiSM provides a strong foundation, deploying it in production introduces practical constraints. Continuous monitoring, validation, and anomaly detection add computational overhead and latency, which can harm user experience in time-sensitive settings. Large-scale enterprise deployments may involve hundreds or thousands of agents, increasing compute, storage, networking, and staffing costs.
Integration is also challenging because many organizations rely on legacy infrastructure or third-party frameworks not designed for agentic security; adding memory protection, access control, and behavioral monitoring often requires cross-component changes and ongoing maintenance. Operationally, monitoring can produce large volumes of alerts, including false positives, which may overwhelm security teams. A phased, risk-driven adoption strategy is therefore often more feasible: start with input validation, restricted tool access, and logging, and then add stronger controls as system maturity and risk exposure increase.

\subsection{Alignment with Emerging AI Regulations and Standards}
TRiSM principles align closely with emerging regulatory frameworks and standards. For example, the EU AI Act~\cite{eu_ai_act} emphasizes risk management, transparency, data governance, and human oversight for high-risk systems, for example capabilities directly supported by TRiSM-aligned governance and controls \cite{yang2025agentic,farooq2025evaluating}. This convergence suggests that TRiSM-based practices can help organizations translate technical safeguards into compliance-ready processes and evidence.

Despite progress, some gaps remain. For example, there is still no consensus on standardized benchmarks and metrics for evaluating agentic systems under TRiSM principles, which limits cross-study comparison and objective progress tracking. Many controls also lack real-world validation beyond lab settings or narrowly scoped tasks, and adversarial learning remains an ongoing challenge as attackers adapt to defenses. Future work should prioritize standardized stress-testing protocols (e.g., red-teaming), longitudinal evaluation to detect drift, and deployment case studies that report both effectiveness and operational costs.

\textbf{Extension to Multimodal and Embodied Agents.} While this survey focuses on LLM-based AMAS, the Agentic AI landscape increasingly encompasses multimodal agents (e.g., vision-language models such as LLaVA, GPT-4V) and embodied agents \cite{xia2018gibson} operating in physical environments (e.g., robotic systems, autonomous vehicles, industrial automation). Future work should systematically extend TRiSM frameworks to address the unique challenges these systems present. Multimodal agents introduce additional security considerations including cross-modal adversarial attacks (e.g., adversarial images triggering unsafe text outputs), visual hallucinations, and sensor spoofing vulnerabilities. Embodied agents face physical safety risks, real-time decision-making constraints under uncertainty, sim-to-real transfer gaps, and complex human-robot trust calibration requirements. Extending the proposed CSS and TUE metrics to capture sensor fusion quality, physical action safety scores, and human-robot collaboration efficacy represents a promising direction for comprehensive agentic AI evaluation.
\subsection{Future Roadmap for Agentic AI TRiSM}

\begin{figure}[t]
\centering
\begin{tikzpicture}
  \path[mindmap,concept color=green!60!black,text=black]
    node[concept] {Agentic AI Roadmap}
    [clockwise from=0]
    child[concept color=blue!80!white] { node[concept] {Scalable Agent Architectures} }
    child[concept color=green!60!white] { node[concept] {Secure Multi-Agent Collaboration} }
    child[concept color=orange!80!white] { node[concept] {Explainable Multi-Agent Decisions} }
    child[concept color=purple!60!white] { node[concept] {Human-Centered Trust Design} }
    child[concept color=red!60!white] { node[concept] {Autonomous Lifecycle Management} }
    child[concept color=cyan!60!white] { 
      node[concept] {Ethics \& Governance}
        [clockwise from=300]
        child[concept color=cyan!30!white] { node[concept] {Controllability} }
    }
    child[concept color=yellow!80!white] { node[concept] {Benchmarking \& Evaluation} }
    child[concept color=pink!60!white] { node[concept] {Cognitive Capability Expansion} };
\end{tikzpicture}
\caption{Agentic AI Roadmap: Mindmap Representation. 
}
\label{fig:roadmap_mindmap}
\end{figure}

Drawing from our findings and best practices from multiple disciplines, we propose several actionable directions for future research and implementation, as shown in Figure \ref{fig:roadmap_mindmap}. These recommendations span both technical system design improvements and governance-level policy initiatives:

The community should create open benchmarks and challenge environments to test AMAS governance. For instance, a suite of scenario-based tasks (with built-in threats and ethical dilemmas) could be used to evaluate how well a TRiSM-governed ASAM that performs relative to one without such controls. This will enable direct comparisons and drive progress on measurable metrics of trust (e.g. frequency of prevented failures or fairness outcomes).

 Future system design must anticipate a continually evolving threat landscape. Techniques from cybersecurity (e.g. adversarial training, AI model “penetration testing” \cite{bacudio2011overview}, and formal verification) should be integrated into the LLM agent development pipeline. Cross-disciplinary collaboration with security experts can yield LLM-specific hardening methods, such as dynamic prompt anomaly detectors or robust tool APIs that constrain agent actions. Additionally, creating red-team/blue-team exercises for AMAS , akin to cyber wargames \cite{schneider2020wargaming} , can help discover vulnerabilities in a controlled way before real adversaries do.

We encourage designing better interfaces and protocols for human oversight of Agentic AI systems. Borrowing from human-computer interaction \cite{AmershiWVFNCSIB19} and cognitive engineering \cite{hexmoor2025behaviour}, researchers could devise dashboards that visualize an agent society’s state, flag important decisions, and allow intuitive human intervention (pausing agents, rolling back actions, etc.). 

Policymakers and industry should collaborate to create regulatory sandboxes for multi-agent AI trials. There is much to learn from other high-stakes domains. For example, the safety engineering field (e.g. aerospace, automotive) has mature practices for redundant controls and failure mode analysis; these could inspire analogous practices in AI agent design. Likewise, ethics boards in biomedical research provide a template for AI ethics committees that review agent behaviors and approve high-risk deployments. We advocate establishing multidisciplinary governance boards that include ethicists, legal experts, domain specialists, and community representatives to oversee significant deployments of autonomous AI.

\section{Conclusion}
\label{conclusion}
Agentic AI systems built as AMAS  are rapidly changing how autonomous systems collaborate, plan, and make decisions in high-stakes domains such as healthcare, finance, and public services. At the same time, these systems expand the risk surface and raise new challenges in trust, risk, and security management (TRiSM). This review addressed those challenges and introduced a TRiSM-oriented framework for LLM-based AMAS organized around five pillars: Explainability, ModelOps, Security, Privacy, and lifecycle Governance.
We contributed a risk taxonomy that captures AMAS-specific threats, including prompt injection, memory poisoning, and collusive or cascading failures. We also proposed practical evaluation metrics to measure inter-agent enablement and the correctness and efficiency of tool use. In addition, we summarized key technical mechanisms to support TRiSM in practice, including explainability methods and security/privacy protections.
Looking ahead, progress will depend on stronger adversarial robustness, clearer governance and audit protocols, and standardized benchmarks that evaluate both trustworthiness and coordination under realistic deployment conditions. Aligning Agentic AI development with TRiSM principles can help ensure these systems remain not only capable and scalable, but also safe, transparent, and accountable for real-world use.

\section*{Acknowledgments}

This research was funded by the European Union’s Horizon Europe research and innovation programme under the AIXPERT project (Grant Agreement No. 101214389), which aims to develop an agentic, multi-layered, GenAI-powered framework for creating explainable, accountable, and transparent AI systems. Additionally, this work was supported in part by the National Science Foundation (NSF); in part by United States Department of Agriculture (USDA); in part by the National Institute of Food and Agriculture (NIFA), through the “Artificial Intelligence (AI) Institute for Agriculture” Program, Accession Number 1029004 for the Project Titled “Robotic Blossom Thinning with Soft Manipulators” under Award AWD003473, Award AWD004595, and Award USDA-NIFA; and in part by United States Department of Agriculture National Science Foundation (USDANSF), Accession Number 1031712, under the Project “ExPanding University of Central Florida (UCF) AI Research To Novel Agricultural EngineeRing Applications (PARTNER)” under Grant 2024-67022-41788. 

\section*{Declaration of Competing Interest}
The authors declare that they have no known competing financial interests or personal relationships that could have appeared to influence the work reported in this paper.

\section*{Author Contributions}
\textbf{S.R.} and \textbf{R.S.} contributed equally to this work. 
\textbf{S.R.:} Conceptualization, Methodology, Literature review, Bibliographic analysis, Writing, Re-writing, Editing, Revision
\textbf{R.S:} Conceptualization, Methodology, Literature review, Bibliographic analysis, Writing, Re-writing, Editing
\textbf{M.K.:} Supervision, Detailed review, Structuring, Writing. 
\textbf{C.E.:} Detailed review,  Writing, Re-writing, Supervision, Revision.  
All authors have read and approved the final manuscript.

 \balance
\bibliographystyle{elsarticle-num}
\bibliography{references}

@article{yang2025agentic,
  title={Agentic web: Weaving the next web with ai agents},
  author={Yang, Yingxuan and Ma, Mulei and Huang, Yuxuan and Chai, Huacan and Gong, Chenyu and Geng, Haoran and Zhou, Yuanjian and Wen, Ying and Fang, Meng and Chen, Muhao and others},
  journal={arXiv preprint arXiv:2507.21206},
  year={2025}
}

@article{farooq2025evaluating,
  title={Evaluating and Regulating Agentic AI: A Study of Benchmarks, Metrics, and Regulation},
  author={Farooq, Azib and Raza, Shaina and Karim, Md Nazmul and Iqbal, Hasan and Vasilakos, Athanasios V and Emmanouilidis, Christos},
  journal={Metrics, and Regulation},
  year={2025}
}

@article{mazeika2024harmbench,
  title={HarmBench: A Standardized Evaluation Framework for Automated Red Teaming and Robust Refusal},
  author={Mazeika, Mantas and Phan, Long and Yin, Xuwang and Zou, Andy and Wang, Zifan and Mu, Norman and Sakhaee, Elham and Li, Nathaniel and Basart, Steven and Li, Bo and Forsyth, David and Hendrycks, Dan},
  journal={arXiv preprint arXiv:2402.04249},
  year={2024}
}

@article{chao2024jailbreakbench,
  title={JailbreakBench: An Open Robustness Benchmark for Jailbreaking Large Language Models},
  author={Chao, Patrick and Debenedetti, Edoardo and Robey, Alexander and Andriushchenko, Maksym and Croce, Francesco and Anand, Vikash and Hendrycks, Dan and Kolter, J. Zico},
  journal={arXiv preprint arXiv:2404.01318},
  year={2024}
}

@ARTICLE{11081880,
  author={Wang, Yuntao and Pan, Yanghe and Guo, Shaolong and Su, Zhou},
  journal={IEEE Open Journal of the Computer Society}, 
  title={Security of Internet of Agents: Attacks and Countermeasures}, 
  year={2025},
  volume={6},
  number={},
  pages={1611-1624},
  doi={10.1109/OJCS.2025.3589638}}

@article{saha2025breaking,
  title={Breaking the code: Security assessment of ai code agents through systematic jailbreaking attacks},
  author={Saha, Shoumik and Chen, Jifan and Mayers, Sam and Gouda, Sanjay Krishna and Wang, Zijian and Kumar, Varun},
  journal={arXiv preprint arXiv:2510.01359},
  year={2025}
}

@article{xu2023tool,
  title={On the tool manipulation capability of open-source large language models},
  author={Xu, Qiantong and Hong, Fenglu and Li, Bo and Hu, Changran and Chen, Zhengyu and Zhang, Jian},
  journal={arXiv preprint arXiv:2305.16504},
  year={2023}
}

@article{li2023apibank,
  title={API-Bank: A Comprehensive Benchmark for Tool-Augmented LLMs},
  author={Li, Minghao and Song, Feifan and Yu, Bowen and Yu, Haiyang and Li, Zhoujun and Huang, Fei and Li, Yongbin},
  journal={arXiv preprint arXiv:2304.08244},
  year={2023}
}

@article{liu2023agentbench,
  title={AgentBench: Evaluating LLMs as Agents},
  author={Liu, Xiao and Yu, Hao and Zhang, Hanchen and Xu, Yifan and Lei, Xuanyu and Lai, Hanyu and Gu, Yu and Ding, Hangliang and Men, Kaiwen and Yang, Kejuan and Zhang, Shudan and Deng, Xiang and Zeng, Aohan and Du, Zhengxiao and Zhang, Chenhui and Shen, Sheng and Zhang, Tianjun and Su, Yu and Sun, Huan and Huang, Minlie and Dong, Yuxiao and Tang, Jie},
  journal={arXiv preprint arXiv:2308.03688},
  year={2023}
}

@article{mialon2023gaia,
  title={GAIA: A Benchmark for General AI Assistants},
  author={Mialon, Gr{\'e}goire and Dess{\`\i}, Roberto and Lomeli, Maria and Nalmpantis, Christoforos and Pasunuru, Ram and Raileanu, Roberta and Rozi{\`e}re, Baptiste and Schick, Timo and Dwivedi-Yu, Jane and Celikyilmaz, Asli and Grave, Edouard and LeCun, Yann and Scialom, Thomas},
  journal={arXiv preprint arXiv:2311.12983},
  year={2023}
}

@article{zhou2023webarena,
  title={WebArena: A Realistic Web Environment for Building Autonomous Agents},
  author={Zhou, Shuyan and Xu, Frank F. and Zhu, Hao and Zhou, Xuhui and Lo, Robert and Sridhar, Abishek and Cheng, Xianyi and Bisk, Yonatan and Fried, Daniel and Alon, Uri and Neubig, Graham},
  journal={arXiv preprint arXiv:2307.13854},
  year={2023}
}

@article{liang2022helm,
  title={Holistic Evaluation of Language Models},
  author={Liang, Percy and Bommasani, Rishi and Lee, Tony and Tsipras, Dimitris and Soylu, Dilara and Yasunaga, Michihiro and Zhang, Yian and Narayanan, Deepak and Wu, Yuhuai and Kumar, Ananya and Newman, Benjamin and Yuan, Binhang and Yan, Bobby and Zhang, Ce and Cosgrove, Christian and Manning, Christopher D. and R{\'e}, Christopher and Acosta-Navas, Diana and Hudson, Drew A. and Zelikman, Eric and Durmus, Esin and Ladhak, Faisal and Rong, Frieda and Ren, Hongyu and Yao, Huaxiu and Wang, Jue and Santus, Enrico and Orr, Laurel and Zheng, John and Yuksekgonul, Mert and others},
  journal={arXiv preprint arXiv:2211.09110},
  year={2022}
}

@misc{mlcommons2024,
  title={MLCommons AI Safety Benchmark v0.5},
  author={{MLCommons AI Safety Working Group}},
  year={2024},
  howpublished={\url{https://mlcommons.org/benchmarks/ai-safety/}},
  note={Accessed: 2024-12-13}
}

@article{walters2025eliza,
  title={Eliza: A web3 friendly ai agent operating system},
  author={Walters, Shaw and Gao, Sam and Nerd, Shakker and Da, Feng and Williams, Warren and Meng, Ting-Chien and Chow, Amie and Han, Hunter and He, Frank and Zhang, Allen and others},
  journal={arXiv preprint arXiv:2501.06781},
  year={2025}
}

@article{daniel1997cadiag,
  title={CADIAG-2 and MYCIN-like systems},
  author={Daniel, Milan and H{\'a}jek, Petr and Nguyen, Phuong Hoang},
  journal={Artificial Intelligence in Medicine},
  volume={9},
  number={3},
  pages={241--259},
  year={1997},
  publisher={Elsevier}
}

@article{wang2025internet,
  title={Internet of agents: Fundamentals, applications, and challenges},
  author={Wang, Yuntao and Guo, Shaolong and Pan, Yanghe and Su, Zhou and Chen, Fahao and Luan, Tom H and Li, Peng and Kang, Jiawen and Niyato, Dusit},
  journal={arXiv preprint arXiv:2505.07176},
  year={2025}
}

@article{guo2024survey,
  title={A survey on semantic communication networks: Architecture, security, and privacy},
  author={Guo, Shaolong and Wang, Yuntao and Zhang, Ning and Su, Zhou and Luan, Tom H and Tian, Zhiyi and Shen, Xuemin},
  journal={IEEE Communications Surveys \& Tutorials},
  year={2024},
  publisher={IEEE}
}

@misc{McKinsey_StateOfAI2025,
  title        = {The State of AI: How organizations are rewiring to capture value},
  author       = {Alex Singla and Alexander Sukharevsky and Lareina Yee and Michael Chui and Bryce Hall},
  year         = {2025},
  howpublished = {\url{https://www.mckinsey.com/capabilities/quantumblack/our-insights/the-state-of-ai}},
  note         = {Accessed: 2025-09-10},
}

@article{langley2025spatial,
  title={Spatial representation and reasoning in an architecture for embodied agents},
  author={Langley, Pat and Katz, Edward P},
  journal={Spatial Cognition \& Computation},
  volume={25},
  number={1},
  pages={15--37},
  year={2025},
  publisher={Taylor \& Francis}
}

@article{acar2018survey,
  title={A survey on homomorphic encryption schemes: Theory and implementation},
  author={Acar, Abbas and Aksu, Hidayet and Uluagac, A Selcuk and Conti, Mauro},
  journal={ACM Computing Surveys (Csur)},
  volume={51},
  number={4},
  pages={1--35},
  year={2018},
  publisher={ACM New York, NY, USA}
}

@article{evans2018pragmatic,
  title={A pragmatic introduction to secure multi-party computation},
  author={Evans, David and Kolesnikov, Vladimir and Rosulek, Mike and others},
  journal={Foundations and Trends{\textregistered} in Privacy and Security},
  volume={2},
  number={2-3},
  pages={70--246},
  year={2018},
  publisher={Now Publishers, Inc.}
}

@article{thiers2024emerging,
  title={The emerging role of ISO 42001 certification in fostering the deployment of responsible generative AI healthcare solutions},
  author={Thiers, Fabio A and Harned, Zach},
  journal={Technology (NIST)},
  volume={55},
  number={58},
  pages={59},
  year={2024}
}

@misc{Stanford_HAI_AIIndex2025,
  title        = {2025 AI Index Report},
  author       = {Stanford Institute for Human-Centered Artificial Intelligence (HAI)},
  year         = {2025},
  howpublished = {\url{https://hai.stanford.edu/ai-index/2025-ai-index-report}},
  note         = {Accessed: 2025-09-10},
}

@misc{Gartner_AI_TRiSM2024,
  title        = {Tackling Trust, Risk and Security in AI Models (AI TRiSM)},
  author       = {Gartner},
  year         = {2024},
  howpublished = {\url{https://www.gartner.com/en/articles/ai-trust-and-ai-risk}},
  note         = {Accessed: 2025-09-10},
}

@article{ai2023artificial,
  title={Artificial intelligence risk management framework (AI RMF 1.0)},
  author={AI, NIST},
  journal={URL: https://nvlpubs. nist. gov/nistpubs/ai/nist. ai},
  pages={100--1},
  year={2023}
}

@misc{EU_AI_Act2025,
  title        = {AI Act},
  author       = {{European Commission}},
  howpublished = {\url{https://digital-strategy.ec.europa.eu/en/policies/regulatory-framework-ai}},
  note         = {Accessed: 2025-09-10},
  year         = {2025}
}

@article{schaar2010privacy,
  title={Privacy by design},
  author={Schaar, Peter},
  journal={Identity in the Information Society},
  volume={3},
  number={2},
  pages={267--274},
  year={2010},
  publisher={Springer}
}

@misc{oag_ccpa_2024,
  title        = {California Consumer Privacy Act (CCPA)},
  author       = {{California Department of Justice, Office of the Attorney General}},
  year         = {2024},
  howpublished = {Online},
  note         = {Updated March 13, 2024, accessed July 9, 2025},
  url          = {https://oag.ca.gov/privacy/ccpa}
}

@article{xi2025rise,
  title={The rise and potential of large language model based agents: A survey},
  author={Xi, Zhiheng and Chen, Wenxiang and Guo, Xin and He, Wei and Ding, Yiwen and Hong, Boyang and Zhang, Ming and Wang, Junzhe and Jin, Senjie and Zhou, Enyu and others},
  journal={Science China Information Sciences},
  volume={68},
  number={2},
  pages={121101},
  year={2025},
  publisher={Springer}
}

@article{raza2024exploring,
  title={Exploring bias and prediction metrics to characterise the fairness of machine learning for equity-centered public health decision-making: A narrative review},
  author={Raza, Shaina and Shaban-Nejad, Arash and Dolatabadi, Elham and Mamiya, Hiroshi},
  journal={IEEE Access},
  year={2024},
  publisher={IEEE}
}

@article{bilal2025llms,
  title={Llms for explainable ai: A comprehensive survey},
  author={Bilal, Ahsan and Ebert, David and Lin, Beiyu},
  journal={arXiv preprint arXiv:2504.00125},
  year={2025}
}

@article{gyevnar2023causal,
  title={Causal explanations for sequential decision-making in multi-agent systems},
  author={Gyevnar, Balint and Wang, Cheng and Lucas, Christopher G and Cohen, Shay B and Albrecht, Stefano V},
  journal={arXiv preprint arXiv:2302.10809},
  year={2023}
}

@inproceedings{yu2025survey,
  title={A survey on trustworthy llm agents: Threats and countermeasures},
  author={Yu, Miao and Meng, Fanci and Zhou, Xinyun and Wang, Shilong and Mao, Junyuan and Pan, Linsey and Chen, Tianlong and Wang, Kun and Li, Xinfeng and Zhang, Yongfeng and others},
  booktitle={Proceedings of the 31st ACM SIGKDD Conference on Knowledge Discovery and Data Mining V. 2},
  pages={6216--6226},
  year={2025}
}

@misc{wang2025surveyllmbasedagentsmedicine,
      title={A Survey of LLM-based Agents in Medicine: How far are we from Baymax?}, 
      author={Wenxuan Wang and Zizhan Ma and Zheng Wang and Chenghan Wu and Jiaming Ji and Wenting Chen and Xiang Li and Yixuan Yuan},
      year={2025},
      eprint={2502.11211},
      archivePrefix={arXiv},
      primaryClass={cs.CL},
      url={https://arxiv.org/abs/2502.11211}, 
}

@misc{zou2025llm,
      title={LLM-Based Human-Agent Collaboration and Interaction Systems: A Survey}, 
      author={Henry Peng Zou and Wei-Chieh Huang and Yaozu Wu and Yankai Chen and Chunyu Miao and Hoang Nguyen and Yue Zhou and Weizhi Zhang and Liancheng Fang and Langzhou He and Yangning Li and Dongyuan Li and Renhe Jiang and Xue Liu and Philip S. Yu},
      year={2025},
      eprint={2505.00753},
      archivePrefix={arXiv},
      primaryClass={cs.CL},
      url={https://arxiv.org/abs/2505.00753}, 
}

@article{luo2025large,
  title={Large language model agent: A survey on methodology, applications and challenges},
  author={Luo, Junyu and Zhang, Weizhi and Yuan, Ye and Zhao, Yusheng and Yang, Junwei and Gu, Yiyang and Wu, Bohan and Chen, Binqi and Qiao, Ziyue and Long, Qingqing and others},
  journal={arXiv preprint arXiv:2503.21460},
  year={2025}
}

@techreport{invicti_prompt_injection_ebook,
  title        = {Prompt Injection Attacks on Applications That Use LLMs},
  author       = {{Invicti Security}},
  institution  = {Invicti Security},
  year         = {2024},
  note         = {E‑book describing prompt injection attack types, real‑world examples, and mitigation strategies},
  url          = {https://www.invicti.com/white-papers/prompt-injection-attacks-on-llm-applications-ebook}
}

@article{langley2006cognitive,
  author = {Langley, Pat and Laird, John E. and Rogers, Seth},
  title = {Cognitive Architectures: Research Issues and Challenges},
  journal = {Cognitive Systems Research},
  volume = {7},
  number = {1},
  pages = {1--24},
  year = {2006},
  publisher = {Elsevier},
  doi = {10.1016/j.cogsys.2005.07.001}
}

@article{taatgen2006modeling,
  title={Modeling paradigms in ACT-R},
  author={Taatgen, Niels A and Lebiere, Christian and Anderson, John R},
  journal={Cognition and multi-agent interaction: From cognitive modeling to social simulation},
  pages={29--52},
  year={2006}
}

@book{anderson1993act,
  author = {Anderson, John R.},
  title = {Rules of the Mind},
  year = {1993},
  publisher = {Lawrence Erlbaum Associates},
  isbn = {9780805812237}
}

@article{laird1987soar,
  title={Soar: An architecture for general intelligence},
  author={Laird, John E and Newell, Allen and Rosenbloom, Paul S},
  journal={Artificial intelligence},
  volume={33},
  number={1},
  pages={1--64},
  year={1987},
  publisher={Elsevier}
}

@misc{guo2024largelanguagemodelbased,
      title={Large Language Model based Multi-Agents: A Survey of Progress and Challenges}, 
      author={Taicheng Guo and Xiuying Chen and Yaqi Wang and Ruidi Chang and Shichao Pei and Nitesh V. Chawla and Olaf Wiest and Xiangliang Zhang},
      year={2024},
      eprint={2402.01680},
      archivePrefix={arXiv},
      primaryClass={cs.CL},
      url={https://arxiv.org/abs/2402.01680}, 
}

@misc{chen2025surveyllmbasedmultiagentsystem,
      title={A Survey on LLM-based Multi-Agent System: Recent Advances and New Frontiers in Application}, 
      author={Shuaihang Chen and Yuanxing Liu and Wei Han and Weinan Zhang and Ting Liu},
      year={2025},
      eprint={2412.17481},
      archivePrefix={arXiv},
      primaryClass={cs.CL},
      url={https://arxiv.org/abs/2412.17481}, 
}

@misc{yan2025selftalkcommunicationcentricsurveyllmbased,
      title={Beyond Self-Talk: A Communication-Centric Survey of LLM-Based Multi-Agent Systems}, 
      author={Bingyu Yan and Xiaoming Zhang and Litian Zhang and Lian Zhang and Ziyi Zhou and Dezhuang Miao and Chaozhuo Li},
      year={2025},
      eprint={2502.14321},
      archivePrefix={arXiv},
      primaryClass={cs.MA},
      url={https://arxiv.org/abs/2502.14321}, 
}

@misc{tran2025multiagentcollaborationmechanismssurvey,
      title={Multi-Agent Collaboration Mechanisms: A Survey of LLMs}, 
      author={Khanh-Tung Tran and Dung Dao and Minh-Duong Nguyen and Quoc-Viet Pham and Barry O'Sullivan and Hoang D. Nguyen},
      year={2025},
      eprint={2501.06322},
      archivePrefix={arXiv},
      primaryClass={cs.AI},
      url={https://arxiv.org/abs/2501.06322}, 
}

@misc{lin2025creativityllmbasedmultiagentsystems,
      title={Creativity in LLM-based Multi-Agent Systems: A Survey}, 
      author={Yi-Cheng Lin and Kang-Chieh Chen and Zhe-Yan Li and Tzu-Heng Wu and Tzu-Hsuan Wu and Kuan-Yu Chen and Hung-yi Lee and Yun-Nung Chen},
      year={2025},
      eprint={2505.21116},
      archivePrefix={arXiv},
      primaryClass={cs.HC},
      url={https://arxiv.org/abs/2505.21116}, 
}

@misc{fang2025trustworthyaisafetybias,
      title={Trustworthy AI on Safety, Bias, and Privacy: A Survey}, 
      author={Xingli Fang and Jianwei Li and Varun Mulchandani and Jung-Eun Kim},
      year={2025},
      eprint={2502.10450},
      archivePrefix={arXiv},
      primaryClass={cs.CR},
      url={https://arxiv.org/abs/2502.10450}, 
}

@misc{langgraph2024,
  author       = {Microsoft Corporation},
  title        = {LangGraph: Agent Orchestration Framework for LLMs},
  year         = {2024},
  url          = {https://www.langchain.com/langgraph},
  note         = {Accessed: 2025-06-03},
  howpublished = {\url{https://www.langchain.com/langgraph}}
}

@article{weyns2009agent,
  title={The agent environment in multi-agent systems: A middleware perspective},
  author={Weyns, Danny and Helleboogh, Alexander and Holvoet, Tom and Schumacher, Michael},
  journal={Multiagent and Grid Systems},
  volume={5},
  number={1},
  pages={93--108},
  year={2009},
  publisher={SAGE Publications Sage UK: London, England}
}

@article{yang2023auto,
  title={Auto-gpt for online decision making: Benchmarks and additional opinions},
  author={Yang, Hui and Yue, Sifu and He, Yunzhong},
  journal={arXiv preprint arXiv:2306.02224},
  year={2023}
}

@misc{gpt-engineer,
  author       = {Anton Osika},
  title        = {gpt-engineer: CLI platform to experiment with codegen},
  year         = {2023},
  publisher    = {GitHub},
  journal      = {GitHub repository},
  howpublished = {\url{https://github.com/AntonOsika/gpt-engineer}},
  note         = {MIT License}
}

@article{acharya2025agentic,
  title={Agentic AI: Autonomous Intelligence for Complex Goals--A Comprehensive Survey},
  author={Acharya, Deepak Bhaskar and Kuppan, Karthigeyan and Divya, B},
  journal={IEEE Access},
  year={2025},
  publisher={IEEE}
}

@article{liu2025agent,
  title={Agent-Environment Alignment via Automated Interface Generation},
  author={Liu, Kaiming and Lei, Xuanyu and Wang, Ziyue and Li, Peng and Liu, Yang},
  journal={arXiv preprint arXiv:2505.21055},
  year={2025}
}

@misc{schick2023toolformerlanguagemodelsteach,
      title={Toolformer: Language Models Can Teach Themselves to Use Tools}, 
      author={Timo Schick and Jane Dwivedi-Yu and Roberto Dessì and Roberta Raileanu and Maria Lomeli and Luke Zettlemoyer and Nicola Cancedda and Thomas Scialom},
      year={2023},
      eprint={2302.04761},
      archivePrefix={arXiv},
      primaryClass={cs.CL},
      url={https://arxiv.org/abs/2302.04761}, 
}

@misc{karpas2022mrklsystemsmodularneurosymbolic,
      title={MRKL Systems: A modular, neuro-symbolic architecture that combines large language models, external knowledge sources and discrete reasoning}, 
      author={Ehud Karpas and Omri Abend and Yonatan Belinkov and Barak Lenz and Opher Lieber and Nir Ratner and Yoav Shoham and Hofit Bata and Yoav Levine and Kevin Leyton-Brown and Dor Muhlgay and Noam Rozen and Erez Schwartz and Gal Shachaf and Shai Shalev-Shwartz and Amnon Shashua and Moshe Tenenholtz},
      year={2022},
      eprint={2205.00445},
      archivePrefix={arXiv},
      primaryClass={cs.CL},
      url={https://arxiv.org/abs/2205.00445}, 
}

@article{langchain2024function,
  title = {LangChain and LangGraph: Comparing Function and Tool Calling Capabilities},
  author = {{LangChain Team}},
  journal = {LangChain Blog},
  year = {2024},
  month = {April},
  url = {https://www.langchain.com/},
  note = {Accessed: 2025-06-03}
}

@inproceedings{xia2018gibson,
  title={Gibson env: Real-world perception for embodied agents},
  author={Xia, Fei and Zamir, Amir R and He, Zhiyang and Sax, Alexander and Malik, Jitendra and Savarese, Silvio},
  booktitle={Proceedings of the IEEE conference on computer vision and pattern recognition},
  pages={9068--9079},
  year={2018}
}

@misc{securiti2025aitrism,
  author       = {{Securiti}},
  title        = {What is {AI TRiSM} and Why It's Essential in the Era of GenAI},
  year         = {2025},
  month        = {May},
  url          = {https://securiti.ai/what-is-ai-trism/},
  note         = {Accessed: 2025-06-03}
}

@misc{eu_ai_act,
  author       = {{European Commission}},
  title        = {{AI Act | Shaping Europe's digital future}},
  year         = {2024},
  url          = {https://digital-strategy.ec.europa.eu/en/policies/regulatory-framework-ai},
  note         = {Accessed: 2025-06-03}
}

@inproceedings{rodden2010measuring,
  title={Measuring the user experience on a large scale: user-centered metrics for web applications},
  author={Rodden, Kerry and Hutchinson, Hilary and Fu, Xin},
  booktitle={Proceedings of the SIGCHI conference on human factors in computing systems},
  pages={2395--2398},
  year={2010}
}

@misc{microsoft_semantickernel,
  author       = {Microsoft},
  title        = {Semantic Kernel Agent Framework},
  year         = {2025},
  howpublished = {\url{https://learn.microsoft.com/en-us/semantic-kernel/frameworks/agent/}},
  note         = {Accessed: 2025-09-10}
}

@standard{ISOIEC23894_2023,
  title        = {Information technology — Artificial intelligence — Guidance on risk management},
  number       = {ISO/IEC 23894:2023},
  series       = {ISO/IEC},
  year         = {2023},
  month        = feb,
  publisher    = {International Organization for Standardization and International Electrotechnical Commission},
  location     = {Geneva},
  edition      = {1},
  pages        = {26},
}

@article{ai2024artificial,
  title={Artificial intelligence risk management framework: Generative artificial intelligence profile},
  author={AI, NIST},
  journal={NIST Trustworthy and Responsible AI Gaithersburg, MD, USA},
  year={2024}
}

@article{barbera2025ai,
  title={AI Privacy Risks \& Mitigations—Large Language Models (LLMs)},
  author={Barber{\'a}, I},
  journal={European Data Protection Board. Available online: https://www. edpb. europa. eu/system/files/2025-04/ai-privacy-risks-and-mitigations-in-llms. pdf (accessed on 12 June 2025)},
  year={2025}
}

@article{beurer2025design,
  title={Design patterns for securing llm agents against prompt injections},
  author={Beurer-Kellner, Luca and Buesser, Beat and Cre{\c{t}}u, Ana-Maria and Debenedetti, Edoardo and Dobos, Daniel and Fabian, Daniel and Fischer, Marc and Froelicher, David and Grosse, Kathrin and Naeff, Daniel and others},
  journal={arXiv preprint arXiv:2506.08837},
  year={2025}
}

@inproceedings{wang2025megaagent,
  title={MegaAgent: A large-scale autonomous LLM-based multi-agent system without predefined SOPs},
  author={Wang, Qian and Wang, Tianyu and Tang, Zhenheng and Li, Qinbin and Chen, Nuo and Liang, Jingsheng and He, Bingsheng},
  booktitle={Findings of the Association for Computational Linguistics: ACL 2025},
  pages={4998--5036},
  year={2025}
}

@article{Sinha2025ModelOps,
  author       = {Santosh Sinha},
  title        = {The Rise of ModelOps: What Comes After MLOps?},
  journal      = {Brim Labs Blog},
  year         = {2025},
  month        = jun,
  day          = {10},
  url          = {https://brimlabs.ai/blog/the-rise-of-modelops-what-comes-after-mlops/},
}

@article{ozgun2025trustworthy,
  title={Trustworthy AI Psychotherapy: Multi-Agent LLM Workflow for Counseling and Explainable Mental Disorder Diagnosis},
  author={Ozgun, Mithat Can and Pei, Jiahuan and Hindriks, Koen and Donatelli, Lucia and Liu, Qingzhi and Sun, Xin and Wang, Junxiao},
  journal={arXiv preprint arXiv:2508.11398},
  year={2025}
}

@article{nagpal2025leveraging,
  title={Leveraging Large Language Models for Effective and Explainable Multi-Agent Credit Assignment},
  author={Nagpal, Kartik and Dong, Dayi and Bouvier, Jean-Baptiste and Mehr, Negar},
  journal={arXiv preprint arXiv:2502.16863},
  year={2025}
}

@inproceedings{lo2025ai,
  title={AI hiring with llms: A context-aware and explainable multi-agent framework for resume screening},
  author={Lo, Frank P-W and Qiu, Jianing and Wang, Zeyu and Yu, Haibao and Chen, Yeming and Zhang, Gao and Lo, Benny},
  booktitle={Proceedings of the Computer Vision and Pattern Recognition Conference},
  pages={4184--4193},
  year={2025}
}

@misc{openai_swarm,
  author       = {OpenAI},
  title        = {Swarm: Lightweight Multi-Agent Orchestration Framework},
  year         = {2024},
  howpublished = {\url{https://github.com/openai/swarm}},
  note         = {Accessed: 2025-09-10}
}

@article{sanwal2025layered,
  title={Layered chain-of-thought prompting for multi-agent llm systems: A comprehensive approach to explainable large language models},
  author={Sanwal, Manish},
  journal={arXiv preprint arXiv:2501.18645},
  year={2025}
}

@misc{openai_agents_sdk,
  author       = {OpenAI},
  title        = {OpenAI Agents SDK (Python)},
  year         = {2025},
  howpublished = {\url{https://github.com/openai/openai-agents-python}},
  note         = {Accessed: 2025-09-10}
}

@misc{strands_agents,
  author       = {AWS Strands Project},
  title        = {Strands Agents SDK (Python)},
  year         = {2025},
  howpublished = {\url{https://github.com/strands-agents/sdk-python}},
  note         = {Accessed: 2025-09-10}
}

@misc{llamaindex_agents,
  author       = {LlamaIndex},
  title        = {LlamaIndex Agents Documentation},
  year         = {2024},
  howpublished = {\url{https://docs.llamaindex.ai/en/stable/use_cases/agents/}},
  note         = {Accessed: 2025-09-10}
}

@standard{ISO_IEC_42005_2025,
  title        = {Information technology — Artificial intelligence (AI) — AI system impact assessment},
  number       = {ISO/IEC 42005:2025},
  year         = {2025},
  month        = may,
  edition      = {1},
  organization = {International Organization for Standardization {\&} International Electrotechnical Commission},
  address      = {Geneva, Switzerland},
  pages        = {39},
  note         = {Published May 2025; reference number 42005; International standard},
}

@article{tomsett2018interpretable,
  title={Interpretable to whom? A role-based model for analyzing interpretable machine learning systems},
  author={Tomsett, Richard and Braines, Dave and Harborne, Dan and Preece, Alun and Chakraborty, Supriyo},
  journal={arXiv preprint arXiv:1806.07552},
  year={2018}
}

@misc{oecd_ai_catalogue,
  title  = {OECD.AI Catalogue of Tools \& Metrics for Trustworthy AI},
  howpublished = {\url{https://oecd.ai/en/}},
  note   = {OECD .AI is a live platform promoting trustworthy, human‑centric AI; accessed 24 Jun 2025},
}

@misc{agarwal2024openxaitransparentevaluationmodel,
      title={OpenXAI: Towards a Transparent Evaluation of Model Explanations}, 
      author={Chirag Agarwal and Dan Ley and Satyapriya Krishna and Eshika Saxena and Martin Pawelczyk and Nari Johnson and Isha Puri and Marinka Zitnik and Himabindu Lakkaraju},
      year={2024},
      eprint={2206.11104},
      archivePrefix={arXiv},
      primaryClass={cs.LG},
      url={https://arxiv.org/abs/2206.11104}, 
}

@article{raza2025humanibench,
  title={Humanibench: A human-centric framework for large multimodal models evaluation},
  author={Raza, Shaina and Narayanan, Aravind and Khazaie, Vahid Reza and Vayani, Ashmal and Chettiar, Mukund S and Singh, Amandeep and Shah, Mubarak and Pandya, Deval},
  journal={arXiv preprint arXiv:2505.11454},
  year={2025}
}

@inproceedings{vayani2025all,
  title={All languages matter: Evaluating lmms on culturally diverse 100 languages},
  author={Vayani, Ashmal and Dissanayake, Dinura and Watawana, Hasindri and Ahsan, Noor and Sasikumar, Nevasini and Thawakar, Omkar and Ademtew, Henok Biadglign and Hmaiti, Yahya and Kumar, Amandeep and Kukreja, Kartik and others},
  booktitle={Proceedings of the Computer Vision and Pattern Recognition Conference},
  pages={19565--19575},
  year={2025}
}

@misc{liu2023gevalnlgevaluationusing,
      title={G-Eval: NLG Evaluation using GPT-4 with Better Human Alignment}, 
      author={Yang Liu and Dan Iter and Yichong Xu and Shuohang Wang and Ruochen Xu and Chenguang Zhu},
      year={2023},
      eprint={2303.16634},
      archivePrefix={arXiv},
      primaryClass={cs.CL},
      url={https://arxiv.org/abs/2303.16634}, 
}

@inproceedings{StrDorFri2024,
title={LLM2Swarm: {R}obot Swarms that Responsively Reason, Plan, and Collaborate through LLMs},
author={Volker Strobel and Marco Dorigo and Mario Fritz},
year={2024},
booktitle={NeurIPS 2024 Workshop on Open-World Agents (OWA-2024)}
}

@misc{liu2023lostmiddlelanguagemodels,
      title={Lost in the Middle: How Language Models Use Long Contexts}, 
      author={Nelson F. Liu and Kevin Lin and John Hewitt and Ashwin Paranjape and Michele Bevilacqua and Fabio Petroni and Percy Liang},
      year={2023},
      eprint={2307.03172},
      archivePrefix={arXiv},
      primaryClass={cs.CL},
      url={https://arxiv.org/abs/2307.03172}, 
}

@article{atkins2023memorypoisoning,
  title={Those Aren't Your Memories, They're Somebody Else's: Seeding Misinformation in Chat Bot Memories},
  author={Atkins, Conor and Zhao, Benjamin Zi Hao and Asghar, Hassan Jameel and Wood, Ian and Kaafar, Mohamed Ali},
  journal={arXiv preprint arXiv:2304.05371},
  year={2023},
  url={https://arxiv.org/abs/2304.05371}
}

@article{ligot2022using,
  title={On using simulation to predict the performance of robot swarms},
  author={Ligot, Antoine and Birattari, Mauro},
  journal={Scientific Data},
  volume={9},
  number={1},
  pages={788},
  year={2022},
  publisher={Nature Publishing Group UK London}
}

@misc{qian2024chatdevcommunicativeagentssoftware,
  publtype={informal},
  author={Chen Qian and Xin Cong and Cheng Yang and Weize Chen and Yusheng Su and Juyuan Xu and Zhiyuan Liu and Maosong Sun},
  title={Communicative Agents for Software Development},
  year={2023},
  cdate={1672531200000},
  journal={CoRR},
  volume={abs/2307.07924},
  url={https://doi.org/10.48550/arXiv.2307.07924}
}

@misc{hipaa164,
  title={{HIPAA Privacy Rule -- 45 CFR Part 164: Security and Privacy Protections for Health Information}},
  author={{U.S. Department of Health and Human Services}},
  year={2003},
  howpublished={\url{https://www.ecfr.gov/current/title-45/subtitle-A/subchapter-C/part-164}},
  note={Accessed: 2025-06-03}
}

@misc{gdpr25,
  title={{General Data Protection Regulation (GDPR) -- Article 25: Data protection by design and by default}},
  author={{European Union}},
  year={2016},
  howpublished={\url{https://gdpr-info.eu/art-25-gdpr/}},
  note={Accessed: 2025-06-03}
}

@misc{owasp2024llmtop10,
  title={{OWASP Top 10 for Large Language Model Applications}},
  author={{Open Worldwide Application Security Project (OWASP)}},
  year={2024},
  howpublished={\url{https://owasp.org/www-project-top-10-for-large-language-model-applications/}},
  note={Accessed: 2025-06-03}
}

@techreport{iso42001,
  title={ISO/IEC 42001:2023 -- Artificial Intelligence Management System (AI MS) -- Requirements},
  author={{International Organization for Standardization}},
  year={2023},
  institution={ISO/IEC},
  note={Available at \url{https://www.iso.org/standard/81230.html}}
}

@misc{tabassi2023ai,
  author       = {Elham Tabassi},
  title        = {Artificial Intelligence Risk Management Framework (AI RMF 1.0)},
  year         = {2023},
  month        = {January 26},
  publisher    = {NIST Trustworthy and Responsible AI, National Institute of Standards and Technology, Gaithersburg, MD},
  url          = {https://tsapps.nist.gov/publication/get_pdf.cfm?pub_id=936225},
  doi          = {10.6028/NIST.AI.100-1},
  language     = {en}
}

@inproceedings{gentry2009fully,
  title={Fully homomorphic encryption using ideal lattices},
  author={Gentry, Craig},
  booktitle={Proceedings of the 41st annual ACM symposium on Theory of computing},
  pages={169--178},
  year={2009},
  organization={ACM}
}

@techreport{iso24029-1,
  title={ISO/IEC TR 24029-1:2021 -- Artificial intelligence (AI) -- Assessment of the robustness of neural networks -- Part 1: Overview},
  author={{International Organization for Standardization}},
  year={2021},
  institution={ISO/IEC},
  note={Available at \url{https://www.iso.org/standard/77608.html}}
}

@inproceedings{costan2016intel,
  title={Intel SGX explained},
  author={Costan, Victor and Devadas, Srinivas},
  booktitle={IACR Cryptology ePrint Archive},
  volume={2016},
  pages={86},
  year={2016}
}

@inproceedings{lindell2009secure,
  title={Secure two-party computation via cut-and-choose oblivious transfer},
  author={Lindell, Yehuda and Pinkas, Benny},
  booktitle={Advances in Cryptology—EUROCRYPT 2007},
  pages={329--346},
  year={2007},
  organization={Springer}
}

@article{sweeney2002k,
  title={k-anonymity: A model for protecting privacy},
  author={Sweeney, Latanya},
  journal={International Journal of Uncertainty, Fuzziness and Knowledge-Based Systems},
  volume={10},
  number={05},
  pages={557--570},
  year={2002},
  publisher={World Scientific}
}

@inproceedings{dwork2006calibrating,
  title={Calibrating noise to sensitivity in private data analysis},
  author={Dwork, Cynthia and McSherry, Frank and Nissim, Kobbi and Smith, Adam},
  booktitle={Theory of Cryptography Conference},
  pages={265--284},
  year={2006},
  organization={Springer}
}

@techreport{owasp2025agentic,
  title        = {Agentic AI – Threats and Mitigations},
  author       = {{OWASP Agentic Security Initiative}},
  institution  = {Open Worldwide Application Security Project (OWASP)},
  year         = {2025},
  month        = {February},
  url          = {https://hal.science/hal-04985337v1/file/Agentic-AI-Threats-and-Mitigations_v1.0.1.pdf},
  note         = {Accessed: 2025-06-03}
}

@misc{raza2025justhumansneedvaccines,
      title={Just as Humans Need Vaccines, So Do Models: Model Immunization to Combat Falsehoods}, 
      author={Shaina Raza and Rizwan Qureshi and Marcelo Lotif and Aman Chadha and Deval Pandya and Christos Emmanouilidis},
      year={2025},
      eprint={2505.17870},
      archivePrefix={arXiv},
      primaryClass={cs.CL},
      url={https://arxiv.org/abs/2505.17870}, 
}

@article{lefevre2022modelops,
  title     = {ModelOps for Enhanced Decision-Making and Governance in Emergency Control Rooms},
  author    = {Lefevre, Kay and Arora, Chetan and Lee, Kevin and Zaslavsky, Arkady and Bouadjenek, Mohamed Reda and Hassani, Ali and Razzak, Imran},
  journal   = {Environment Systems and Decisions},
  volume    = {42},
  number    = {3},
  pages     = {402--416},
  year      = {2022},
  month     = {September},
  publisher = {Springer},
  doi       = {10.1007/s10669-022-09855-1},
  url       = {https://doi.org/10.1007/s10669-022-09855-1}
}

@inproceedings{lundberg2017unified,
  title     = {A Unified Approach to Interpreting Model Predictions},
  author    = {Lundberg, Scott M. and Lee, Su-In},
  booktitle = {Proceedings of the 31st International Conference on Neural Information Processing Systems},
  pages     = {4765--4774},
  year      = {2017},
  publisher = {Curran Associates Inc.},
  url       = {https://proceedings.neurips.cc/paper_files/paper/2017/file/8a20a8621978632d76c43dfd28b67767-Paper.pdf}
}

@article{kaur2024building,
  title = {Building Trust with AI TRiSM: Managing Risks in the Era of Agentic AI},
  author = {Kaur, Jagreet},
  journal = {Akira AI Blog},
  year = {2024},
  month = {December},
  url = {https://www.akira.ai/blog/ai-trism-with-ai-agents},
  note = {Accessed: 2025-06-03}
}

@misc{ko2025sevensecuritychallengessolved,
      title={Seven Security Challenges That Must be Solved in Cross-domain Multi-agent LLM Systems}, 
      author={Ronny Ko and Jiseong Jeong and Shuyuan Zheng and Chuan Xiao and Taewan Kim and Makoto Onizuka and Wonyong Shin},
      year={2025},
      eprint={2505.23847},
      archivePrefix={arXiv},
      primaryClass={cs.CR},
      url={https://arxiv.org/abs/2505.23847}, 
}

@article{litan2024ai,
  title = {AI Trust and AI Risk: Tackling Trust, Risk and Security in AI Models},
  author = {Litan, Avivah},
  journal = {Gartner},
  year = {2024},
  month = {December},
  url = {https://www.gartner.com/en/articles/ai-trust-and-ai-risk},
  note = {Accessed: 2025-06-03}
}

@misc{superagi,
  author       = {TransformerOptimus and contributors},
  title        = {SuperAGI: A Dev-First Open Source Autonomous AI Agent Framework},
  year         = {2023},
  publisher    = {GitHub},
  howpublished = {\url{https://github.com/TransformerOptimus/SuperAGI}},
  note         = {MIT License}
}

@misc{xie2023openagentsopenplatformlanguage,
      title={OpenAgents: An Open Platform for Language Agents in the Wild}, 
      author={Tianbao Xie and Fan Zhou and Zhoujun Cheng and Peng Shi and Luoxuan Weng and Yitao Liu and Toh Jing Hua and Junning Zhao and Qian Liu and Che Liu and Leo Z. Liu and Yiheng Xu and Hongjin Su and Dongchan Shin and Caiming Xiong and Tao Yu},
      year={2023},
      eprint={2310.10634},
      archivePrefix={arXiv},
      primaryClass={cs.CL},
      url={https://arxiv.org/abs/2310.10634}, 
}

@article{chen2023agentverse,
  title={Agentverse: Facilitating multi-agent collaboration and exploring emergent behaviors in agents},
  author={Chen, Weize and Su, Yusheng and Zuo, Jingwei and Yang, Cheng and Yuan, Chenfei and Qian, Chen and Chan, Chi-Min and Qin, Yujia and Lu, Yaxi and Xie, Ruobing and others},
  journal={arXiv preprint arXiv:2308.10848},
  year={2023}
}

@misc{crewAI,
  author       = {João Moura and contributors},
  title        = {CrewAI: Framework for Orchestrating Role-Playing, Autonomous AI Agents},
  year         = {2023},
  publisher    = {GitHub},
  howpublished = {\url{https://github.com/crewAIInc/crewAI}},
  note         = {MIT License}
}

@inproceedings{li2023camel,
  title={CAMEL: Communicative Agents for "Mind" Exploration of Large Language Model Society},
  author={Li, Guohao and Hammoud, Hasan Abed Al Kader and Itani, Hani and Khizbullin, Dmitrii and Ghanem, Bernard},
  booktitle={Thirty-seventh Conference on Neural Information Processing Systems},
  year={2023}
}

@article{shen2023hugginggpt,
  title={Hugginggpt: Solving ai tasks with chatgpt and its friends in hugging face},
  author={Shen, Yongliang and Song, Kaitao and Tan, Xu and Li, Dongsheng and Lu, Weiming and Zhuang, Yueting},
  journal={Advances in Neural Information Processing Systems},
  volume={36},
  pages={38154--38180},
  year={2023}
}

@article{he2024webvoyager,
  title={WebVoyager: Building an end-to-end web agent with large multimodal models},
  author={He, Hongliang and Yao, Wenlin and Ma, Kaixin and Yu, Wenhao and Dai, Yong and Zhang, Hongming and Lan, Zhenzhong and Yu, Dong},
  journal={arXiv preprint arXiv:2401.13919},
  year={2024}
}

@misc{shinn2023reflexionlanguageagentsverbal,
      title={Reflexion: Language Agents with Verbal Reinforcement Learning}, 
      author={Noah Shinn and Federico Cassano and Edward Berman and Ashwin Gopinath and Karthik Narasimhan and Shunyu Yao},
      year={2023},
      eprint={2303.11366},
      archivePrefix={arXiv},
      primaryClass={cs.AI},
      url={https://arxiv.org/abs/2303.11366}, 
}

@misc{wang2023voyageropenendedembodiedagent,
      title={Voyager: An Open-Ended Embodied Agent with Large Language Models}, 
      author={Guanzhi Wang and Yuqi Xie and Yunfan Jiang and Ajay Mandlekar and Chaowei Xiao and Yuke Zhu and Linxi Fan and Anima Anandkumar},
      year={2023},
      eprint={2305.16291},
      archivePrefix={arXiv},
      primaryClass={cs.AI},
      url={https://arxiv.org/abs/2305.16291}, 
}

@inproceedings{hong2024metagpt,
      title={Meta{GPT}: Meta Programming for A Multi-Agent Collaborative Framework},
      author={Sirui Hong and Mingchen Zhuge and Jonathan Chen and Xiawu Zheng and Yuheng Cheng and Jinlin Wang and Ceyao Zhang and Zili Wang and Steven Ka Shing Yau and Zijuan Lin and Liyang Zhou and Chenyu Ran and Lingfeng Xiao and Chenglin Wu and J{\"u}rgen Schmidhuber},
      booktitle={The Twelfth International Conference on Learning Representations},
      year={2024},
      url={https://openreview.net/forum?id=VtmBAGCN7o}
}

@misc{yao2023reactsynergizingreasoningacting,
      title={ReAct: Synergizing Reasoning and Acting in Language Models}, 
      author={Shunyu Yao and Jeffrey Zhao and Dian Yu and Nan Du and Izhak Shafran and Karthik Narasimhan and Yuan Cao},
      year={2023},
      eprint={2210.03629},
      archivePrefix={arXiv},
      primaryClass={cs.CL},
      url={https://arxiv.org/abs/2210.03629}, 
}

@misc{nakajima_babyagi_archive_2024,
  author       = {Nakajima, Yohei},
  title        = {BabyAGI (Archived Version)},
  year         = {2024},
  publisher    = {GitHub},
  howpublished = {\url{https://github.com/yoheinakajima/babyagi_archive}},
  note         = {Accessed: 2025-06-03}
}

@misc{Saravia_Prompt_Engineering_Guide_Web_2024,
  author       = {Saravia, Elvis},
  title        = {Prompt Engineering Guide},
  year         = {2024},
  howpublished = {\url{https://www.promptingguide.ai/}},
  note         = {Accessed: 2025-06-03}
}

@misc{Saravia_LLM_Agents_2024,
  author       = {Saravia, Elvis},
  title        = {LLM Agents},
  year         = {2024},
  howpublished = {\url{https://www.promptingguide.ai/research/llm-agents}},
  note         = {Accessed: 2025-06-03}
}

@article{kitchenham2004procedures,
  title={Procedures for performing systematic reviews},
  author={Kitchenham, Barbara},
  journal={Keele, UK, Keele University},
  volume={33},
  number={2004},
  pages={1--26},
  year={2004},
  publisher={Citeseer}
}

@article{yang2024multi,
  title={Multi-llm-agent systems: Techniques and business perspectives},
  author={Yang, Yingxuan and Peng, Qiuying and Wang, Jun and Zhang, Weinan},
  journal={arXiv preprint arXiv:2411.14033},
  year={2024}
}

@article{russell1997rationality,
  title={Rationality and intelligence},
  author={Russell, Stuart J},
  journal={Artificial intelligence},
  volume={94},
  number={1-2},
  pages={57--77},
  year={1997},
  publisher={Elsevier}
}

@article{parasuraman2010complacency,
  title={Complacency and Bias in Human Use of Automation: An Attentional Integration},
  author={Parasuraman, Raja and Manzey, Dietrich H.},
  journal={Human Factors},
  volume={52},
  number={3},
  pages={381--410},
  year={2010},
  publisher={SAGE Publications},
  doi={10.1177/0018720810376055},
  url={https://journals.sagepub.com/doi/10.1177/0018720810376055}
}

@article{qian2023chatdev,
  title={Chatdev: Communicative agents for software development},
  author={Qian, Chen and Liu, Wei and Liu, Hongzhang and Chen, Nuo and Dang, Yufan and Li, Jiahao and Yang, Cheng and Chen, Weize and Su, Yusheng and Cong, Xin and others},
  journal={arXiv preprint arXiv:2307.07924},
  year={2023}
}

@article{wang2024agent,
  title   = {Agent AI with LangGraph: A Modular Framework for Enhancing Machine Translation Using Large Language Models},
  author  = {Wang, Jialin and Duan, Zhihua},
  journal = {arXiv preprint arXiv:2412.03801},
  year    = {2024},
  url     = {https://arxiv.org/abs/2412.03801}
}

@article{wu2023autogen,
  title={Autogen: Enabling next-gen llm applications via multi-agent conversation},
  author={Wu, Qingyun and Bansal, Gagan and Zhang, Jieyu and Wu, Yiran and Li, Beibin and Zhu, Erkang and Jiang, Li and Zhang, Xiaoyun and Zhang, Shaokun and Liu, Jiale and others},
  journal={arXiv preprint arXiv:2308.08155},
  year={2023}
}

@article{hong2023metagpt,
  title={Metagpt: Meta programming for multi-agent collaborative framework},
  author={Hong, Sirui and Zheng, Xiawu and Chen, Jonathan and Cheng, Yuheng and Wang, Jinlin and Zhang, Ceyao and Wang, Zili and Yau, Steven Ka Shing and Lin, Zijuan and Zhou, Liyang and others},
  journal={arXiv preprint arXiv:2308.00352},
  volume={3},
  number={4},
  pages={6},
  year={2023}
}

@article{muscettola1998remote,
  title={Remote agent: To boldly go where no AI system has gone before},
  author={Muscettola, Nicola and Nayak, P Pandurang and Pell, Barney and Williams, Brian C},
  journal={Artificial intelligence},
  volume={103},
  number={1-2},
  pages={5--47},
  year={1998},
  publisher={Elsevier}
}

@incollection{lim2012memory,
  title={Memory models for intelligent social companions},
  author={Lim, Mei Yii},
  booktitle={Human-computer interaction: The agency perspective},
  pages={241--262},
  year={2012},
  publisher={Springer}
}

@article{habbal2024artificial,
  title={Artificial Intelligence Trust, risk and security management (AI trism): Frameworks, applications, challenges and future research directions},
  author={Habbal, Adib and Ali, Mohamed Khalif and Abuzaraida, Mustafa Ali},
  journal={Expert Systems with Applications},
  volume={240},
  pages={122442},
  year={2024},
  publisher={Elsevier}
}

@inproceedings{hannebauer1999formal,
  title={From formal workflow models to intelligent agents},
  author={Hannebauer, Markus},
  booktitle={Proceedings of the AAAI-99 Workshop on Agent Based Systems in the Business Context},
  pages={19--24},
  year={1999}
}

@inproceedings{cranshaw2017calendar,
  title={Calendar. help: Designing a workflow-based scheduling agent with humans in the loop},
  author={Cranshaw, Justin and Elwany, Emad and Newman, Todd and Kocielnik, Rafal and Yu, Bowen and Soni, Sandeep and Teevan, Jaime and Monroy-Hern{\'a}ndez, Andr{\'e}s},
  booktitle={Proceedings of the 2017 CHI Conference on Human Factors in Computing Systems},
  pages={2382--2393},
  year={2017}
}

@article{boland2017doing,
  title={Doing a systematic review: a student s guide},
  author={Boland, Angela and Cherry, Gemma and Dickson, Rumona},
  year={2017},
  publisher={Sage}
}

@inproceedings{AmershiWVFNCSIB19,
  author    = {Saleema Amershi and Daniel S. Weld and Mihaela Vorvoreanu and Adam Fourney and Besmira Nushi and Penny Collisson and Jina Suh and Shamsi T. Iqbal and Paul N. Bennett and Kori Inkpen and Jaime Teevan and Ruth Kikin-Gil and Eric Horvitz},
  title     = {Guidelines for Human-AI Interaction},
  booktitle = {Proceedings of the 2019 CHI Conference on Human Factors in Computing Systems},
  year      = {2019},
  publisher = {ACM},
  isbn      = {978-1-4503-5970-2},
  url       = {https://dl.acm.org/doi/10.1145/3290605.3300233}
}

@article{schneider2020wargaming,
  title={Wargaming Cyber Security},
  author={Schneider, Jacquelyn},
  journal={War on the Rocks},
  year={2020},
  month={September},
  url={https://warontherocks.com/2020/09/wargaming-cyber-security/}
}

@article{bacudio2011overview,
  title={An Overview of Penetration Testing},
  author={Bacudio, Aileen G. and Yuan, Xiaohong and Chu, Bei-Tseng Bill and Jones, Monique},
  journal={International Journal of Network Security \& Its Applications (IJNSA)},
  volume={3},
  number={6},
  pages={19--38},
  year={2011},
  month={November},
  doi={10.5121/ijnsa.2011.3602},
  url={https://airccse.org/journal/nsa/1111nsa02.pdf}
}

@misc{autogen_agents_tutorial,
  title        = {Agents — AutoGen AgentChat User Guide},
  author       = {{Microsoft Open Source}},
  year         = {2024},
  howpublished = {\url{https://microsoft.github.io/autogen/stable//user-guide/agentchat-user-guide/tutorial/agents.html}},
  note         = {Accessed: 2025-06-02}
}

@techreport{keele2007guidelines,
  title={Guidelines for performing systematic literature reviews in software engineering},
  author={Keele, Staffs and others},
  year={2007},
  institution={Technical report, ver. 2.3 ebse technical report. ebse}
}

@online{IBM2025AITRiSM,
  author       = {Alice Gomstyn and Alexandra Jonker},
  title        = {What Is AI TRiSM?},
  year         = {2025},
  month        = mar,
  url          = {https://www.ibm.com/think/topics/ai-trism},
  note         = {Accessed: 2025-06-02}
}

@article{govaerts2010development,
  title={Development of a software tool using deterministic logic for the optimization of cochlear implant processor programming},
  author={Govaerts, Paul J and Vaerenberg, Bart and De Ceulaer, Geert and Daemers, Kristin and De Beukelaer, Carina and Schauwers, Karen},
  journal={Otology \& Neurotology},
  volume={31},
  number={6},
  pages={908--918},
  year={2010},
  publisher={LWW}
}

@article{qiao2024agent,
  title={Agent planning with world knowledge model},
  author={Qiao, Shuofei and Fang, Runnan and Zhang, Ningyu and Zhu, Yuqi and Chen, Xiang and Deng, Shumin and Jiang, Yong and Xie, Pengjun and Huang, Fei and Chen, Huajun},
  journal={Advances in Neural Information Processing Systems},
  volume={37},
  pages={114843--114871},
  year={2024}
}

@article{schneider2025generative,
  title={Generative to agentic ai: Survey, conceptualization, and challenges},
  author={Schneider, Johannes},
  journal={arXiv preprint arXiv:2504.18875},
  year={2025}
}

@article{borghoff2025human,
  title={Human-artificial interaction in the age of agentic AI: a system-theoretical approach},
  author={Borghoff, Uwe M and Bottoni, Paolo and Pareschi, Remo},
  journal={Frontiers in Human Dynamics},
  volume={7},
  pages={1579166},
  year={2025},
  publisher={Frontiers Media SA}
}

@article{de2025open,
  title={Open Challenges in Multi-Agent Security: Towards Secure Systems of Interacting AI Agents},
  author={de Witt, Christian Schroeder},
  journal={arXiv preprint arXiv:2505.02077},
  year={2025}
}

@article{ke2025meta,
  title={Meta-Design Matters: A Self-Design Multi-Agent System},
  author={Ke, Zixuan and Xu, Austin and Ming, Yifei and Nguyen, Xuan-Phi and Xiong, Caiming and Joty, Shafiq},
  journal={arXiv preprint arXiv:2505.14996},
  year={2025}
}

@inproceedings{miao2025autonomous,
  title={Autonomous LLM-enhanced adversarial attack for text-to-motion},
  author={Miao, Honglei and Ma, Fan and Quan, Ruijie and Zhan, Kun and Yang, Yi},
  booktitle={Proceedings of the AAAI Conference on Artificial Intelligence},
  volume={39},
  number={6},
  pages={6144--6152},
  year={2025}
}

@article{tran2025multi,
  title={Multi-Agent Collaboration Mechanisms: A Survey of LLMs},
  author={Tran, Khanh-Tung and Dao, Dung and Nguyen, Minh-Duong and Pham, Quoc-Viet and O'Sullivan, Barry and Nguyen, Hoang D},
  journal={arXiv preprint arXiv:2501.06322},
  year={2025}
}

@article{feng2025multi,
  title={Multi-agent embodied ai: Advances and future directions},
  author={Feng, Zhaohan and Xue, Ruiqi and Yuan, Lei and Yu, Yang and Ding, Ning and Liu, Meiqin and Gao, Bingzhao and Sun, Jian and Wang, Gang},
  journal={arXiv preprint arXiv:2505.05108},
  year={2025}
}

@article{lazaros2024federated,
  title={Federated Learning: Navigating the Landscape of Collaborative Intelligence},
  author={Lazaros, Konstantinos and Koumadorakis, Dimitrios E and Vrahatis, Aristidis G and Kotsiantis, Sotiris},
  journal={Electronics},
  volume={13},
  number={23},
  pages={4744},
  year={2024},
  publisher={MDPI}
}

@article{luzon2024tutorial,
  title={A tutorial on federated learning from theory to practice: Foundations, software frameworks, exemplary use cases, and selected trends},
  author={Luz{\'o}n, M Victoria and Rodr{\'\i}guez-Barroso, Nuria and Argente-Garrido, Alberto and Jim{\'e}nez-L{\'o}pez, Daniel and Moyano, Jose M and Del Ser, Javier and Ding, Weiping and Herrera, Francisco},
  journal={IEEE/CAA Journal of Automatica Sinica},
  volume={11},
  number={4},
  pages={824--850},
  year={2024},
  publisher={IEEE}
}

@article{dawid2025agentic,
  title={Agentic Workflows for Economic Research: Design and Implementation},
  author={Dawid, Herbert and Harting, Philipp and Wang, Hankui and Wang, Zhongli and Yi, Jiachen},
  journal={arXiv preprint arXiv:2504.09736},
  year={2025}
}

@book{hexmoor2025behaviour,
  title={Behaviour based AI, cognitive processes, and emergent behaviors in autonomous agents},
  author={Hexmoor, Henry and Lammens, Johan and Caicedo, Guido and Shapiro, Stuart C},
  volume={1},
  year={2025},
  publisher={WIT Press}
}

@article{ghafarollahi2024sciagents,
  title={Sciagents: Automating scientific discovery through multi-agent intelligent graph reasoning},
  author={Ghafarollahi, Alireza and Buehler, Markus J},
  journal={arXiv preprint arXiv:2409.05556},
  year={2024}
}

@article{chen2025human,
  title={Human-in-the-Loop Robot Learning for Smart Manufacturing: A Human-Centric Perspective},
  author={Chen, Hongpeng and Li, Shufei and Fan, Junming and Duan, Anqing and Yang, Chenguang and Navarro-Alarcon, David and Zheng, Pai},
  journal={IEEE Transactions on Automation Science and Engineering},
  year={2025},
  publisher={IEEE}
}

@article{bran2023chemcrow,
  title={Chemcrow: Augmenting large-language models with chemistry tools},
  author={Bran, Andres M and Cox, Sam and Schilter, Oliver and Baldassari, Carlo and White, Andrew D and Schwaller, Philippe},
  journal={arXiv preprint arXiv:2304.05376},
  year={2023}
}

@inproceedings{ribeiro2016should,
  title={" Why should i trust you?" Explaining the predictions of any classifier},
  author={Ribeiro, Marco Tulio and Singh, Sameer and Guestrin, Carlos},
  booktitle={Proceedings of the 22nd ACM SIGKDD international conference on knowledge discovery and data mining},
  pages={1135--1144},
  year={2016}
}

@article{rosenfeld2019explainability,
  title={Explainability in human--agent systems},
  author={Rosenfeld, Avi and Richardson, Ariella},
  journal={Autonomous agents and multi-agent systems},
  volume={33},
  pages={673--705},
  year={2019},
  publisher={Springer}
}

@article{yossef2024explainable,
  title={Explainable artificial intelligence framework for FRP composites design},
  author={Yossef, Mostafa and Noureldin, Mohamed and Alqabbany, Aghyad},
  journal={Composite Structures},
  volume={341},
  pages={118190},
  year={2024},
  publisher={Elsevier}
}

@misc{Euler2023AutoGPTRCE,
  author       = {Lukas Euler},
  title        = {Hacking Auto-GPT and Escaping Its Docker Container},
  howpublished = {\url{https://positive.security/blog/auto-gpt-rce}},
  year         = {2023},
  note         = {Accessed 27 Jun 2025}
}

@article{tupayachi2024towards,
  title={Towards next-generation urban decision support systems through ai-powered construction of scientific ontology using large language models—A case in optimizing intermodal freight transportation},
  author={Tupayachi, Jose and Xu, Haowen and Omitaomu, Olufemi A and Camur, Mustafa Can and Sharmin, Aliza and Li, Xueping},
  journal={Smart Cities},
  volume={7},
  number={5},
  pages={2392--2421},
  year={2024},
  publisher={MDPI}
}

@article{bai2024derived,
  title={A derived information framework for a dynamic knowledge graph and its application to smart cities},
  author={Bai, Jiaru and Lee, Kok Foong and Hofmeister, Markus and Mosbach, Sebastian and Akroyd, Jethro and Kraft, Markus},
  journal={Future Generation Computer Systems},
  volume={152},
  pages={112--126},
  year={2024},
  publisher={Elsevier}
}

@article{tian2025outlook,
  title={An Outlook on the Opportunities and Challenges of Multi-Agent AI Systems},
  author={Tian, Fangqiao and Luo, An and Du, Jin and Xian, Xun and Specht, Robert and Wang, Ganghua and Bi, Xuan and Zhou, Jiawei and Srinivasa, Jayanth and Kundu, Ashish and others},
  journal={arXiv preprint arXiv:2505.18397},
  year={2025}
}

@inproceedings{liu2024ava,
  title={AVA: Towards Autonomous Visualization Agents through Visual Perception-Driven Decision-Making},
  author={Liu, Shusen and Miao, Haichao and Li, Zhimin and Olson, Matthew and Pascucci, Valerio and Bremer, P-T},
  booktitle={Computer Graphics Forum},
  volume={43},
  number={3},
  pages={e15093},
  year={2024},
  organization={Wiley Online Library}
}

@article{hu2024overview,
  title={An overview: Attention mechanisms in multi-agent reinforcement learning},
  author={Hu, Kai and Xu, Keer and Xia, Qingfeng and Li, Mingyang and Song, Zhiqiang and Song, Lipeng and Sun, Ning},
  journal={Neurocomputing},
  pages={128015},
  year={2024},
  publisher={Elsevier}
}

@article{gu2024artificial,
  title={Artificial intelligence co-piloted auditing},
  author={Gu, Hanchi and Schreyer, Marco and Moffitt, Kevin and Vasarhelyi, Miklos},
  journal={International Journal of Accounting Information Systems},
  volume={54},
  pages={100698},
  year={2024},
  publisher={Elsevier}
}

@article{chin2025automating,
  title={Automating Security Audit Using Large Language Model based Agent: An Exploration Experiment},
  author={Chin, Jia Hui and Zhang, Pu and Cheong, Yu Xin and Pan, Jonathan},
  journal={arXiv preprint arXiv:2505.10732},
  year={2025}
}

@article{sapkota2025ai,
title = {AI Agents vs. Agentic AI: A Conceptual taxonomy, applications and challenges},
journal = {Information Fusion},
volume = {126},
pages = {103599},
year = {2026},
issn = {1566-2535},
doi = {https://doi.org/10.1016/j.inffus.2025.103599},
url = {https://www.sciencedirect.com/science/article/pii/S1566253525006712},
author = {Ranjan Sapkota and Konstantinos I. Roumeliotis and Manoj Karkee}
}

@article{singh2025explainable,
  title={Explainable Reinforcement Learning Agents Using World Models},
  author={Singh, Madhuri and Alabdulkarim, Amal and Mansi, Gennie and Riedl, Mark O},
  journal={arXiv preprint arXiv:2505.08073},
  year={2025}
}

@article{phillips2021four,
  title={Four principles of explainable artificial intelligence},
  author={Phillips, P Jonathon and Phillips, P Jonathon and Hahn, Carina A and Fontana, Peter C and Yates, Amy N and Greene, Kristen and Broniatowski, David A and Przybocki, Mark A},
  year={2021},
  publisher={US Department of Commerce, National Institute of Standards and Technology}
}

@article{he2024security,
  title={Security of ai agents},
  author={He, Yifeng and Wang, Ethan and Rong, Yuyang and Cheng, Zifei and Chen, Hao},
  journal={arXiv preprint arXiv:2406.08689},
  year={2024}
}

@article{feretzakis2024privacy,
  title={Privacy-Preserving Techniques in Generative AI and Large Language Models: A Narrative Review},
  author={Feretzakis, Georgios and Papaspyridis, Konstantinos and Gkoulalas-Divanis, Aris and Verykios, Vassilios S},
  journal={Information},
  volume={15},
  number={11},
  pages={697},
  year={2024},
  publisher={MDPI}
}

@incollection{khan2024role,
  title={Role-based access control (rbac) and attribute-based access control (abac)},
  author={Khan, Javed Akhtar},
  booktitle={Improving security, privacy, and trust in cloud computing},
  pages={113--126},
  year={2024},
  publisher={IGI Global Scientific Publishing}
}

@article{zhang2024breaking,
  title={Breaking agents: Compromising autonomous llm agents through malfunction amplification},
  author={Zhang, Boyang and Tan, Yicong and Shen, Yun and Salem, Ahmed and Backes, Michael and Zannettou, Savvas and Zhang, Yang},
  journal={arXiv preprint arXiv:2407.20859},
  year={2024}
}

@article{standen2025adversarial,
  title={Adversarial Machine Learning Attacks and Defences in Multi-Agent Reinforcement Learning},
  author={Standen, Maxwell and Kim, Junae and Szabo, Claudia},
  journal={ACM Computing Surveys},
  volume={57},
  number={5},
  pages={1--35},
  year={2025},
  publisher={ACM New York, NY}
}

@article{zhen2025novel,
  title={A novel malware detection method based on audit logs and graph neural network},
  author={Zhen, Yewei and Tian, Donghai and Fu, Xiaohu and Hu, Changzhen},
  journal={Engineering Applications of Artificial Intelligence},
  volume={152},
  pages={110524},
  year={2025},
  publisher={Elsevier}
}

@article{raza2025responsible,
  title={Who is Responsible? The Data, Models, Users or Regulations? A Comprehensive Survey on Responsible Generative AI for a Sustainable Future},
  author={Raza, Shaina and Qureshi, Rizwan and Zahid, Anam and Fioresi, Joseph and Sadak, Ferhat and Saeed, Muhammad and Sapkota, Ranjan and Jain, Aditya and Zafar, Anas and Hassan, Muneeb Ul and others},
  journal={arXiv preprint arXiv:2502.08650},
  year={2025}
}

@inproceedings{zhong2023towards,
  title={Towards safe ai: Sandboxing dnns-based controllers in stochastic games},
  author={Zhong, Bingzhuo and Cao, Hongpeng and Zamani, Majid and Caccamo, Marco},
  booktitle={Proceedings of the AAAI Conference on Artificial Intelligence},
  volume={37},
  number={12},
  pages={15340--15349},
  year={2023}
}

@misc{guardrails_ai_guardrails_2024,
    title = {Guardrails {AI} {\textbar} {Your} {Enterprise} {AI} needs {Guardrails} — guardrailsai.com},
    url = {https://www.guardrailsai.com/docs/},
    urldate = {2024-02-01},
    author = {Guardrails AI},
    month = feb,
    year = {2024},
}

@article{raza_fair_2024,
    title = {{FAIR} {Enough}: {How} {Can} {We} {Develop} and {Assess} a {FAIR}-{Compliant} {Dataset} for {Large} {Language} {Models}' {Training}?},
    journal = {arXiv preprint arXiv:2401.11033},
    author = {Raza, Shaina and Ghuge, Shardul and Ding, Chen and Pandya, Deval},
    year = {2024},
}

@misc{ganguli_red_2022,
    title = {Red {Teaming} {Language} {Models} to {Reduce} {Harms}: {Methods}, {Scaling} {Behaviors}, and {Lessons} {Learned}},
    shorttitle = {Red {Teaming} {Language} {Models} to {Reduce} {Harms}},
    url = {http://arxiv.org/abs/2209.07858},
    urldate = {2024-01-19},
    publisher = {arXiv},
    author = {Ganguli, Deep and Lovitt, Liane and Kernion, Jackson and Askell, Amanda and Bai, Yuntao and Kadavath, Saurav and Mann, Ben and Perez, Ethan and Schiefer, Nicholas and Ndousse, Kamal and Jones, Andy and Bowman, Sam and Chen, Anna and Conerly, Tom and DasSarma, Nova and Drain, Dawn and Elhage, Nelson and El-Showk, Sheer and Fort, Stanislav and Hatfield-Dodds, Zac and Henighan, Tom and Hernandez, Danny and Hume, Tristan and Jacobson, Josh and Johnston, Scott and Kravec, Shauna and Olsson, Catherine and Ringer, Sam and Tran-Johnson, Eli and Amodei, Dario and Brown, Tom and Joseph, Nicholas and McCandlish, Sam and Olah, Chris and Kaplan, Jared and Clark, Jack},
    month = nov,
    year = {2022},
    note = {arXiv:2209.07858 [cs]},
}
\newpage
\section*{Appendix}
\renewcommand{\thefigure}{A.\arabic{figure}}
\renewcommand{\thetable}{A.\arabic{table}}
\setcounter{figure}{0}
\setcounter{table}{0}

\begin{table}[h]
\centering
\footnotesize
\caption{List of abbreviations used in this paper.}
\label{tab:abbreviations}
\begin{tabular}{ll}
\hline
\textbf{Abbreviation} & \textbf{Description} \\
\hline
AI & Artificial Intelligence \\
AMAS & Agentic Multi-Agent Systems \\
CoT & Chain-of-Thought \\
CSS & Component Synergy Score \\
DP & Differential Privacy \\
FHE & Fully Homomorphic Encryption \\
HITL & Human-in-the-Loop \\
HE & Homomorphic Encryption \\
IAM & Identity and Access Management \\
LLM & Large Language Model \\
MLOps & Machine Learning Operations \\
ModelOps & Model Lifecycle Operations, Monitoring, and Governance \\
MPC & Secure Multi-Party Computation \\
PII & Personally Identifiable Information \\
RAG & Retrieval-Augmented Generation \\
RBAC & Role-Based Access Control \\
RMF & Risk Management Framework \\
TEEs & Trusted Execution Environments \\
TRiSM & Trust, Risk, and Security Management \\
TUE & Tool Utilization Efficacy \\
XAI & Explainable Artificial Intelligence \\
\hline
\end{tabular}
\end{table}

\begin{figure}[h]
  \centering
  \includegraphics[width=0.4\textwidth]{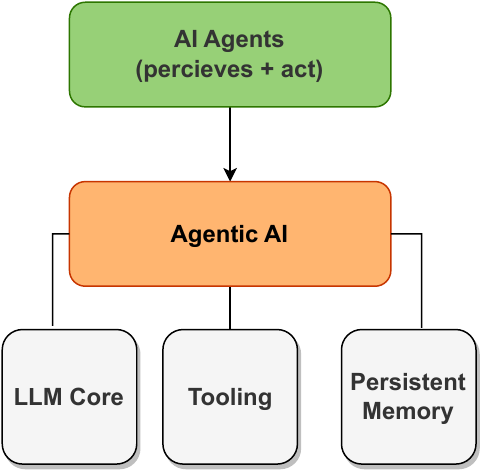}
  \caption{Traditional AI Agent vs. Agentic AI.}
  \label{fig:agentic_diagram}
\end{figure}

\begin{table}[h]
\centering
\footnotesize
\caption{Key terminologies for LLM‐based agentic AI systems.}
\renewcommand{\arraystretch}{.9}
\resizebox{\textwidth}{!}{
\begin{tabular}{p{0.20\textwidth} p{0.8\textwidth}}
\toprule
\textbf{Term} & \textbf{Definition} \\
\midrule
Agentic AI System &
A multi-agent artificial intelligence system powered by large language models (LLMs), in which multiple autonomous agents coordinate, communicate, and reason collectively to plan and execute tasks over extended horizons. Such systems typically incorporate persistent memory, tool use, and dynamic role assignment to support adaptive, goal-driven behavior. \\

Autonomy Model &
The internal decision-making mechanism that governs how an agent selects goals, plans actions, and executes tasks without continuous human intervention, often leveraging planning, reasoning traces, feedback signals, and environmental context. \\

Chain of Thought (CoT) &
A prompting and reasoning paradigm in which an LLM explicitly generates intermediate reasoning steps before producing a final output, enabling multi-step problem solving, improved transparency, and more controllable agent behavior. \\

Counterfactual Analysis &
An interpretability and diagnostic technique that evaluates how changes to inputs, agent roles, or internal states would alter system outcomes, helping reveal causal dependencies, failure modes, and contribution disparities across agents. \\

Explainability &
The ability of an AI system or individual agent to provide human-understandable rationales for its decisions, actions, or outputs, often supported by techniques such as feature attribution, decision provenance, or structured reasoning traces. \\

Foundation Model (LLM) &
A large pretrained language model (e.g., GPT-4, LLaMA) that serves as the core reasoning and generation component of an agent, providing linguistic competence, contextual understanding, tool invocation, and general-purpose reasoning capabilities. \\

Shared Memory (Persistent Memory) &
A centralized or distributed memory substrate, commonly implemented using vector databases or structured logs, that enables agents to store, retrieve, and share contextual information across time, supporting long-term coherence and coordination. \\

ModelOps &
A set of operational practices for managing models and agent prompts throughout their lifecycle, including versioning, deployment, monitoring, auditing, and retirement, with an emphasis on reliability, reproducibility, and governance. \\

Application Security &
Protective mechanisms and design practices that safeguard agentic systems against adversarial behaviors such as prompt injection, identity spoofing, privilege escalation, and cross-agent manipulation. \\

Model Privacy &
Privacy-preserving techniques that ensure sensitive data remains protected during model training, inference, and inter-agent communication, including differential privacy, secure computation, and trusted execution environments. \\

Prompt Injection &
A class of security attacks in which maliciously crafted inputs introduce hidden or indirect instructions that override or corrupt an agent’s intended behavior, potentially propagating harmful actions across agent interactions. \\

Retrieval-Augmented Generation (RAG) &
An architectural approach in which agents dynamically retrieve relevant external knowledge (e.g., documents or facts) from a datastore and condition LLM generation on this retrieved context to improve factuality and grounding. \\

Role-Specialized Coordination &
A coordination strategy where agents are assigned distinct functional roles (e.g., planner, verifier, executor) and interact via structured communication protocols to collectively solve complex tasks more efficiently and robustly. \\

Decision Provenance Graph &
A structured, graph-based representation that records data flows, reasoning steps, and inter-agent interactions, enabling post-hoc auditing, accountability analysis, and system-level interpretability. \\

Tool-Use Interface &
The mechanism that allows agents to issue structured commands to external tools, APIs, or execution environments and integrate returned results into subsequent reasoning and decision-making steps. \\

Trust Score &
A composite quantitative indicator of an agent’s reliability and alignment, typically combining task performance, safety compliance, behavioral consistency, and calibration of expressed confidence over time. \\

Composite Metric &
An aggregated evaluation measure that combines multiple dimensions—such as trustworthiness, explainability, coordination quality, and user-centered performance—to benchmark and compare agentic AI systems holistically. \\

\bottomrule
\end{tabular}
}
\label{tab:appendix_terminology}
\end{table}

\end{document}